\documentclass[10pt]{article}
\pdftrailerid{}
\usepackage[margin=1in]{geometry}
\usepackage{microtype}
\usepackage{amsmath,amssymb,mathtools}
\usepackage{booktabs}
\usepackage{graphicx}
\usepackage[dvipsnames]{xcolor}
\usepackage[most]{tcolorbox}
\usepackage{enumitem}
\usepackage{caption}
\usepackage{longtable}
\usepackage[section]{placeins}
\usepackage[round,authoryear]{natbib}
\usepackage[colorlinks=true,linkcolor=MidnightBlue,citecolor=MidnightBlue,urlcolor=MidnightBlue]{hyperref}
\usepackage{algorithm} 
\usepackage{algpseudocode}
\algtext*{EndFor}
\usepackage{placeins}
 \usepackage{multirow}
\usepackage{pifont}
\usepackage{xcolor}
\usepackage{capt-of}
\usepackage[nameinlink,capitalise]{cleveref}
\definecolor{bestcolor}{named}{Green}

\newcommand{\fpilotabs}{\color{Green}\ensuremath{\mathtt{FinPILOT}}}

\newcommand{\cmark}{\textcolor{green!60!black}{\ding{51}}}
\newcommand{\xmark}{\textcolor{red!70!black}{\ding{55}}}

\makeatletter
\newcommand{\NUMe}[1]{\@ifundefined{num@#1}{\textbf{\textcolor{red}{??#1??}}\PackageWarning{numbers}{Undefined number #1}}{\@nameuse{num@#1}}}
\makeatother
\expandafter\def\csname num@DoneCBC\endcsname{99.9}
\expandafter\def\csname num@DoneCBCQ\endcsname{99.9}
\expandafter\def\csname num@DoneCCQL\endcsname{99.9}
\expandafter\def\csname num@DoneCDQN\endcsname{$-$5.2}
\expandafter\def\csname num@DoneCIQL\endcsname{99.9}
\expandafter\def\csname num@DoneCMB\endcsname{99.8}
\expandafter\def\csname num@DoneCmin\endcsname{99.5}
\expandafter\def\csname num@DoneCtabMB\endcsname{99.7}
\expandafter\def\csname num@DoneCtabSC\endcsname{99.9}
\expandafter\def\csname num@DoneNJCons\endcsname{CQL}
\expandafter\def\csname num@DoneNJEBC\endcsname{90.3}
\expandafter\def\csname num@DoneNJEBCQa\endcsname{90.3}
\expandafter\def\csname num@DoneNJEBCQsmall\endcsname{65.9}
\expandafter\def\csname num@DoneNJECQLa\endcsname{90.3}
\expandafter\def\csname num@DoneNJECQLsmall\endcsname{86.0}
\expandafter\def\csname num@DoneNJECons\endcsname{90.3}
\expandafter\def\csname num@DoneNJEDQN\endcsname{$-$5.7}
\expandafter\def\csname num@DoneNJEIQLa\endcsname{90.3}
\expandafter\def\csname num@DoneNJEMB\endcsname{99.7}
\expandafter\def\csname num@DoneNJEMBa\endcsname{97.1}
\expandafter\def\csname num@DoneNJEMBo\endcsname{99.7}
\expandafter\def\csname num@DoneNJEMBone\endcsname{97.1}
\expandafter\def\csname num@DoneNJEbest\endcsname{99.7}
\expandafter\def\csname num@DoneNJEdiff\endcsname{$-$9.1}
\expandafter\def\csname num@DoneNJEdiffci\endcsname{$-$9.1 [$-$10.1, $-$8.0]}
\expandafter\def\csname num@DoneNJEtabMB\endcsname{99.8}
\expandafter\def\csname num@DoneNJEtabSC\endcsname{90.3}
\expandafter\def\csname num@DoneNJLossCons\endcsname{9.5}
\expandafter\def\csname num@DoneNJLossMB\endcsname{55.8}
\expandafter\def\csname num@DoneNJN\endcsname{20}
\expandafter\def\csname num@DoneNJNtrans\endcsname{9,089}
\expandafter\def\csname num@DoneNJRBC\endcsname{97.6}
\expandafter\def\csname num@DoneNJRBCQa\endcsname{98.8}
\expandafter\def\csname num@DoneNJRBCQsmall\endcsname{49.5}
\expandafter\def\csname num@DoneNJRCQLa\endcsname{99.7}
\expandafter\def\csname num@DoneNJRCQLsmall\endcsname{70.5}
\expandafter\def\csname num@DoneNJRCons\endcsname{99.7}
\expandafter\def\csname num@DoneNJRDQN\endcsname{$-$3.8}
\expandafter\def\csname num@DoneNJRIQLa\endcsname{99.8}
\expandafter\def\csname num@DoneNJRMB\endcsname{44.0}
\expandafter\def\csname num@DoneNJRMBa\endcsname{47.4}
\expandafter\def\csname num@DoneNJRMBo\endcsname{47.4}
\expandafter\def\csname num@DoneNJRMBone\endcsname{47.4}
\expandafter\def\csname num@DoneNJRbest\endcsname{99.8}
\expandafter\def\csname num@DoneNJRdiff\endcsname{$+$57.0}
\expandafter\def\csname num@DoneNJRdiffci\endcsname{$+$57.0 [$+$51.1, $+$63.2]}
\expandafter\def\csname num@DoneNJRtabMB\endcsname{42.8}
\expandafter\def\csname num@DoneNJRtabSC\endcsname{99.9}
\expandafter\def\csname num@DoneNJhidMB\endcsname{0.81}
\expandafter\def\csname num@DoneNJhidMBone\endcsname{0.50}
\expandafter\def\csname num@DoneNJuHidden\endcsname{0.133}
\expandafter\def\csname num@DoneNJuLogged\endcsname{0.045}
\expandafter\def\csname num@DoneNJuOther\endcsname{0.113}
\expandafter\def\csname num@DoneNJuRatio\endcsname{1.17}
\expandafter\def\csname num@DoneNVCons\endcsname{CQL}
\expandafter\def\csname num@DoneNVEBC\endcsname{93.6}
\expandafter\def\csname num@DoneNVEBCQa\endcsname{93.6}
\expandafter\def\csname num@DoneNVEBCQsmall\endcsname{46.5}
\expandafter\def\csname num@DoneNVECQLa\endcsname{93.7}
\expandafter\def\csname num@DoneNVECQLsmall\endcsname{85.1}
\expandafter\def\csname num@DoneNVECons\endcsname{93.7}
\expandafter\def\csname num@DoneNVEDQN\endcsname{$-$3.5}
\expandafter\def\csname num@DoneNVEIQLa\endcsname{93.7}
\expandafter\def\csname num@DoneNVEMB\endcsname{82.0}
\expandafter\def\csname num@DoneNVEMBa\endcsname{98.2}
\expandafter\def\csname num@DoneNVEMBo\endcsname{98.2}
\expandafter\def\csname num@DoneNVEMBone\endcsname{98.2}
\expandafter\def\csname num@DoneNVEbest\endcsname{98.2}
\expandafter\def\csname num@DoneNVEdiff\endcsname{$+$13.0}
\expandafter\def\csname num@DoneNVEdiffci\endcsname{$+$13.0 [$+$7.9, $+$18.2]}
\expandafter\def\csname num@DoneNVEtabMB\endcsname{99.6}
\expandafter\def\csname num@DoneNVEtabSC\endcsname{93.8}
\expandafter\def\csname num@DoneNVLossCons\endcsname{4.5}
\expandafter\def\csname num@DoneNVLossMB\endcsname{53.2}
\expandafter\def\csname num@DoneNVN\endcsname{20}
\expandafter\def\csname num@DoneNVNtrans\endcsname{12,224}
\expandafter\def\csname num@DoneNVRBC\endcsname{99.5}
\expandafter\def\csname num@DoneNVRBCQa\endcsname{99.6}
\expandafter\def\csname num@DoneNVRBCQsmall\endcsname{43.2}
\expandafter\def\csname num@DoneNVRCQLa\endcsname{99.8}
\expandafter\def\csname num@DoneNVRCQLsmall\endcsname{88.6}
\expandafter\def\csname num@DoneNVRCons\endcsname{99.8}
\expandafter\def\csname num@DoneNVRDQN\endcsname{$-$2.7}
\expandafter\def\csname num@DoneNVRIQLa\endcsname{99.8}
\expandafter\def\csname num@DoneNVRMB\endcsname{46.6}
\expandafter\def\csname num@DoneNVRMBa\endcsname{59.3}
\expandafter\def\csname num@DoneNVRMBo\endcsname{59.3}
\expandafter\def\csname num@DoneNVRMBone\endcsname{59.3}
\expandafter\def\csname num@DoneNVRbest\endcsname{99.8}
\expandafter\def\csname num@DoneNVRdiff\endcsname{$+$54.9}
\expandafter\def\csname num@DoneNVRdiffci\endcsname{$+$54.9 [$+$49.4, $+$60.5]}
\expandafter\def\csname num@DoneNVRtabMB\endcsname{46.2}
\expandafter\def\csname num@DoneNVRtabSC\endcsname{99.8}
\expandafter\def\csname num@DoneNVhidMB\endcsname{0.19}
\expandafter\def\csname num@DoneNVhidMBone\endcsname{0.10}
\expandafter\def\csname num@DoneNVuHidden\endcsname{0.193}
\expandafter\def\csname num@DoneNVuLogged\endcsname{0.036}
\expandafter\def\csname num@DoneNVuOther\endcsname{0.138}
\expandafter\def\csname num@DoneNVuRatio\endcsname{1.41}
\expandafter\def\csname num@DoneSPCons\endcsname{IQL}
\expandafter\def\csname num@DoneSPEBC\endcsname{85.1}
\expandafter\def\csname num@DoneSPEBCQa\endcsname{85.0}
\expandafter\def\csname num@DoneSPEBCQsmall\endcsname{61.7}
\expandafter\def\csname num@DoneSPECQLa\endcsname{85.0}
\expandafter\def\csname num@DoneSPECQLsmall\endcsname{82.1}
\expandafter\def\csname num@DoneSPECons\endcsname{85.1}
\expandafter\def\csname num@DoneSPEDQN\endcsname{$-$5.6}
\expandafter\def\csname num@DoneSPEIQLa\endcsname{85.1}
\expandafter\def\csname num@DoneSPEMB\endcsname{99.7}
\expandafter\def\csname num@DoneSPEMBa\endcsname{99.7}
\expandafter\def\csname num@DoneSPEMBo\endcsname{99.7}
\expandafter\def\csname num@DoneSPEMBone\endcsname{98.9}
\expandafter\def\csname num@DoneSPEbest\endcsname{99.7}
\expandafter\def\csname num@DoneSPEdiff\endcsname{$-$14.4}
\expandafter\def\csname num@DoneSPEdiffci\endcsname{$-$14.4 [$-$16.1, $-$12.8]}
\expandafter\def\csname num@DoneSPEtabMB\endcsname{99.8}
\expandafter\def\csname num@DoneSPEtabSC\endcsname{85.0}
\expandafter\def\csname num@DoneSPLossCons\endcsname{14.7}
\expandafter\def\csname num@DoneSPLossMB\endcsname{33.3}
\expandafter\def\csname num@DoneSPN\endcsname{20}
\expandafter\def\csname num@DoneSPNtrans\endcsname{11,376}
\expandafter\def\csname num@DoneSPRBC\endcsname{99.9}
\expandafter\def\csname num@DoneSPRBCQa\endcsname{99.9}
\expandafter\def\csname num@DoneSPRBCQsmall\endcsname{62.4}
\expandafter\def\csname num@DoneSPRCQLa\endcsname{99.9}
\expandafter\def\csname num@DoneSPRCQLsmall\endcsname{98.8}
\expandafter\def\csname num@DoneSPRCons\endcsname{99.9}
\expandafter\def\csname num@DoneSPRDQN\endcsname{$-$3.1}
\expandafter\def\csname num@DoneSPRIQLa\endcsname{99.9}
\expandafter\def\csname num@DoneSPRMB\endcsname{66.6}
\expandafter\def\csname num@DoneSPRMBa\endcsname{67.0}
\expandafter\def\csname num@DoneSPRMBo\endcsname{69.2}
\expandafter\def\csname num@DoneSPRMBone\endcsname{71.3}
\expandafter\def\csname num@DoneSPRbest\endcsname{99.9}
\expandafter\def\csname num@DoneSPRdiff\endcsname{$+$33.6}
\expandafter\def\csname num@DoneSPRdiffci\endcsname{$+$33.6 [$+$31.2, $+$36.0]}
\expandafter\def\csname num@DoneSPRtabMB\endcsname{60.9}
\expandafter\def\csname num@DoneSPRtabSC\endcsname{99.9}
\expandafter\def\csname num@DoneSPhidMB\endcsname{0.43}
\expandafter\def\csname num@DoneSPhidMBone\endcsname{0.37}
\expandafter\def\csname num@DoneSPuHidden\endcsname{0.147}
\expandafter\def\csname num@DoneSPuLogged\endcsname{0.031}
\expandafter\def\csname num@DoneSPuOther\endcsname{0.134}
\expandafter\def\csname num@DoneSPuRatio\endcsname{1.08}
\expandafter\def\csname num@DthreeAdvApv\endcsname{$-$0.02}
\expandafter\def\csname num@DthreeAdvCCk\endcsname{$+$0.12}
\expandafter\def\csname num@DthreeAdvDataMax\endcsname{0.21}
\expandafter\def\csname num@DthreeAdvExclMass\endcsname{$+$0.13}
\expandafter\def\csname num@DthreeAdvExclShare\endcsname{$+$0.09}
\expandafter\def\csname num@DthreeAdvFifteen\endcsname{$+$4.1}
\expandafter\def\csname num@DthreeAdvHaz\endcsname{$-$0.81}
\expandafter\def\csname num@DthreeAdvNeg\endcsname{36}
\expandafter\def\csname num@DthreeAdvNlog\endcsname{$-$0.09}
\expandafter\def\csname num@DthreeAdvNtr\endcsname{$-$0.21}
\expandafter\def\csname num@DthreeAdvOptExcl\endcsname{$+$0.07}
\expandafter\def\csname num@DthreeAdvPos\endcsname{42}
\expandafter\def\csname num@DthreeAdvSACo\endcsname{$-$0.01}
\expandafter\def\csname num@DthreeAdvSixty\endcsname{$-$21.2}
\expandafter\def\csname num@DthreeAdvState\endcsname{$+$0.03}
\expandafter\def\csname num@DthreeAdvThirty\endcsname{$-$5.3}
\expandafter\def\csname num@DthreeAdvZero\endcsname{$+$12.3}
\expandafter\def\csname num@DthreeApv\endcsname{$-$0.30}
\expandafter\def\csname num@DthreeApvci\endcsname{[$-$0.50, $-$0.08]}
\expandafter\def\csname num@DthreeBC\endcsname{84.4}
\expandafter\def\csname num@DthreeBCQ\endcsname{89.2}
\expandafter\def\csname num@DthreeCCk\endcsname{$+$0.42}
\expandafter\def\csname num@DthreeCCkci\endcsname{[$+$0.20, $+$0.61]}
\expandafter\def\csname num@DthreeCCs\endcsname{$+$0.41}
\expandafter\def\csname num@DthreeCCsci\endcsname{[$+$0.19, $+$0.61]}
\expandafter\def\csname num@DthreeCQL\endcsname{84.2}
\expandafter\def\csname num@DthreeCorrSacoExcl\endcsname{$-$0.85}
\expandafter\def\csname num@DthreeDQN\endcsname{$-$6.2}
\expandafter\def\csname num@DthreeExclMass\endcsname{$+$0.28}
\expandafter\def\csname num@DthreeExclMassci\endcsname{[$+$0.05, $+$0.51]}
\expandafter\def\csname num@DthreeExclShare\endcsname{$+$0.33}
\expandafter\def\csname num@DthreeExclShareci\endcsname{[$+$0.09, $+$0.54]}
\expandafter\def\csname num@DthreeIQL\endcsname{85.2}
\expandafter\def\csname num@DthreeMB\endcsname{86.6}
\expandafter\def\csname num@DthreeN\endcsname{80}
\expandafter\def\csname num@DthreeNlog\endcsname{$+$0.12}
\expandafter\def\csname num@DthreeNlogci\endcsname{[$-$0.09, $+$0.31]}
\expandafter\def\csname num@DthreeNqFifteen\endcsname{20}
\expandafter\def\csname num@DthreeNqSixty\endcsname{20}
\expandafter\def\csname num@DthreeNqThirty\endcsname{20}
\expandafter\def\csname num@DthreeNqZero\endcsname{20}
\expandafter\def\csname num@DthreeNtr\endcsname{$+$0.03}
\expandafter\def\csname num@DthreeNtrci\endcsname{[$-$0.18, $+$0.23]}
\expandafter\def\csname num@DthreePosFifteen\endcsname{14}
\expandafter\def\csname num@DthreePosSixty\endcsname{2}
\expandafter\def\csname num@DthreePosThirty\endcsname{6}
\expandafter\def\csname num@DthreePosZero\endcsname{20}
\expandafter\def\csname num@DthreeSACo\endcsname{$-$0.33}
\expandafter\def\csname num@DthreeSACoci\endcsname{[$-$0.53, $-$0.10]}
\expandafter\def\csname num@DthreeSCApv\endcsname{$-$0.91}
\expandafter\def\csname num@DthreeSCCCk\endcsname{$+$0.88}
\expandafter\def\csname num@DthreeSCExclMass\endcsname{$+$0.90}
\expandafter\def\csname num@DthreeSCExclShare\endcsname{$+$0.84}
\expandafter\def\csname num@DthreeSCHaz\endcsname{$+$0.04}
\expandafter\def\csname num@DthreeSCNlog\endcsname{$-$0.60}
\expandafter\def\csname num@DthreeSCNtr\endcsname{$-$0.67}
\expandafter\def\csname num@DthreeSCOptExcl\endcsname{$+$0.94}
\expandafter\def\csname num@DthreeSCSACo\endcsname{$-$0.92}
\expandafter\def\csname num@DthreeSCState\endcsname{$-$0.50}
\expandafter\def\csname num@DthreeState\endcsname{$-$0.50}
\expandafter\def\csname num@DthreeStateci\endcsname{[$-$0.66, $-$0.32]}
\expandafter\def\csname num@DthreetabSC\endcsname{93.9}
\expandafter\def\csname num@DtwoAleaDiag\endcsname{0.92}
\expandafter\def\csname num@DtwoAleaX\endcsname{0.98}
\expandafter\def\csname num@DtwoCBC\endcsname{97.3}
\expandafter\def\csname num@DtwoCCQL\endcsname{99.5}
\expandafter\def\csname num@DtwoCIQL\endcsname{97.2}
\expandafter\def\csname num@DtwoCMBP\endcsname{99.0}
\expandafter\def\csname num@DtwoCMBPa\endcsname{99.5}
\expandafter\def\csname num@DtwoCMOPOa\endcsname{93.3}
\expandafter\def\csname num@DtwoCSAC\endcsname{11.7}
\expandafter\def\csname num@DtwoCTDBC\endcsname{98.4}
\expandafter\def\csname num@DtwoCTDthree\endcsname{$-$3.4}
\expandafter\def\csname num@DtwoCons\endcsname{TD3+BC}
\expandafter\def\csname num@DtwoEBC\endcsname{76.9}
\expandafter\def\csname num@DtwoEBCcrash\endcsname{0}
\expandafter\def\csname num@DtwoEBCspeed\endcsname{0.37}
\expandafter\def\csname num@DtwoECQL\endcsname{63.2}
\expandafter\def\csname num@DtwoECQLcrash\endcsname{0}
\expandafter\def\csname num@DtwoECQLone\endcsname{61.1}
\expandafter\def\csname num@DtwoECQLonecrash\endcsname{0}
\expandafter\def\csname num@DtwoECQLonespeed\endcsname{0.45}
\expandafter\def\csname num@DtwoECQLspeed\endcsname{0.47}
\expandafter\def\csname num@DtwoECQLten\endcsname{66.5}
\expandafter\def\csname num@DtwoECQLtencrash\endcsname{0}
\expandafter\def\csname num@DtwoECQLtenspeed\endcsname{0.48}
\expandafter\def\csname num@DtwoECons\endcsname{82.2}
\expandafter\def\csname num@DtwoEIQL\endcsname{77.8}
\expandafter\def\csname num@DtwoEIQLb\endcsname{78.2}
\expandafter\def\csname num@DtwoEIQLbcrash\endcsname{0}
\expandafter\def\csname num@DtwoEIQLbspeed\endcsname{0.39}
\expandafter\def\csname num@DtwoEIQLcrash\endcsname{0}
\expandafter\def\csname num@DtwoEIQLspeed\endcsname{0.38}
\expandafter\def\csname num@DtwoEMBP\endcsname{98.5}
\expandafter\def\csname num@DtwoEMBPa\endcsname{99.4}
\expandafter\def\csname num@DtwoEMBPacrash\endcsname{0}
\expandafter\def\csname num@DtwoEMBPaspeed\endcsname{1.00}
\expandafter\def\csname num@DtwoEMBPb\endcsname{99.3}
\expandafter\def\csname num@DtwoEMBPbcrash\endcsname{0}
\expandafter\def\csname num@DtwoEMBPbspeed\endcsname{1.00}
\expandafter\def\csname num@DtwoEMBPc\endcsname{98.6}
\expandafter\def\csname num@DtwoEMBPccrash\endcsname{0}
\expandafter\def\csname num@DtwoEMBPcrash\endcsname{0}
\expandafter\def\csname num@DtwoEMBPcspeed\endcsname{0.99}
\expandafter\def\csname num@DtwoEMBPd\endcsname{98.0}
\expandafter\def\csname num@DtwoEMBPdcrash\endcsname{0}
\expandafter\def\csname num@DtwoEMBPdspeed\endcsname{0.98}
\expandafter\def\csname num@DtwoEMBPspeed\endcsname{1.00}
\expandafter\def\csname num@DtwoEMOPO\endcsname{56.7}
\expandafter\def\csname num@DtwoEMOPOa\endcsname{88.5}
\expandafter\def\csname num@DtwoEMOPOacrash\endcsname{0}
\expandafter\def\csname num@DtwoEMOPOaspeed\endcsname{0.94}
\expandafter\def\csname num@DtwoEMOPOb\endcsname{57.7}
\expandafter\def\csname num@DtwoEMOPObcrash\endcsname{0}
\expandafter\def\csname num@DtwoEMOPObspeed\endcsname{0.82}
\expandafter\def\csname num@DtwoEMOPOc\endcsname{8.3}
\expandafter\def\csname num@DtwoEMOPOccrash\endcsname{0}
\expandafter\def\csname num@DtwoEMOPOcrash\endcsname{0}
\expandafter\def\csname num@DtwoEMOPOcspeed\endcsname{0.53}
\expandafter\def\csname num@DtwoEMOPOspeed\endcsname{0.90}
\expandafter\def\csname num@DtwoESAC\endcsname{5.7}
\expandafter\def\csname num@DtwoESACcrash\endcsname{0}
\expandafter\def\csname num@DtwoESACspeed\endcsname{0.86}
\expandafter\def\csname num@DtwoETDBC\endcsname{82.2}
\expandafter\def\csname num@DtwoETDBCcrash\endcsname{0}
\expandafter\def\csname num@DtwoETDBCfive\endcsname{85.1}
\expandafter\def\csname num@DtwoETDBCfivecrash\endcsname{0}
\expandafter\def\csname num@DtwoETDBCfivespeed\endcsname{0.52}
\expandafter\def\csname num@DtwoETDBCone\endcsname{79.4}
\expandafter\def\csname num@DtwoETDBConecrash\endcsname{0}
\expandafter\def\csname num@DtwoETDBConespeed\endcsname{0.41}
\expandafter\def\csname num@DtwoETDBCspeed\endcsname{0.46}
\expandafter\def\csname num@DtwoETDBCten\endcsname{87.3}
\expandafter\def\csname num@DtwoETDBCtencrash\endcsname{0}
\expandafter\def\csname num@DtwoETDBCtenspeed\endcsname{0.61}
\expandafter\def\csname num@DtwoETDthree\endcsname{4.0}
\expandafter\def\csname num@DtwoETDthreecrash\endcsname{0}
\expandafter\def\csname num@DtwoETDthreespeed\endcsname{1.00}
\expandafter\def\csname num@DtwoEbest\endcsname{99.4}
\expandafter\def\csname num@DtwoEdiff\endcsname{$-$16.0}
\expandafter\def\csname num@DtwoEdiffMOPO\endcsname{$+$24.8}
\expandafter\def\csname num@DtwoEdiffMOPOci\endcsname{$+$24.8 [$+$1.8, $+$46.7]}
\expandafter\def\csname num@DtwoEdiffci\endcsname{$-$16.0 [$-$16.7, $-$15.4]}
\expandafter\def\csname num@DtwoLossCons\endcsname{17.2}
\expandafter\def\csname num@DtwoLossMBP\endcsname{61.8}
\expandafter\def\csname num@DtwoLossMOPO\endcsname{72.4}
\expandafter\def\csname num@DtwoMbpCmin\endcsname{98.2}
\expandafter\def\csname num@DtwoMbpEmin\endcsname{96.8}
\expandafter\def\csname num@DtwoModelErr\endcsname{0.0012}
\expandafter\def\csname num@DtwoMopoCfail\endcsname{1}
\expandafter\def\csname num@DtwoMopoEfail\endcsname{5}
\expandafter\def\csname num@DtwoN\endcsname{10}
\expandafter\def\csname num@DtwoNC\endcsname{5}
\expandafter\def\csname num@DtwoNtrans\endcsname{6,975}
\expandafter\def\csname num@DtwoRBC\endcsname{87.1}
\expandafter\def\csname num@DtwoRBCcrash\endcsname{0}
\expandafter\def\csname num@DtwoRBCspeed\endcsname{0.37}
\expandafter\def\csname num@DtwoRCQL\endcsname{71.0}
\expandafter\def\csname num@DtwoRCQLcrash\endcsname{1}
\expandafter\def\csname num@DtwoRCQLone\endcsname{68.9}
\expandafter\def\csname num@DtwoRCQLonecrash\endcsname{1}
\expandafter\def\csname num@DtwoRCQLonespeed\endcsname{0.45}
\expandafter\def\csname num@DtwoRCQLspeed\endcsname{0.47}
\expandafter\def\csname num@DtwoRCQLten\endcsname{73.5}
\expandafter\def\csname num@DtwoRCQLtencrash\endcsname{2}
\expandafter\def\csname num@DtwoRCQLtenspeed\endcsname{0.48}
\expandafter\def\csname num@DtwoRCons\endcsname{87.9}
\expandafter\def\csname num@DtwoRIQL\endcsname{88.4}
\expandafter\def\csname num@DtwoRIQLb\endcsname{88.9}
\expandafter\def\csname num@DtwoRIQLbcrash\endcsname{0}
\expandafter\def\csname num@DtwoRIQLbspeed\endcsname{0.39}
\expandafter\def\csname num@DtwoRIQLcrash\endcsname{0}
\expandafter\def\csname num@DtwoRIQLspeed\endcsname{0.38}
\expandafter\def\csname num@DtwoRMBP\endcsname{27.6}
\expandafter\def\csname num@DtwoRMBPa\endcsname{28.2}
\expandafter\def\csname num@DtwoRMBPacrash\endcsname{75}
\expandafter\def\csname num@DtwoRMBPaspeed\endcsname{1.00}
\expandafter\def\csname num@DtwoRMBPb\endcsname{28.1}
\expandafter\def\csname num@DtwoRMBPbcrash\endcsname{75}
\expandafter\def\csname num@DtwoRMBPbspeed\endcsname{1.00}
\expandafter\def\csname num@DtwoRMBPc\endcsname{28.0}
\expandafter\def\csname num@DtwoRMBPccrash\endcsname{75}
\expandafter\def\csname num@DtwoRMBPcrash\endcsname{76}
\expandafter\def\csname num@DtwoRMBPcspeed\endcsname{0.98}
\expandafter\def\csname num@DtwoRMBPd\endcsname{27.9}
\expandafter\def\csname num@DtwoRMBPdcrash\endcsname{75}
\expandafter\def\csname num@DtwoRMBPdspeed\endcsname{0.97}
\expandafter\def\csname num@DtwoRMBPspeed\endcsname{1.00}
\expandafter\def\csname num@DtwoRMOPO\endcsname{17.0}
\expandafter\def\csname num@DtwoRMOPOa\endcsname{22.3}
\expandafter\def\csname num@DtwoRMOPOacrash\endcsname{75}
\expandafter\def\csname num@DtwoRMOPOaspeed\endcsname{0.94}
\expandafter\def\csname num@DtwoRMOPOb\endcsname{14.5}
\expandafter\def\csname num@DtwoRMOPObcrash\endcsname{65}
\expandafter\def\csname num@DtwoRMOPObspeed\endcsname{0.80}
\expandafter\def\csname num@DtwoRMOPOc\endcsname{1.6}
\expandafter\def\csname num@DtwoRMOPOccrash\endcsname{48}
\expandafter\def\csname num@DtwoRMOPOcrash\endcsname{71}
\expandafter\def\csname num@DtwoRMOPOcspeed\endcsname{0.51}
\expandafter\def\csname num@DtwoRMOPOspeed\endcsname{0.89}
\expandafter\def\csname num@DtwoRSAC\endcsname{$-$0.2}
\expandafter\def\csname num@DtwoRSACcrash\endcsname{75}
\expandafter\def\csname num@DtwoRSACspeed\endcsname{0.85}
\expandafter\def\csname num@DtwoRTDBC\endcsname{87.9}
\expandafter\def\csname num@DtwoRTDBCcrash\endcsname{7}
\expandafter\def\csname num@DtwoRTDBCfive\endcsname{55.8}
\expandafter\def\csname num@DtwoRTDBCfivecrash\endcsname{47}
\expandafter\def\csname num@DtwoRTDBCfivespeed\endcsname{0.51}
\expandafter\def\csname num@DtwoRTDBCone\endcsname{89.5}
\expandafter\def\csname num@DtwoRTDBConecrash\endcsname{1}
\expandafter\def\csname num@DtwoRTDBConespeed\endcsname{0.41}
\expandafter\def\csname num@DtwoRTDBCspeed\endcsname{0.45}
\expandafter\def\csname num@DtwoRTDBCten\endcsname{30.1}
\expandafter\def\csname num@DtwoRTDBCtencrash\endcsname{71}
\expandafter\def\csname num@DtwoRTDBCtenspeed\endcsname{0.59}
\expandafter\def\csname num@DtwoRTDthree\endcsname{$-$0.3}
\expandafter\def\csname num@DtwoRTDthreecrash\endcsname{69}
\expandafter\def\csname num@DtwoRTDthreespeed\endcsname{1.00}
\expandafter\def\csname num@DtwoRbest\endcsname{89.5}
\expandafter\def\csname num@DtwoRdiff\endcsname{$+$59.5}
\expandafter\def\csname num@DtwoRdiffMOPO\endcsname{$+$69.9}
\expandafter\def\csname num@DtwoRdiffMOPOci\endcsname{$+$69.9 [$+$62.7, $+$76.6]}
\expandafter\def\csname num@DtwoRdiffci\endcsname{$+$59.5 [$+$56.8, $+$61.8]}
\expandafter\def\csname num@DtwoURatioDiag\endcsname{1.41}
\expandafter\def\csname num@DtwoURatioMax\endcsname{2.17}
\expandafter\def\csname num@DtwoURatioX\endcsname{1.14}
\expandafter\def\csname num@HfiveDtwo\endcsname{yes}
\expandafter\def\csname num@HfiveNJ\endcsname{yes}
\expandafter\def\csname num@HfiveNV\endcsname{yes}
\expandafter\def\csname num@HfiveSP\endcsname{yes}
\expandafter\def\csname num@HfiveTj\endcsname{\textbf{no}}
\expandafter\def\csname num@HfiveTv\endcsname{\textbf{no}}
\expandafter\def\csname num@HfourDone\endcsname{yes}
\expandafter\def\csname num@HfourDtwo\endcsname{yes}
\expandafter\def\csname num@HoneDtwo\endcsname{yes$^a$}
\expandafter\def\csname num@HoneDtwoMopo\endcsname{\textbf{no} in $E$}
\expandafter\def\csname num@HoneNJ\endcsname{yes}
\expandafter\def\csname num@HoneNV\endcsname{\textbf{no} ($E$)}
\expandafter\def\csname num@HoneSP\endcsname{yes}
\expandafter\def\csname num@HoneTj\endcsname{yes}
\expandafter\def\csname num@HoneTv\endcsname{yes}
\expandafter\def\csname num@HsevenDthree\endcsname{\textbf{no} (1/5)}
\expandafter\def\csname num@HsevenDthreeOk\endcsname{1}
\expandafter\def\csname num@HsevenDthreeTot\endcsname{5}
\expandafter\def\csname num@HsevenT\endcsname{yes}
\expandafter\def\csname num@HsixDtwo\endcsname{yes}
\expandafter\def\csname num@HsixNJ\endcsname{yes}
\expandafter\def\csname num@HsixNV\endcsname{yes}
\expandafter\def\csname num@HsixSP\endcsname{yes}
\expandafter\def\csname num@HsixTj\endcsname{yes}
\expandafter\def\csname num@HsixTv\endcsname{yes}
\expandafter\def\csname num@HthreeDtwo\endcsname{yes}
\expandafter\def\csname num@HthreeNJ\endcsname{yes}
\expandafter\def\csname num@HthreeNV\endcsname{yes}
\expandafter\def\csname num@HthreeSP\endcsname{yes}
\expandafter\def\csname num@HthreeTj\endcsname{yes}
\expandafter\def\csname num@HthreeTv\endcsname{yes}
\expandafter\def\csname num@HtwoDtwo\endcsname{yes}
\expandafter\def\csname num@HtwoNJ\endcsname{yes}
\expandafter\def\csname num@HtwoNV\endcsname{yes}
\expandafter\def\csname num@HtwoSP\endcsname{yes}
\expandafter\def\csname num@HtwoTj\endcsname{yes}
\expandafter\def\csname num@HtwoTv\endcsname{yes}
\expandafter\def\csname num@TdiagCCk\endcsname{$+$0.93}
\expandafter\def\csname num@TdiagCCkSC\endcsname{$+$0.93}
\expandafter\def\csname num@TdiagCCs\endcsname{$+$0.66}
\expandafter\def\csname num@TdiagCCsSC\endcsname{$+$0.68}
\expandafter\def\csname num@TdiagCCsup\endcsname{$+$0.06}
\expandafter\def\csname num@TdiagCCsupSC\endcsname{$+$0.14}
\expandafter\def\csname num@TdiagExcl\endcsname{$+$0.64}
\expandafter\def\csname num@TdiagExclSC\endcsname{$+$0.64}
\expandafter\def\csname num@TdiagN\endcsname{1600}
\expandafter\def\csname num@TdiagNlog\endcsname{$-$0.15}
\expandafter\def\csname num@TdiagNlogSC\endcsname{$-$0.14}
\expandafter\def\csname num@TdiagSACo\endcsname{$-$0.11}
\expandafter\def\csname num@TdiagSACoSC\endcsname{$-$0.14}
\expandafter\def\csname num@TdiagState\endcsname{$-$0.10}
\expandafter\def\csname num@TdiagStateSC\endcsname{$-$0.12}
\expandafter\def\csname num@TfourBeatFifteen\endcsname{0}
\expandafter\def\csname num@TfourBeatSixty\endcsname{0}
\expandafter\def\csname num@TfourBeatThirty\endcsname{0}
\expandafter\def\csname num@TfourBeatZero\endcsname{8}
\expandafter\def\csname num@TfourLamFifteen\endcsname{0.1}
\expandafter\def\csname num@TfourLamSixty\endcsname{1.0}
\expandafter\def\csname num@TfourLamThirty\endcsname{0.2}
\expandafter\def\csname num@TfourLamZero\endcsname{0.0}
\expandafter\def\csname num@TfourLoseFifteen\endcsname{7}
\expandafter\def\csname num@TfourLoseSixty\endcsname{8}
\expandafter\def\csname num@TfourLoseThirty\endcsname{8}
\expandafter\def\csname num@TfourLoseZero\endcsname{0}
\expandafter\def\csname num@ToneJumpEBC\endcsname{89.7}
\expandafter\def\csname num@ToneJumpEGATE\endcsname{100.0}
\expandafter\def\csname num@ToneJumpEMB\endcsname{100.0}
\expandafter\def\csname num@ToneJumpEMBa\endcsname{91.0}
\expandafter\def\csname num@ToneJumpEMBo\endcsname{100.0}
\expandafter\def\csname num@ToneJumpESC\endcsname{89.8}
\expandafter\def\csname num@ToneJumpEdiff\endcsname{$-$10.7}
\expandafter\def\csname num@ToneJumpEdiffabs\endcsname{10.7}
\expandafter\def\csname num@ToneJumpLam\endcsname{1.0}
\expandafter\def\csname num@ToneJumpRBC\endcsname{96.2}
\expandafter\def\csname num@ToneJumpRGATE\endcsname{96.0}
\expandafter\def\csname num@ToneJumpRMB\endcsname{43.6}
\expandafter\def\csname num@ToneJumpRMBa\endcsname{96.0}
\expandafter\def\csname num@ToneJumpRMBo\endcsname{96.0}
\expandafter\def\csname num@ToneJumpRSC\endcsname{96.4}
\expandafter\def\csname num@ToneJumpRdiff\endcsname{$+$54.6}
\expandafter\def\csname num@ToneJumpRdiffabs\endcsname{54.6}
\expandafter\def\csname num@ToneVaultEBC\endcsname{92.6}
\expandafter\def\csname num@ToneVaultEGATE\endcsname{100.0}
\expandafter\def\csname num@ToneVaultEMB\endcsname{100.0}
\expandafter\def\csname num@ToneVaultEMBa\endcsname{94.0}
\expandafter\def\csname num@ToneVaultEMBo\endcsname{100.0}
\expandafter\def\csname num@ToneVaultESC\endcsname{92.8}
\expandafter\def\csname num@ToneVaultEdiff\endcsname{$-$7.6}
\expandafter\def\csname num@ToneVaultEdiffabs\endcsname{7.6}
\expandafter\def\csname num@ToneVaultLam\endcsname{1.0}
\expandafter\def\csname num@ToneVaultRBC\endcsname{97.0}
\expandafter\def\csname num@ToneVaultRGATE\endcsname{98.9}
\expandafter\def\csname num@ToneVaultRMB\endcsname{48.6}
\expandafter\def\csname num@ToneVaultRMBa\endcsname{99.3}
\expandafter\def\csname num@ToneVaultRMBo\endcsname{99.3}
\expandafter\def\csname num@ToneVaultRSC\endcsname{97.2}
\expandafter\def\csname num@ToneVaultRdiff\endcsname{$+$50.2}
\expandafter\def\csname num@ToneVaultRdiffabs\endcsname{50.2}
\expandafter\def\csname num@TthreeSGATE\endcsname{100.0}
\expandafter\def\csname num@TthreeSMB\endcsname{100.0}
\expandafter\def\csname num@TthreeSNAIVE\endcsname{94.2}
\expandafter\def\csname num@TthreeSSC\endcsname{78.5}
\expandafter\def\csname num@TthreeSbad\endcsname{36.9}
\expandafter\def\csname num@TthreeXGATE\endcsname{100.0}
\expandafter\def\csname num@TthreeXMB\endcsname{62.6}
\expandafter\def\csname num@TthreeXNAIVE\endcsname{100.0}
\expandafter\def\csname num@TthreeXSC\endcsname{100.0}
\expandafter\def\csname num@TthreeXnGATE\endcsname{62.6}
\expandafter\def\csname num@TthreeXnMB\endcsname{62.6}
\expandafter\def\csname num@TthreeXnNAIVE\endcsname{66.7}
\expandafter\def\csname num@TthreeXnSC\endcsname{100.0}
\expandafter\def\csname num@TtwoEMBanine\endcsname{99.9}
\expandafter\def\csname num@TtwoEMBazero\endcsname{91.0}
\expandafter\def\csname num@TtwoESCnine\endcsname{99.3}
\expandafter\def\csname num@TtwoESCseven\endcsname{99.2}
\expandafter\def\csname num@TtwoESCzero\endcsname{89.8}
\expandafter\def\csname num@TtwoOmegaNine\endcsname{0.88}
\expandafter\def\csname num@TtwoRMBanine\endcsname{96.0}
\expandafter\def\csname num@TtwoRMBnine\endcsname{43.6}
\expandafter\def\csname num@TtwoRMBzero\endcsname{43.6}

\setlist{topsep=2pt}

\title{{Plan Before You Trade: \\Inference-Time Optimization for RL Trading Agents}}
\author{\normalsize 
\makebox[\textwidth][c]{%
\begin{tabular}{c@{\hspace{2cm}}c}
\textbf{Eun Go} & \textbf{Rohan Deb} \\
Siebel School of Computing and Data Science & Siebel School of Computing and Data Science\\
University of Illinois Urbana-Champaign & University of Illinois Urbana-Champaign \\
{\texttt{eungo2@illinois.edu}} &
{\texttt{rd22@illinois.edu}}
\end{tabular}
}
\\[1.2em]
\makebox[\textwidth][c]{%
\begin{tabular}{c}
\textbf{Arindam Banerjee} \\
Siebel School of Computing and Data Science \\
University of Illinois Urbana-Champaign \\
{\texttt{arindamb@illinois.edu}}
\end{tabular}
}
}
\date{}

\begin{document}
\maketitle

\begin{abstract} \noindent 
Reinforcement learning agents for portfolio management are typically
trained and deployed as static policies, with no mechanism for policy adjustments using price forecasts at inference time. We propose $\fpilotabs$ (\textbf{Fin}ancial \textbf{P}lugin \textbf{I}nference-time \textbf{L}earning for \textbf{O}ptimal \textbf{T}rading), a plugin inference-time policy optimization framework inspired by Model Predictive Control (MPC). Under the price-taker assumption, portfolio actions do not affect future market prices, allowing a multi-step price forecaster to provide imagined market trajectories without iterative action-conditioned rollouts. At each decision step, $\fpilotabs$ uses the forecaster's predicted price trajectory to construct an allocation-based imagined return objective, and optimizes the pre-trained policy at inference-time before executing one step of the trade. Our framework is compatible with any pre-trained policy and adapts the policy using the forecaster's predictions without modifying the original RL training pipeline.
Evaluated with five pre-trained policies (PPO, SAC, A2C, TD3, DDPG) across DJ30 equities and foreign exchange (FX) over three rolling held-out test
years (30 settings in total), $\fpilotabs$ improves Total Return in 28 of
30 settings and Sharpe ratio in 27 of 30, with gains spanning both
stochastic and deterministic policies. Direct single- and multi-period portfolio
optimizers supplied with the same forecasts do not reproduce these aggregate
gains under our evaluation protocol. Further, using controlled forecasts at
calibrated quality levels, we show that both the magnitude and consistency of
$\fpilotabs$'s performance increase as forecast error is reduced.
\end{abstract}

\section{Introduction}
\label{sec:intro}
Reinforcement learning for financial trading (RLFT) has emerged as a
principled framework for sequential portfolio decision-making, with recent
benchmarks reporting competitive performance across a range of market
conditions~\citep{finrl_meta_NEURIPS2022_0bf54b80,
trademaster/NEURIPS2023_b8f6f7f2}.
At the same time, machine learning for
trading has developed along two largely separate lines: forecasting-based and RL-based methods.

The forecasting-based line of work treats prediction as the central
problem, using increasingly sophisticated models for short-horizon returns
or price movements, ranging from the linear-to-tree spectrum systematized by
\citet{gu_kelly_r2_10.1093/rfs/hhaa009} to LSTM and Transformer-based
movement predictors~\citep{feng2019enhancingstockmovementprediction,
yoo2021accurate}.
Out-of-sample financial predictability is typically weak, with $R^2$ often
below 1\% at monthly horizons~\citep{gu_kelly_r2_10.1093/rfs/hhaa009}. Moreover, in this paradigm, the mechanism translating
forecasts into portfolio allocations is designed separately from the
predictor.

The RL-based line of work instead learns the trading decision directly.
Building on the deep RL formulation of
\citet{jiang2017deepreinforcementlearningframework}, subsequent methods add
risk-aware objectives~\citep{deeptrader_Wang_Huang_Tu_Zhang_Xu_2021},
augmented state representations carrying price-prediction
signals~\citep{SARL_Ye_Pei_Wang_Chen_Zhu_Xiao_Li_2020}, and standardized
benchmarking platforms~\citep{trademaster/NEURIPS2023_b8f6f7f2}.
Forecasts can therefore be incorporated during policy training, but this
does not provide a general way to attach an independently trained forecaster
to an already pre-trained policy at inference time.

A related line of model-based RL performs planning through learned world
models, including Dreamer~\citep{Hafner2020Dream,Hafner2025DreamerV3},
MuZero~\citep{Schrittwieser2020MuZero}, and
TD-MPC2~\citep{hansen2024tdmpc}.
Most directly related to our setting,
\citet{deb2026modelpredictivecontroldifferentiable} perform inference-time
gradient-based policy adaptation through a learned differentiable world
model on continuous-control offline RL benchmarks.
This demonstrates inference-time policy optimization in that setting, but
leaves open how to connect such adaptation to an independently trained
financial forecaster.

The world-model perspective suggests a natural route.
Model Predictive Control (MPC) repeatedly optimizes over a finite future
horizon, executes the first resulting action, and replans after observing the
next state~\citep{mpc-garcia,mpc-mayne}.
Translating this paradigm to RLFT raises two key questions:
\textbf{(i)} What should an imagined world model contain in financial
markets?
\textbf{(ii)} Given a forecasted future, should one optimize portfolio
actions directly, as in classical financial
MPC~\citep{boyd_cvx_DBLP:journals/ftopt/BoydBDKKNS17}, or adapt the
parameters of a pre-trained policy?

For \textbf{(i)}, price-taking portfolio management has a useful structural
decomposition. Under the price-taker assumption, the agent's allocation does
not affect subsequent market prices, so the uncertain market component of an
imagined rollout can be supplied by an exogenous multi-step forecaster rather
than an action-conditioned transition model. Conditional on a forecasted
market path and the policy's allocations, portfolio wealth, transaction
costs, weight drift, and rewards follow exactly from known portfolio
mechanics. Thus, imagined rollouts decompose into an independently trained
market forecaster and closed-form transaction-cost-aware portfolio
accounting, without requiring a learned reward model.

For \textbf{(ii)}, we adapt the pre-trained policy itself rather than
optimizing an action sequence. At each real decision step, the forecaster
provides an $H$-step future trajectory, against which we optimize the actor
parameters using imagined portfolio returns and a downside-risk penalty.
Only one real action is then executed before the procedure is repeated, and
the adapted actor parameters are carried forward to the next decision.
The forecaster is trained independently, and the RL policy has no access to
its outputs during the original training procedure. We call this framework
$\fpilotabs$ (\emph{\textbf{Fin}ancial \textbf{P}lugin \textbf{I}nference-time \textbf{L}earning for \textbf{O}ptimal
\textbf{T}rading}).

We evaluate $\fpilotabs$ across five RL algorithms, two asset classes (DJ30 and FX), and
three rolling held-out test years. Across 30 settings, $\fpilotabs$ improves
Total Return in 28 and Sharpe ratio in 27, and direct portfolio optimizers
supplied with the same forecasts do not reproduce these aggregate gains.
Controlled forecast-skill experiments further show that the magnitude and
consistency of the gains increase as forecast error is reduced.

\textbf{Contributions.}
We summarize our contributions as follows:
\begin{enumerate}
    \item \textbf{A modular forecast-to-policy interface.}
    We introduce $\fpilotabs$, which uses an independently trained multi-step
    forecaster to adapt a pre-trained RL trading actor at inference time.
    Exploiting the price-taker structure of portfolio management, imagined
    rollouts combine exogenous forecasted market paths with exact
    transaction-cost-aware portfolio accounting, so no learned
    action-conditioned transition or reward model is needed and the original
    RL training pipeline is left unchanged.

    \item \textbf{Broad paired evaluation.}
    Across five RL algorithms, two asset classes, and three rolling held-out
    years, $\fpilotabs$ improves the frozen pre-trained policy in 28 of 30
    comparisons on Total Return and 27 of 30 on Sharpe ratio, while direct
    portfolio optimizers given the same forecasts do not reproduce these
    gains under our protocol.

    \item \textbf{A controlled forecast-skill protocol.}
    We rescale the errors of the forecasters used in our main experiments to
an exact target $R^2_\star$ and show that both the magnitude and
consistency of $\fpilotabs$'s Sharpe gains increase with forecast skill,
while the original low-skill forecasts already yield consistent Sharpe
improvements.
\end{enumerate}

\section{Problem Formulation}
\label{sec:problem_formulation}
\begin{figure}[ht]
  \centering
\includegraphics[width=\linewidth]{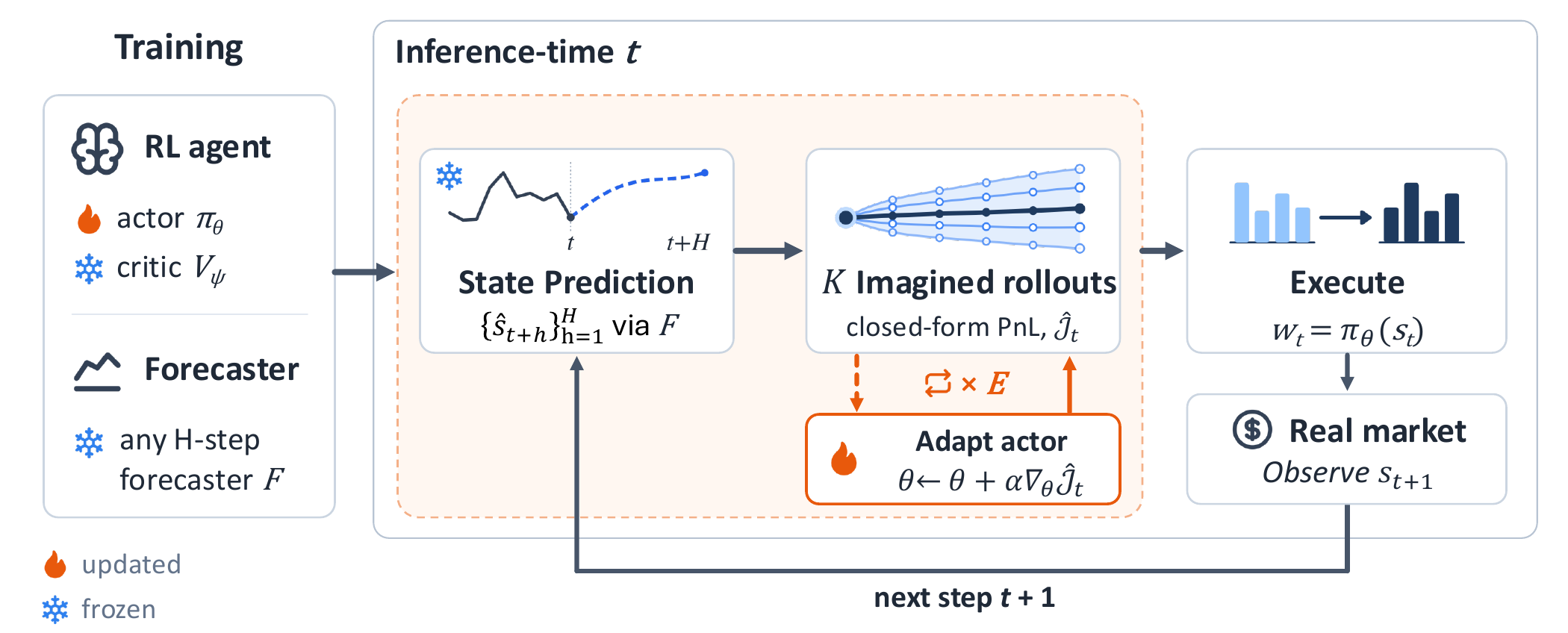}
\caption{Overview of $\fpilotabs$.
\textbf{Training:} an RL agent is
pre-trained without any forecaster, and an $H$-step forecaster $F$ is trained separately.
\textbf{Inference (at time $t$):} $K$ imagined trajectories are sampled once
around $F$'s predicted trajectory and reused across $E$ epochs, during which only the
actor is updated by gradient ascent on the risk-penalized objective
$\widehat{\mathcal{J}}_t$ (closed-form PnL with a terminal bootstrap from the frozen critic). The adapted actor executes $w_t=\pi_\theta(s_t)$, observes
$s_{t+1}$, and repeats. Mini-plots are schematic.}
\label{fig:architecture}
\end{figure}

Following the RLFT literature and the TradeMaster benchmarking framework \citep{liu2022finrldeepreinforcementlearning,trademaster/NEURIPS2023_b8f6f7f2,earnmore_10.1145/3589334.3645615}, we formalize portfolio management over a
universe of $N$ assets as a Markov Decision Process (MDP) defined by the
5-tuple $(S,A,\mathbb{P},R,\gamma)$, where $S$
is the state space, $A$ is the action space,
$\mathbb{P}:\mathcal{S}\times\mathcal{A}\times\mathcal{S}\to[0,1]$ is the
state-transition function determined by market dynamics,
$R:S\times A \to R$ is the scalar reward function,
and $\gamma\in[0,1)$ is the discount factor. At each trading step $t$
(e.g., a day), the agent allocates its capital across the assets for step
$t+1$.

\textbf{State Space.}
At each time step $t$, the state $s_t \in S$ encodes all information available to the agent. For each asset $i \in \{1, ..., N\}$, we compute a vector of 11 temporal features from historical daily prices (open, high, low, close, adjusted-close), following the feature set of TradeMaster. We refer the reader to Appendix~\ref{app:features} for the full feature definitions.

\textbf{Action Space.}
The action $a_t \in \mathbb{R}^{N+1}$ is mapped via softmax to a portfolio weight vector $w_t = (w_{0,t}, w_{1,t}, \ldots, w_{N,t})$ on the simplex, where $w_{0,t}$ is the cash weight and $w_{i,t}$ ($i \geq 1$) is the fraction allocated to asset $i$. Short selling and leverage are precluded by construction since softmax outputs satisfy $w_{i,t} \geq 0$ and $\sum_i w_{i,t} = 1$. Cash earns zero return and provides an uninvested reserve. Because the action space is continuous, any actor-critic algorithm that outputs continuous actions can be used to learn the policy.

\textbf{State Transition.}
The transition function $\mathbb{P}$ is included in the MDP 5-tuple for completeness, but the agent does not learn a model of $\mathbb{P}$ during training: all five algorithms we consider are model-free and update their actor and critic directly from observed transitions.  Under the standard
price-taker assumption used in the RLFT literature
\citep{liu2022finrldeepreinforcementlearning,
trademaster/NEURIPS2023_b8f6f7f2}, the market-state transition is
independent of the agent's portfolio allocation, i.e., $\mathbb{P}(s_{t+1}\mid s_t,a_t) = \mathbb{P}(s_{t+1}\mid s_t).$ This approximation is reasonable when the agent's trades are small relative to available market liquidity, such that their price impact is negligible. It may break down for sufficiently large or aggressive trades, for which the agent's own trading activity can affect subsequent market prices. At inference time, $\fpilotabs$ uses a separately trained price forecaster $F$
to predict an $H$-step future market-state trajectory for imagined
rollouts during inference-time adaptation (Section~\ref{sec:methodology}). Under the
price-taker assumption, this does not require an action-conditioned
transition model.


\textbf{Reward Function.}
The training reward is the step-wise portfolio Profit and Loss (PnL).
We index each reward by the decision step that generates it. At time $t$,
after selecting the target portfolio $w_t$, the agent incurs a transaction
cost $\delta_t
=
c_{\mathrm{tx}} U_t
\left\|w_t-w_t^{-}\right\|_1,$ where $U_t$ is the portfolio value immediately before rebalancing and
$w_t^{-}$ denotes the post-price-movement, pre-rebalancing portfolio
weights. Given the realized next-period price relatives $x_{t+1}$,
portfolio value evolves as $U_{t+1} =
\left(U_t-\delta_t\right)
w_t^\top x_{t+1}.$
The reward associated with the decision at time $t$ is therefore $r_t = U_{t+1}-U_t$. The imagined rollouts in $\fpilotabs$ use the same portfolio accounting,
with realized price relatives replaced by forecast-derived ones; the full
multi-step formulation is given in Section~\ref{sec:methodology}.

The pre-trained agent maximizes expected discounted cumulative reward, \[\pi_\theta^*
=
\arg\max_{\pi_\theta}
\mathbb{E}_{\pi_\theta}
\left[
\sum_{k=0}^{T-t-1}
\gamma^k r_{t+k}
\right].
\]
No explicit volatility, drawdown, or downside-risk penalty is used during
RL training. Risk sensitivity is introduced only during inference-time
adaptation.

\section{$\fpilotabs$}
\label{sec:methodology}
We assume a reinforcement learning agent has been pre-trained on historical
market data using a model-free actor-critic algorithm with continuous action
outputs. We denote the actor parameters by $\theta$ and critic parameters by
$\psi$, with $\pi_\theta$ the actor and $\mathcal V_\psi$ the critic.
During training, the agent learns from observed market transitions with no
access to forecasted future prices.

\subsection{Inference-Time Policy Adaptation}
\label{sec:inference-timempc}
\begin{algorithm}[!t]
\caption{$\fpilotabs$: inference-time policy adaptation}
\label{alg:finpilot}
\begin{algorithmic}[1]
\Require Pre-trained actor $\pi_\theta$, frozen critic $\mathcal{V}_\psi$,
         forecaster $F$, horizon $H$, particles $K$, epochs $E$,
         step size $\alpha$, discount $\gamma$, risk weight $\lambda$,
         noise scale $\sigma$
\For{$t=1,\ldots,T$}
    \State Observe $s_t$, $w_t^{-}$, and portfolio value $U_t$
    \State $\{\hat{s}_{t+h}\}_{h=1}^H \gets F(s_{\leq t})$
    \State Sample and cache $K$ perturbed trajectories
    $\{\tilde{s}_{t:t+H}^{(k)}\}_{k=1}^K$
    \For{$e=1,\ldots,E$}
        \State Compute $\{J_t^{(k)}(\theta)\}_{k=1}^K$
        using~\eqref{eq:imagined-return}
        \State Compute the Monte Carlo estimate $\widehat{\mathcal J}_t$
        of~\eqref{eq:risk-objective}
        \State $\theta \gets
        \theta+\alpha\nabla_\theta\widehat{\mathcal J}_t$
    \EndFor
    \State Execute $w_t=\pi_\theta(s_t)$
\EndFor
\end{algorithmic}
\end{algorithm}

At inference time, we introduce an independently trained $H$-step forecaster
$F$, absent during RL training, that provides imagined future market
trajectories for policy adaptation. $\fpilotabs$ is forecaster-agnostic: $F$
may be deterministic or probabilistic, is trained only on data preceding the
evaluation period, and remains frozen during inference. Algorithm~\ref{alg:finpilot}
summarizes the inference-time adaptation procedure; a fully expanded version
is provided in Algorithm~\ref{alg:finpilot-full} in the appendix.
Figure~\ref{fig:architecture} illustrates the overall framework.

\paragraph{Forecast uncertainty.}
$\fpilotabs$ does not prescribe a particular representation of forecast
uncertainty. It only requires a collection of imagined future trajectories
over the planning horizon. A probabilistic forecaster can provide these
trajectories directly by sampling from its predictive distribution. For deterministic point forecasters, such as the XGBoost and
Theta + VAR(1)-ridge models we use, we construct $K$ forecast-error
scenarios around the predicted trajectory.

Specifically, for each forecast horizon $h$, we estimate a cross-asset
residual covariance matrix $\Sigma_h^{\mathrm{val}}$ from forecast errors
on the validation split and freeze it before test-time evaluation. At
decision time $t$, particle $k$ uses
$\varepsilon_{t,h}^{(k)}
\sim
\mathcal{N}
\left(
0,\,
\sigma^2\Sigma_h^{\mathrm{val}}
\right)$
to perturb the corresponding point forecast and form an imagined
trajectory
$\{\tilde{s}_{t+h}^{(k)}\}_{h=1}^{H}$.
The $K$ sampled trajectories are generated once at each decision step and
held fixed across all $E$ inference-time adaptation epochs. Gradients are
not propagated through either the forecaster or the trajectory-construction
procedure.


\paragraph{Closed-form imagined return.}
A useful property of portfolio management is that, conditional on a
market trajectory and portfolio allocations, the resulting portfolio
reward is available in closed form. Thus, $\fpilotabs$ does not require
a learned reward model: the forecaster supplies the imagined market
trajectory, while portfolio PnL follows deterministically from the
corresponding price relatives, allocations, and transaction costs. 

Consider a single imagined trajectory, omitting the particle index for
clarity. At the beginning of the rollout, $U_t$ and $w_t^{-}$ are the
current real portfolio value and post-price-movement, pre-rebalancing
portfolio weights, respectively. We set $\tilde{s}_t=s_t$. At imagined
step $t+h$, for $h=0,\ldots,H-1$, the actor produces the target allocation
$w_{t+h}=\pi_\theta(\tilde{s}_{t+h}).$
The transaction cost incurred when rebalancing to $w_{t+h}$ is $\delta_{t+h}
=
c_{\mathrm{tx}}U_{t+h}
\left\|w_{t+h}-w_{t+h}^{-}\right\|_1,$
where $U_{t+h}$ is the portfolio value immediately before rebalancing.
We apply $c_{\mathrm{tx}}=10$ bps to the full $\ell_1$ turnover over the
complete portfolio, including cash. Thus, a complete transfer from one
portfolio sleeve to another has turnover $2$ and incurs a 20 bp cost.

Given the forecast-derived next-period price relatives
$\tilde{x}_{t+h+1}$, imagined portfolio value evolves as
\[
U_{t+h+1}
=
\left(U_{t+h}-\delta_{t+h}\right)
w_{t+h}^{\top}\tilde{x}_{t+h+1}.
\]
Consistent with the real-environment reward defined in
Section~\ref{sec:problem_formulation}, we index the imagined reward by
the decision step that generates it: $\hat r_{t+h} = U_{t+h+1}-U_{t+h}.$

The resulting post-price-movement, pre-rebalancing portfolio weights for
the next imagined step are
\[
w_{t+h+1}^{-}
=
\frac{
w_{t+h}\odot\tilde{x}_{t+h+1}
}{
w_{t+h}^{\top}\tilde{x}_{t+h+1}
}.
\]

The imagined $H$-step return for one trajectory is therefore
\begin{equation}
\label{eq:imagined-return}
J_t(\theta)
=
\sum_{h=0}^{H-1}
\gamma^h \hat r_{t+h}
+
\gamma^H
\mathcal{V}_{\psi}(\tilde{s}_{t+H}),
\end{equation}
where $\mathcal{V}_{\psi}$ is the frozen critic used for the terminal
bootstrap. For SAC, TD3, and DDPG, which learn action-value critics, the
terminal value is
$\mathcal{V}_\psi(\tilde s) = Q_\psi(\tilde s, \pi_\theta(\tilde s))$,
taking the minimum over the twin critics for SAC and TD3.

\paragraph{Risk-aware objective.}
Because the imagined trajectory is stochastic under the forecast-error
model, the inference-time objective is defined over its induced
distribution. Let
    $\mu_t(\theta) = \mathbb{E}[J_t(\theta)]$
and define the downside variance $\mathcal{D}_t(\theta)
    =
    \mathbb{E}
    \left[
        \min
        \left(
            J_t(\theta)-\mu_t(\theta),\,0
        \right)^2
    \right].$
We optimize
\begin{equation}
\label{eq:risk-objective}
    \mathcal{J}_t(\theta)
    =
    \mu_t(\theta)
    -
    \lambda
    \sqrt{
        \mathcal{D}_t(\theta)+\epsilon
    }.
\end{equation}

In practice, we approximate these expectations using $K$ Monte Carlo forecast particles. For particle $k$, let
$J_t^{(k)} = J_t(\theta;\tilde{s}^{(k)})$. We compute
    $\bar{J}_t
    =
    \frac{1}{K}
    \sum_{k=1}^{K}
    J_t^{(k)},$
and
    $\widehat{\mathcal{D}}_t
    =
    \frac{1}{K}
    \sum_{k=1}^{K}
    \min
    \left(
        J_t^{(k)}-\bar{J}_t,\,0
    \right)^2.$ The resulting Monte Carlo objective is
    $\widehat{\mathcal{J}}_t = \bar{J}_t - \lambda \sqrt{\widehat{\mathcal{D}}_t+\epsilon
    }.$
In practice, the actor update uses Adam with gradient clipping, and both the
terminal value and the particle mean $\bar J_t$ inside the downside term are
treated as constants with respect to $\theta$. Consequently, the actor
gradient flows only through the closed-form imagined PnL; the terminal
bootstrap enters by shifting each particle's return, which determines which
particles fall in the downside term and how strongly they are weighted.

\section{Experimentation and Results}
\label{sec:experiments}

We apply $\fpilotabs$ to pre-trained policies obtained using
PPO~\citep{schulman2017proximalpolicyoptimizationalgorithms},
SAC~\citep{sac-pmlr-v80-haarnoja18b},
A2C~\citep{a2c-pmlr-v48-mniha16},
TD3~\citep{td3_pmlr-v80-fujimoto18a}, and
DDPG~\citep{ddpg_lillicrap2019continuouscontroldeepreinforcement},
five RL algorithms commonly used in RLFT benchmarks.
For each trained checkpoint, the frozen baseline (Base) and $\fpilotabs$ start
from identical parameters and are evaluated over the same test period, making
every comparison paired. Base keeps its parameters fixed throughout the test
period, whereas $\fpilotabs$ adapts the actor before each portfolio decision and
carries the adapted parameters forward to the next decision. The adaptation
uses a separately trained forecaster that is never observed by the policy
during RL training.

Our experiments address three questions.
\textbf{(Q1)} Does forecast-driven inference-time adaptation consistently
improve the corresponding frozen Base policy across algorithms, test years,
and two asset classes?
\textbf{(Q2)} Given the same forecast, how does adapting a pre-trained policy
compare with directly optimizing portfolio allocations?
\textbf{(Q3)} How do the gains from inference-time adaptation change as
forecast quality improves?

\subsection{Experimental Setup}
\label{sec:experimental-setup}

\begin{figure}[!t]
    \centering
    \includegraphics[width=0.9\linewidth]{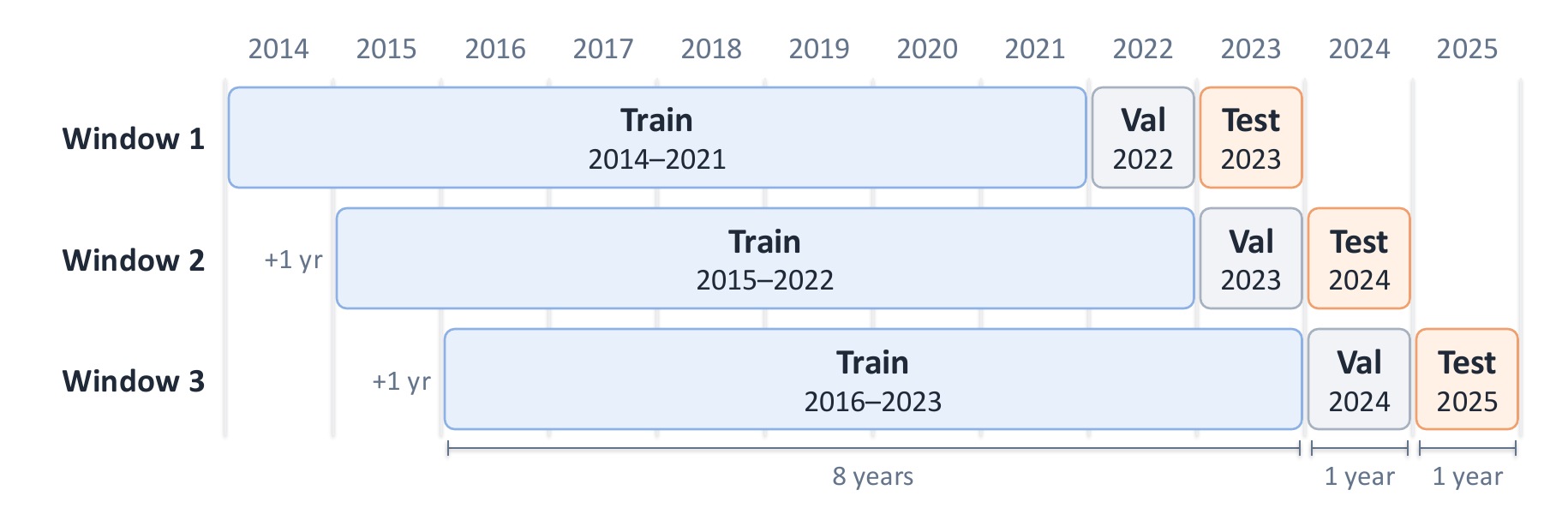}
    \caption{
    \textbf{Rolling evaluation protocol.}
    For each held-out test year $Y\in\{2023,2024,2025\}$, the preceding
    eight years are used for training and year $Y-1$ for validation.
    The complete train/validation/test window advances by one year at
    each evaluation step. RL policies and forecasters are refit using
    only the corresponding training period.
    }
    \label{fig:rolling-protocol}
\end{figure}

\paragraph{Markets, protocol, and agents.} 
We use the TradeMaster
portfolio-management environment~\citep{trademaster/NEURIPS2023_b8f6f7f2} with
two universes: 28 Dow Jones~30 constituents (DJ30) and 22 currency pairs quoted
against the U.S.\ dollar (FX). Both use the same rolling protocol
(Figure~\ref{fig:rolling-protocol}): for test year $Y\in\{2023,2024,2025\}$, all
models are trained on years $Y{-}9$ through $Y{-}2$, selected on $Y{-}1$, and
evaluated once on $Y$. All five agents are pre-trained separately for each
window, with batch size, hidden size, and learning rate selected on the
validation year (Appendix~\ref{app:pre-trained-setup}); no agent observes a
forecast during training.

\paragraph{Forecasters and $\fpilotabs$ configuration.}
To reflect that $\fpilotabs$ is
forecaster-agnostic, we use a different forecaster $F$ per market: a multi-horizon
XGBoost model of per-step price differences for DJ30, and an equal-weight
ensemble of a Theta model and a ridge-regularized VAR(1) for FX
(Theta + VAR(1)-ridge; $R^2$ in Table~\ref{tab:r2_forecaster}). Only the risk weight
$\lambda \in \{0.05, 0.10, 0.15\}$ is selected per market, algorithm, and
window, by validation Sortino ratio; all other inference-time hyperparameters
are shared across settings. The horizon $H = 50$ (also used by the
multi-period baseline) was fixed during preliminary development on DJ30 data
through 2021, before every validation and test year. Full settings are in
Appendix~\ref{app:finpilot-config}, and a post-hoc sensitivity analysis over step
size and adaptation epochs, including settings that outperform ours on some
metrics, is in Appendix~\ref{app:alpha}.

\paragraph{Baselines.}
Besides each agent's frozen policy (\emph{Base}), we compare against two
families of non-RL portfolios (Table~\ref{tab:nonrl}).
\emph{Passive}: buy-and-hold and daily-rebalanced $1/N$.
\emph{Forecast-driven optimization}: single-period ($H{=}1$) and multi-period
($H{=}50$) convex trading optimizers~\citep{boyd_cvx_DBLP:journals/ftopt/BoydBDKKNS17} that receive exactly
the same forecasts as $\fpilotabs$.

\paragraph{Evaluation metrics.}
Because the training reward captures only profit, we evaluate trained policies along five financial metrics standard in the RLFT literature~\citep{trademaster/NEURIPS2023_b8f6f7f2, deeptrader_Wang_Huang_Tu_Zhang_Xu_2021, SARL_Ye_Pei_Wang_Chen_Zhu_Xiao_Li_2020, finrl_meta_NEURIPS2022_0bf54b80}.
We report Total Return (TR), Sharpe ratio (SR), Sortino ratio
(SoR), Calmar ratio (CR), and Maximum Drawdown (MDD), following standard
RL-for-trading evaluation practice.
Formal definitions are provided in Appendix~\ref{app:metrics}.
Unless otherwise stated, aggregate results average the three per-window
seed means; complete per-window results with seed-level variability are
reported in Tables~\ref{tab:dj30_final} and~\ref{tab:fx_final} (Appendix~\ref{app:empirical}).

\subsection{Improvement over Frozen RL Policies (Q1)}
\label{sec:frozen-policy-results}

\begin{table}[t]
\centering
\caption{Performance averaged over the three rolling test windows (2023--2025).
Each entry is the mean of the per-window seed means ($n=5$ seeds per window).
Values after $\pm$ are seed standard deviations pooled over the three windows
(square root of the mean per-window seed variance), reflecting variability
across seeds within a window, not across test years. Base is the frozen
pre-trained policy; $\fpilotabs$ adapts its actor at inference time.
Per-window results are in Figure~\ref{fig:window-improvement} and
Tables~\ref{tab:dj30_final} and~\ref{tab:fx_final}. Better mean per algorithm
and metric in \textbf{bold}.}
\label{tab:main}   
\resizebox{\textwidth}{!}{{\footnotesize\setlength{\tabcolsep}{3pt}\renewcommand{\arraystretch}{1.1}\setlength{\aboverulesep}{0.25ex}\setlength{\belowrulesep}{0.35ex}
\begin{tabular}{ll ccccc ccccc}
\toprule
 & & \multicolumn{5}{c}{DJ30} & \multicolumn{5}{c}{FX} \\
\cmidrule(lr){3-7}\cmidrule(lr){8-12}
Algo & Method & Ret.\,(\%)$\uparrow$ & Sharpe$\uparrow$ & Sortino$\uparrow$ & Calmar$\uparrow$ & MDD\,(\%)$\downarrow$ & Ret.\,(\%)$\uparrow$ & Sharpe$\uparrow$ & Sortino$\uparrow$ & Calmar$\uparrow$ & MDD\,(\%)$\downarrow$ \\
\midrule
PPO & Base & 13.42\,{\tiny$\pm$0.29} & 1.10\,{\tiny$\pm$0.02} & 1.64\,{\tiny$\pm$0.03} & 1.48\,{\tiny$\pm$0.03} & \textbf{9.86}\,{\tiny$\pm$0.10} & 1.08\,{\tiny$\pm$0.18} & 0.15\,{\tiny$\pm$0.03} & 0.40\,{\tiny$\pm$0.05} & 0.73\,{\tiny$\pm$0.06} & 4.87\,{\tiny$\pm$0.14} \\
 & $\fpilotabs$ & \textbf{23.62}\,{\tiny$\pm$1.46} & \textbf{1.32}\,{\tiny$\pm$0.06} & \textbf{1.92}\,{\tiny$\pm$0.09} & \textbf{1.84}\,{\tiny$\pm$0.08} & 12.44\,{\tiny$\pm$0.24} & \textbf{1.29}\,{\tiny$\pm$0.73} & \textbf{0.28}\,{\tiny$\pm$0.15} & \textbf{0.77}\,{\tiny$\pm$0.26} & \textbf{0.94}\,{\tiny$\pm$0.51} & \textbf{4.06}\,{\tiny$\pm$0.44} \\
\cmidrule(lr){1-12}
SAC & Base & 13.77\,{\tiny$\pm$0.05} & 1.13\,{\tiny$\pm$0.00} & 1.68\,{\tiny$\pm$0.00} & 1.51\,{\tiny$\pm$0.00} & \textbf{9.79}\,{\tiny$\pm$0.01} & 1.20\,{\tiny$\pm$0.06} & 0.16\,{\tiny$\pm$0.01} & 0.42\,{\tiny$\pm$0.01} & 0.75\,{\tiny$\pm$0.01} & 4.80\,{\tiny$\pm$0.03} \\
 & $\fpilotabs$ & \textbf{15.95}\,{\tiny$\pm$0.04} & \textbf{1.22}\,{\tiny$\pm$0.00} & \textbf{1.89}\,{\tiny$\pm$0.00} & \textbf{1.73}\,{\tiny$\pm$0.00} & 9.89\,{\tiny$\pm$0.01} & \textbf{1.29}\,{\tiny$\pm$0.04} & \textbf{0.20}\,{\tiny$\pm$0.01} & \textbf{0.58}\,{\tiny$\pm$0.02} & \textbf{0.85}\,{\tiny$\pm$0.01} & \textbf{4.38}\,{\tiny$\pm$0.01} \\
\cmidrule(lr){1-12}
A2C & Base & 13.73\,{\tiny$\pm$0.10} & 1.12\,{\tiny$\pm$0.01} & 1.67\,{\tiny$\pm$0.01} & 1.50\,{\tiny$\pm$0.01} & \textbf{9.81}\,{\tiny$\pm$0.04} & 1.20\,{\tiny$\pm$0.02} & 0.16\,{\tiny$\pm$0.00} & 0.42\,{\tiny$\pm$0.01} & 0.75\,{\tiny$\pm$0.01} & 4.81\,{\tiny$\pm$0.01} \\
 & $\fpilotabs$ & \textbf{27.17}\,{\tiny$\pm$0.69} & \textbf{1.45}\,{\tiny$\pm$0.03} & \textbf{2.23}\,{\tiny$\pm$0.04} & \textbf{2.46}\,{\tiny$\pm$0.10} & 10.59\,{\tiny$\pm$0.14} & \textbf{1.44}\,{\tiny$\pm$0.15} & \textbf{0.33}\,{\tiny$\pm$0.03} & \textbf{0.85}\,{\tiny$\pm$0.05} & \textbf{0.79}\,{\tiny$\pm$0.03} & \textbf{3.92}\,{\tiny$\pm$0.11} \\
\cmidrule(lr){1-12}
TD3 & Base & 12.93\,{\tiny$\pm$4.81} & 0.99\,{\tiny$\pm$0.38} & 1.43\,{\tiny$\pm$0.46} & 1.25\,{\tiny$\pm$0.62} & \textbf{11.45}\,{\tiny$\pm$1.73} & 0.54\,{\tiny$\pm$1.11} & 0.09\,{\tiny$\pm$0.19} & 0.30\,{\tiny$\pm$0.26} & 0.72\,{\tiny$\pm$0.44} & 5.19\,{\tiny$\pm$0.83} \\
 & $\fpilotabs$ & \textbf{36.30}\,{\tiny$\pm$9.41} & \textbf{1.55}\,{\tiny$\pm$0.34} & \textbf{2.49}\,{\tiny$\pm$0.60} & \textbf{2.55}\,{\tiny$\pm$0.71} & 14.21\,{\tiny$\pm$3.84} & \textbf{3.42}\,{\tiny$\pm$1.44} & \textbf{0.53}\,{\tiny$\pm$0.25} & \textbf{1.06}\,{\tiny$\pm$0.37} & \textbf{1.21}\,{\tiny$\pm$0.20} & \textbf{4.47}\,{\tiny$\pm$1.09} \\
\cmidrule(lr){1-12}
DDPG & Base & 10.80\,{\tiny$\pm$4.30} & 0.85\,{\tiny$\pm$0.31} & 1.23\,{\tiny$\pm$0.45} & 1.08\,{\tiny$\pm$0.54} & \textbf{11.40}\,{\tiny$\pm$1.70} & $-$0.21\,{\tiny$\pm$1.66} & $-$0.11\,{\tiny$\pm$0.33} & $-$0.05\,{\tiny$\pm$0.56} & 0.28\,{\tiny$\pm$0.59} & 5.19\,{\tiny$\pm$1.05} \\
 & $\fpilotabs$ & \textbf{36.19}\,{\tiny$\pm$7.78} & \textbf{1.55}\,{\tiny$\pm$0.32} & \textbf{2.64}\,{\tiny$\pm$0.68} & \textbf{2.73}\,{\tiny$\pm$0.75} & 13.04\,{\tiny$\pm$2.18} & \textbf{1.48}\,{\tiny$\pm$1.83} & \textbf{0.19}\,{\tiny$\pm$0.39} & \textbf{0.51}\,{\tiny$\pm$0.57} & \textbf{0.33}\,{\tiny$\pm$0.33} & \textbf{5.00}\,{\tiny$\pm$1.04} \\
\bottomrule
\end{tabular}}}
\end{table}

We first ask whether inference-time adaptation improves the static
deployment of the same pre-trained RL policy.
Table~\ref{tab:main} reports performance averaged over the three
rolling test windows. On DJ30, $\fpilotabs$ increases the three-window mean Total Return, Sharpe,
Sortino, and Calmar ratios for all five pre-trained policies.
The improvement is not restricted to stochastic-policy methods:
TD3 and DDPG exhibit some of the largest return gains.
These improvements are accompanied by higher mean Maximum Drawdown on
DJ30, however, indicating that the larger realized returns do not
uniformly translate into lower peak-to-trough risk. On FX, $\fpilotabs$ again improves the three-window mean Return, Sharpe,
Sortino, and Calmar ratios for all five policies.
In contrast to DJ30, mean Maximum Drawdown also decreases for all five
policies, showing that the effect of inference-time adaptation on
drawdown is market dependent.

\begin{figure}[!t]
    \centering
    \includegraphics[width=0.83\linewidth]{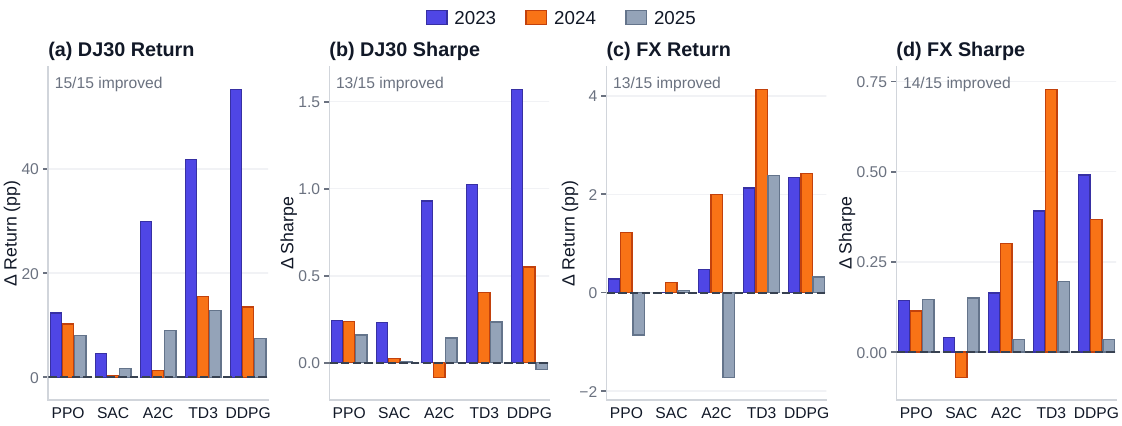}
    \caption{
    \textbf{Per-window improvement from $\fpilotabs$.}
    Bars show the change relative to the corresponding frozen Base policy
    for each RL policy and held-out test year.
    Positive values indicate improvement from inference-time adaptation.
    Colors denote test years.
    Across the 15 algorithm--window pairs in each market,
    Total Return improves in 15/15 DJ30 and 13/15 FX cases, while
    Sharpe ratio improves in 13/15 and 14/15 cases, respectively.
    }
    \label{fig:window-improvement}
\end{figure}

Because averaging across windows can obscure whether a gain is driven by a
single favorable period, Figure~\ref{fig:window-improvement} reports
per-window changes in Total Return and Sharpe ratio. Across the two markets,
$\fpilotabs$ improves Total Return in \textbf{28 of 30} and Sharpe ratio in
\textbf{27 of 30} policy--market--window comparisons, showing that the
aggregate gains are broadly distributed across evaluation periods.
Complete per-window results with seed-level variability are reported in
Appendix~\ref{app:empirical}.

\subsection{The same forecast, consumed differently (Q2)}
\label{sec:direct-optimization}

\noindent
\begin{minipage}[t]{0.43\linewidth}
\vspace{0pt}
Given the same forecast, one can bypass the pre-trained RL policy and
optimize allocations directly. We compare $\fpilotabs$ with single-period
optimization (SPO) and multi-period optimization (MPO), which optimize
the current allocation and an $H$-step allocation sequence, respectively.
$\fpilotabs$ instead optimizes the parameters of the pre-trained actor (Table~\ref{tab:method-comparison}).
\end{minipage}
\hfill
\begin{minipage}[t]{0.54\linewidth}
\centering

\captionof{table}{Comparison of inference-time forecast integration mechanisms.}
\label{tab:method-comparison}


\setlength{\tabcolsep}{3pt}
\resizebox{\linewidth}{!}{%
\begin{tabular}{lcccc}
\toprule
Method
& Pre-trained RL
& Forecast
& Optimization target
& Horizon \\
\midrule

Base RL
& \cmark
& \xmark
& None
& -- \\

SPO
& \xmark
& \cmark
& Current allocation
& $1$ \\

MPO
& \xmark
& \cmark
& Allocation sequence
& $H$ \\

$\fpilotabs$
& \cmark
& \cmark
& Policy parameters
& $H$ \\

\bottomrule
\end{tabular}%
}
\end{minipage}

\begin{table}[!t]

\caption{Comparison with non-RL baselines, averaged over the three rolling
test windows (2023--2025). $\fpilotabs$ entries are the mean of the
per-window seed means ($n=5$ seeds per window); all baselines are
deterministic. SPO and MPO receive the same forecasts as
$\fpilotabs$, whereas the passive allocators use no forecast. Best value per column within each market in \textbf{bold} (values tied at two decimals are all bolded). $^{\dagger}$On FX, the Boyd
optimizers remain essentially in cash in every window, so their near-zero
returns and drawdowns reflect inactivity rather than skill; they are
excluded when marking the best MDD, and their Sharpe and Sortino ratios,
computed from near-zero returns, are numerically unstable. Per-window
results are in Tables~\ref{tab:nonrl_dj30_per_year}
and~\ref{tab:nonrl_fx_per_year}.}
\label{tab:nonrl}
\begin{center}
\setlength{\tabcolsep}{5pt}
\renewcommand{\arraystretch}{1.08}
\resizebox{0.97\textwidth}{!}{%
\begin{tabular}{lcccc@{\hspace{14pt}}cccc}
\toprule
& \multicolumn{4}{c}{\textbf{DJ30}}
& \multicolumn{4}{c}{\textbf{FX}} \\
\cmidrule(lr){2-5}
\cmidrule(lr){6-9}
Method
& Return (\%) $\uparrow$ & Sharpe $\uparrow$ & Sortino $\uparrow$ & MDD (\%) $\downarrow$
& Return (\%) $\uparrow$ & Sharpe $\uparrow$ & Sortino $\uparrow$ & MDD (\%) $\downarrow$ \\
\midrule
Buy-and-hold
& $14.70$ & $1.18$ & $1.77$ & $10.02$
& $1.31$ & $0.17$ & $0.43$ & $5.01$ \\
$1/N$ (daily rebalanced)
& $14.25$ & $1.13$ & $1.68$ & $10.12$
& $1.26$ & $0.16$ & $0.42$ & $5.02$ \\
\midrule
\multicolumn{9}{l}{\emph{Direct optimization}} \\
Boyd SPO ($H{=}1$)
& $8.95$ & $0.52$ & $0.63$ & $17.62$
& $0.00^{\dagger}$ & $-0.71$ & $-0.80$ & $<\!0.01^{\dagger}$ \\
Boyd MPO ($H{=}50$)
& $6.25$ & $0.37$ & $0.49$ & $19.14$
& $0.00^{\dagger}$ & $-2.98$ & $-3.51$ & $<\!0.01^{\dagger}$ \\
\midrule
\multicolumn{9}{l}{\emph{Policy optimization}} \\
FinPILOT-PPO
& $23.62$ & $1.32$ & $1.92$ & $12.44$
& $1.29$ & $0.28$ & $0.77$ & $4.06$ \\
FinPILOT-SAC
& $15.95$ & $1.22$ & $1.89$ & {\boldmath$9.89$}
& $1.29$ & $0.20$ & $0.58$ & $4.38$ \\
FinPILOT-A2C
& $27.17$ & $1.45$ & $2.23$ & $10.59$
& $1.44$ & $0.33$ & $0.85$ & {\boldmath$3.92$} \\
FinPILOT-TD3
& {\boldmath$36.30$} & {\boldmath$1.55$} & $2.49$ & $14.21$
& {\boldmath$3.42$} & {\boldmath$0.53$} & {\boldmath$1.06$} & $4.47$ \\
FinPILOT-DDPG
& $36.19$ & {\boldmath$1.55$} & {\boldmath$2.64$} & $13.04$
& $1.48$ & $0.19$ & $0.51$ & $5.00$ \\
\bottomrule
\end{tabular}%
}
\end{center}
\end{table}

Under this evaluation protocol, directly optimizing allocations from the
forecast does not reproduce the aggregate performance obtained by adapting
the pre-trained policies (Table~\ref{tab:nonrl}).
On DJ30, every $\fpilotabs$-adapted policy achieves higher mean Total Return,
Sharpe ratio, and Sortino ratio than both SPO and MPO.
On FX, SPO and MPO select essentially cash-only portfolios and
therefore obtain near-zero mean returns, whereas the adapted RL policies
remain exposed to the market and achieve positive three-window mean returns. These results should not be interpreted as a general claim that policy
optimization dominates direct portfolio optimization.
Rather, they show that access to the forecast alone is not sufficient to
reproduce the gains of $\fpilotabs$ under our evaluation protocol:
the mechanism by which the forecast is incorporated into the decision
rule matters.

\subsection{How much forecast skill is needed (Q3)}
\label{sec:cheating}

A natural question is how much further the framework could go with a stronger forecaster. To answer this, we construct controlled forecasts at calibrated $R^2$ levels.


\paragraph{Controlled forecast-skill condition.}
To vary forecast quality while preserving the horizon-wise error pattern of
the forecasters used in our main experiments, we rescale their errors. For
origin $t$ and asset $i$, let $\hat d_{t,i,h}$ denote the original forecast
and $d_{t,i,h}$ the realized one-day price change at step $t+h$,
$h=1,\dots,H$. We set
$\tilde d_{t,i,h} = d_{t,i,h} + c_{t,i}(\hat d_{t,i,h}-d_{t,i,h})$ with
$c_{t,i}=\sqrt{(1-R^2_\star)/(1-R^2_{t,i})}$, where $R^2_{t,i}$ is the
original forecast's $R^2$ over the $H$-step horizon relative to a
trailing-mean baseline (Appendix~\ref{app:r2-def}). This scales the squared
error by $c_{t,i}^2$ without changing the baseline, so each controlled
forecast attains exactly $R^2_\star$ at every origin and asset. Targets
below the original skill ($c_{t,i}>1$) amplify the original errors, while
targets above it ($c_{t,i}<1$) move the forecast toward the realized path,
with $c_{t,i}=1$ and $c_{t,i}=0$ recovering the original forecast and the
realized path, respectively. Because it uses realized future prices, this
construction is an oracle diagnostic for mapping forecast skill to planning
performance, not a deployable forecaster.

\begin{figure}[!t]
    \centering
    \includegraphics[width=0.80\linewidth]{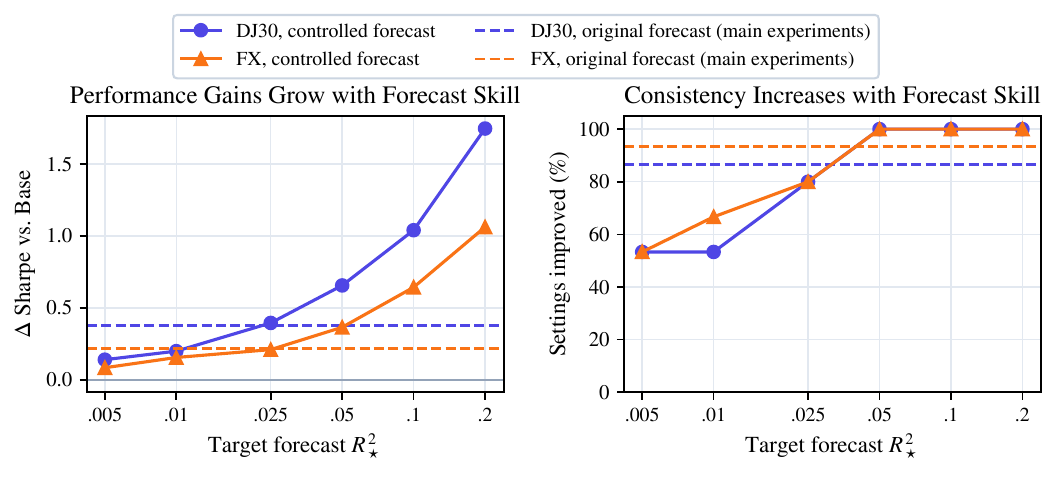}
    \caption{
    \textbf{Performance scales with controlled forecast skill.}
    Left: mean Sharpe improvement over the corresponding frozen Base policy
    as target forecast quality $R^2_\star$ increases.
    Right: percentage of the 15 policy--window settings per market in which
    $\fpilotabs$ improves Sharpe.
    Solid curves denote controlled forecasts, while dashed horizontal lines
    show the corresponding results obtained with the original forecasts
    used in the main experiments.
    Controlled forecasts use realized future prices by construction and are
    therefore diagnostic rather than deployable.
    }
    \label{fig:forecast-skill}
\end{figure}

\paragraph{Scaling with forecast skill.}
Figure~\ref{fig:forecast-skill} shows that stronger predictive signals
produce both larger and more consistent gains from inference-time
adaptation. In the weakest controlled-signal regime, adaptation is not
uniformly beneficial across policies and market windows. As
$R^2_\star$ increases, however, the mean Sharpe improvement rises in both
markets and the fraction of improved settings increases correspondingly.
By $R^2_\star=0.05$, all 15 policy--window comparisons in each market
improve over Base in Sharpe, and this remains true for the stronger
controlled forecasts shown in Figure~\ref{fig:forecast-skill}. We do not read $R^2_\star=0.05$ as a practical threshold: such skill is hard
to attain in financial prediction, the construction is oracle-based, and
forecasts with similar $R^2$ can differ in cross-asset, temporal, and
directional error structure.

\paragraph{Original vs.\ controlled forecasts.}
The original forecasts operate in the low-skill region of the controlled
sweep. Their mean Sharpe improvements are close to those obtained with
controlled forecasts at $R^2_\star=0.025$: $+0.38$ versus $+0.39$ on
DJ30 and $+0.22$ versus $+0.21$ on FX. Yet with the original forecasts,
$\fpilotabs$ improves Sharpe more consistently, in 13/15 and 14/15
policy--window comparisons on DJ30 and FX, respectively, compared with
12/15 for the controlled $R^2_\star=0.025$ condition in each market. This is consistent with the controlled construction enforcing the same $R^2_\star$ at every origin and asset, whereas the original forecaster's skill
varies, and suggests that aggregate $R^2$ alone does not fully characterize
decision value.

Taken together, these results answer Q3: $\fpilotabs$ already extracts useful
decision signals from the low-skill forecasters used in the main
experiments, while stronger controlled predictive signals produce larger
and more consistent gains. Because the policy and adaptation procedure are
held fixed throughout this sweep, the result supports the interpretation
that the improvements arise from predictive information in the forecast
rather than simply from performing additional gradient updates at inference
time.
\section{Conclusion}

We introduced $\fpilotabs$, an inference-time policy adaptation framework that
uses an independently trained financial forecaster to adapt a pre-trained RL
trading policy before each decision. Across five pre-trained policies, two
markets, and three rolling held-out test years, $\fpilotabs$ improves Total
Return in 28 of 30 policy--market--window comparisons and Sharpe ratio in
27 of 30. Same-forecast direct portfolio optimizers do not reproduce these
aggregate gains, while controlled forecast-skill experiments show that the
magnitude and consistency of the benefit increase with predictive signal
quality. These results support inference-time policy adaptation as a modular
interface between separately trained forecasters and pre-trained RL trading
policies.
\section*{Limitations and Future Work}

Our study is limited to historical backtests on DJ30 and FX under the
price-taker assumption, which neglects the market impact of large or
aggressive trades. The formulation also assumes long-only, unlevered, daily
portfolios. Our evaluation covers only three
recent test years (2023--2025), so broader validation across assets, market
regimes, frequencies, and portfolio constraints is needed. The benefit of adaptation
depends on its magnitude: we fix the step size and epoch count across all
settings, but larger updates help when the forecast is informative and hurt
when it is weak (e.g., DJ30 2024; Appendix~\ref{app:alpha}), so no single
setting is best in every market and year. The controlled forecast-skill
experiments use realized future prices and are diagnostics, not deployable
results. Future work could extend $\fpilotabs$ to market-impact-aware
environments, richer portfolio constraints and risk objectives, update
magnitudes that adapt to estimated forecast skill, and probabilistic
forecasters.

\bibliographystyle{abbrvnat}
\bibliography{arxiv_v2/references}

@inproceedings{
Hafner2020Dream,
title={Dream to Control: Learning Behaviors by Latent Imagination},
author={Danijar Hafner and Timothy Lillicrap and Jimmy Ba and Mohammad Norouzi},
booktitle={International Conference on Learning Representations},
year={2020},
url={https://openreview.net/forum?id=S1lOTC4tDS}
}

@misc{feng2019enhancingstockmovementprediction,
      title={Enhancing Stock Movement Prediction with Adversarial Training}, 
      author={Fuli Feng and Huimin Chen and Xiangnan He and Ji Ding and Maosong Sun and Tat-Seng Chua},
      year={2019},
      eprint={1810.09936},
      archivePrefix={arXiv},
      primaryClass={q-fin.TR},
      url={https://arxiv.org/abs/1810.09936}, 
}

@article{mpc-mayne,
title = {Constrained model predictive control: Stability and optimality},
journal = {Automatica},
volume = {36},
number = {6},
pages = {789-814},
year = {2000},
issn = {0005-1098},
doi = {https://doi.org/10.1016/S0005-1098(99)00214-9},
url = {https://www.sciencedirect.com/science/article/pii/S0005109899002149},
author = {D.Q. Mayne and J.B. Rawlings and C.V. Rao and P.O.M. Scokaert}
}

@article{campbell2008predicting,
    author = {Campbell, John Y. and Thompson, Samuel B.},
    title = {Predicting Excess Stock Returns Out of Sample: Can Anything Beat the Historical Average?},
    journal = {The Review of Financial Studies},
    volume = {21},
    number = {4},
    pages = {1509-1531},
    year = {2008},
    month = {07},
    issn = {0893-9454},
    doi = {10.1093/rfs/hhm055},
    url = {https://doi.org/10.1093/rfs/hhm055},
    eprint = {https://academic.oup.com/rfs/article-pdf/21/4/1509/24453399/hhm055.pdf},
}

@article{mpc-garcia,
title = {Model predictive control: Theory and practice—A survey},
journal = {Automatica},
volume = {25},
number = {3},
pages = {335-348},
year = {1989},
issn = {0005-1098},
doi = {https://doi.org/10.1016/0005-1098(89)90002-2},
url = {https://www.sciencedirect.com/science/article/pii/0005109889900022},
author = {Carlos E. García and David M. Prett and Manfred Morari}
}

@inproceedings{yoo2021accurate,
author = {Yoo, Jaemin and Soun, Yejun and Park, Yong-chan and Kang, U},
title = {Accurate Multivariate Stock Movement Prediction via Data-Axis Transformer with Multi-Level Contexts},
year = {2021},
isbn = {9781450383325},
publisher = {Association for Computing Machinery},
address = {New York, NY, USA},
url = {https://doi.org/10.1145/3447548.3467297},
doi = {10.1145/3447548.3467297},
booktitle = {Proceedings of the 27th ACM SIGKDD Conference on Knowledge Discovery \& Data Mining},
pages = {2037–2045},
numpages = {9},
location = {Virtual Event, Singapore},
series = {KDD '21}
}

@article{sharpe1994sharpe,
  author  = {Sharpe, William F.},
  title   = {The {Sharpe} Ratio},
  journal = {Journal of Portfolio Management},
  volume  = {21},
  number  = {1},
  pages   = {49--58},
  year    = {1994}
}

@InProceedings{sac-pmlr-v80-haarnoja18b,
  title = 	 {Soft Actor-Critic: Off-Policy Maximum Entropy Deep Reinforcement Learning with a Stochastic Actor},
  author =       {Haarnoja, Tuomas and Zhou, Aurick and Abbeel, Pieter and Levine, Sergey},
  booktitle = 	 {Proceedings of the 35th International Conference on Machine Learning},
  pages = 	 {1861--1870},
  year = 	 {2018},
  editor = 	 {Dy, Jennifer and Krause, Andreas},
  volume = 	 {80},
  series = 	 {Proceedings of Machine Learning Research},
  month = 	 {10--15 Jul},
  publisher =    {PMLR},
  url = 	 {https://proceedings.mlr.press/v80/haarnoja18b.html}
}

@InProceedings{a2c-pmlr-v48-mniha16,
  title = 	 {Asynchronous Methods for Deep Reinforcement Learning},
  author = 	 {Mnih, Volodymyr and Badia, Adria Puigdomenech and Mirza, Mehdi and Graves, Alex and Lillicrap, Timothy and Harley, Tim and Silver, David and Kavukcuoglu, Koray},
  booktitle = 	 {Proceedings of The 33rd International Conference on Machine Learning},
  pages = 	 {1928--1937},
  year = 	 {2016},
  editor = 	 {Balcan, Maria Florina and Weinberger, Kilian Q.},
  volume = 	 {48},
  series = 	 {Proceedings of Machine Learning Research},
  address = 	 {New York, New York, USA},
  month = 	 {20--22 Jun},
  publisher =    {PMLR},
  url = 	 {https://proceedings.mlr.press/v48/mniha16.html}
}

@inproceedings{earnmore_10.1145/3589334.3645615,
author = {Zhang, Wentao and Zhao, Yilei and Sun, Shuo and Ying, Jie and Xie, Yonggang and Song, Zitao and Wang, Xinrun and An, Bo},
title = {Reinforcement Learning with Maskable Stock Representation for Portfolio Management in Customizable Stock Pools},
year = {2024},
isbn = {9798400701719},
publisher = {Association for Computing Machinery},
address = {New York, NY, USA},
url = {https://doi.org/10.1145/3589334.3645615},
doi = {10.1145/3589334.3645615},
booktitle = {Proceedings of the ACM Web Conference 2024},
pages = {187–198},
numpages = {12},
location = {Singapore, Singapore},
series = {WWW '24}
}

@Article{Schrittwieser2020MuZero,
author={Schrittwieser, Julian
and Antonoglou, Ioannis
and Hubert, Thomas
and Simonyan, Karen
and Sifre, Laurent
and Schmitt, Simon
and Guez, Arthur
and Lockhart, Edward
and Hassabis, Demis
and Graepel, Thore
and Lillicrap, Timothy
and Silver, David},
title={Mastering Atari, Go, chess and shogi by planning with a learned model},
journal={Nature},
year={2020},
month={Dec},
day={01},
volume={588},
number={7839},
pages={604-609},
issn={1476-4687},
doi={10.1038/s41586-020-03051-4},
url={https://doi.org/10.1038/s41586-020-03051-4}
}

@Article{Hafner2025DreamerV3,
author={Hafner, Danijar
and Pasukonis, Jurgis
and Ba, Jimmy
and Lillicrap, Timothy},
title={Mastering diverse control tasks through world models},
journal={Nature},
year={2025},
month={Apr},
day={01},
volume={640},
number={8059},
pages={647-653},
issn={1476-4687},
doi={10.1038/s41586-025-08744-2},
url={https://doi.org/10.1038/s41586-025-08744-2}
}

@inproceedings{trademaster/NEURIPS2023_b8f6f7f2,
 author = {Sun, Shuo and Qin, Molei and Zhang, Wentao and Xia, Haochong and Zong, Chuqiao and Ying, Jie and Xie, Yonggang and Zhao, Lingxuan and Wang, Xinrun and An, Bo},
 booktitle = {Advances in Neural Information Processing Systems},
 editor = {A. Oh and T. Naumann and A. Globerson and K. Saenko and M. Hardt and S. Levine},
 pages = {59047--59061},
 publisher = {Curran Associates, Inc.},
 title = {TradeMaster: A Holistic Quantitative Trading Platform Empowered by Reinforcement Learning},
 url = {https://proceedings.neurips.cc/paper_files/paper/2023/file/b8f6f7f2ba4137124ac976286eacb611-Paper-Datasets_and_Benchmarks.pdf},
 volume = {36},
 year = {2023}
}

@inproceedings{finrl_meta_NEURIPS2022_0bf54b80,
 author = {Liu, Xiao-Yang and Xia, Ziyi and Rui, Jingyang and Gao, Jiechao and Yang, Hongyang and Zhu, Ming and Wang, Christina and Wang, Zhaoran and Guo, Jian},
 booktitle = {Advances in Neural Information Processing Systems},
 editor = {S. Koyejo and S. Mohamed and A. Agarwal and D. Belgrave and K. Cho and A. Oh},
 pages = {1835--1849},
 publisher = {Curran Associates, Inc.},
 title = {FinRL-Meta: Market Environments and Benchmarks for Data-Driven Financial Reinforcement Learning},
 url = {https://proceedings.neurips.cc/paper_files/paper/2022/file/0bf54b80686d2c4dc0808c2e98d430f7-Paper-Datasets_and_Benchmarks.pdf},
 volume = {35},
 year = {2022}
}

@misc{liu2022finrldeepreinforcementlearning,
      title={FinRL: A Deep Reinforcement Learning Library for Automated Stock Trading in Quantitative Finance}, 
      author={Xiao-Yang Liu and Hongyang Yang and Qian Chen and Runjia Zhang and Liuqing Yang and Bowen Xiao and Christina Dan Wang},
      year={2022},
      eprint={2011.09607},
      archivePrefix={arXiv},
      primaryClass={q-fin.TR},
      url={https://arxiv.org/abs/2011.09607}, 
}

@misc{deb2026modelpredictivecontroldifferentiable,
      title={Inference Time Policy Optimization for Offline RL with Differentiable World Models}, 
      author={Rohan Deb and Stephen J. Wright and Arindam Banerjee},
      year={2026},
      eprint={2603.22430},
      archivePrefix={arXiv},
      primaryClass={cs.LG},
      url={https://arxiv.org/abs/2603.22430}, 
}

@article{boyd_cvx_DBLP:journals/ftopt/BoydBDKKNS17,
  author       = {Stephen P. Boyd and
                  Enzo Busseti and
                  Steven Diamond and
                  Ronald N. Kahn and
                  Kwangmoo Koh and
                  Peter Nystrup and
                  Jan Speth},
  title        = {Multi-Period Trading via Convex Optimization},
  journal      = {Found. Trends Optim.},
  volume       = {3},
  number       = {1},
  pages        = {1--76},
  year         = {2017},
  url          = {https://doi.org/10.1561/2400000023},
  doi          = {10.1561/2400000023},
  bibsource    = {dblp computer science bibliography, https://dblp.org}
}

@article{SARL_Ye_Pei_Wang_Chen_Zhu_Xiao_Li_2020, title={Reinforcement-Learning Based Portfolio Management with Augmented Asset Movement Prediction States}, volume={34}, url={https://ojs.aaai.org/index.php/AAAI/article/view/5462}, DOI={10.1609/aaai.v34i01.5462}, abstractNote={&lt;p&gt;Portfolio management (PM) is a fundamental financial planning task that aims to achieve investment goals such as maximal profits or minimal risks. Its decision process involves continuous derivation of valuable information from various data sources and sequential decision optimization, which is a prospective research direction for reinforcement learning (RL). In this paper, we propose SARL, a novel State-Augmented RL framework for PM. Our framework aims to address two unique challenges in financial PM: (1) &lt;em&gt;data heterogeneity&lt;/em&gt; – the collected information for each asset is usually diverse, noisy and imbalanced (e.g., news articles); and (2) &lt;em&gt;environment uncertainty&lt;/em&gt; – the financial market is versatile and non-stationary. To incorporate heterogeneous data and enhance robustness against environment uncertainty, our SARL augments the asset information with their price movement prediction as additional states, where the prediction can be solely based on financial data (e.g., asset prices) or derived from alternative sources such as news. Experiments on two real-world datasets, (i) Bitcoin market and (ii) HighTech stock market with 7-year Reuters news articles, validate the effectiveness of SARL over existing PM approaches, both in terms of accumulated profits and risk-adjusted profits. Moreover, extensive simulations are conducted to demonstrate the importance of our proposed state augmentation, providing new insights and boosting performance significantly over standard RL-based PM method and other baselines.&lt;/p&gt;}, number={01}, journal={Proceedings of the AAAI Conference on Artificial Intelligence}, author={Ye, Yunan and Pei, Hengzhi and Wang, Boxin and Chen, Pin-Yu and Zhu, Yada and Xiao, Ju and Li, Bo}, year={2020}, month={Apr.}, pages={1112-1119} }

@misc{jiang2017deepreinforcementlearningframework,
      title={A Deep Reinforcement Learning Framework for the Financial Portfolio Management Problem}, 
      author={Zhengyao Jiang and Dixing Xu and Jinjun Liang},
      year={2017},
      eprint={1706.10059},
      archivePrefix={arXiv},
      primaryClass={q-fin.CP},
      url={https://arxiv.org/abs/1706.10059}, 
}

@misc{schulman2017proximalpolicyoptimizationalgorithms,
      title={Proximal Policy Optimization Algorithms}, 
      author={John Schulman and Filip Wolski and Prafulla Dhariwal and Alec Radford and Oleg Klimov},
      year={2017},
      eprint={1707.06347},
      archivePrefix={arXiv},
      primaryClass={cs.LG},
      url={https://arxiv.org/abs/1707.06347}, 
}

@InProceedings{td3_pmlr-v80-fujimoto18a,
  title = 	 {Addressing Function Approximation Error in Actor-Critic Methods},
  author =       {Fujimoto, Scott and van Hoof, Herke and Meger, David},
  booktitle = 	 {Proceedings of the 35th International Conference on Machine Learning},
  pages = 	 {1587--1596},
  year = 	 {2018},
  editor = 	 {Dy, Jennifer and Krause, Andreas},
  volume = 	 {80},
  series = 	 {Proceedings of Machine Learning Research},
  month = 	 {10--15 Jul},
  publisher =    {PMLR},
  url = 	 {https://proceedings.mlr.press/v80/fujimoto18a.html}
}

@misc{ddpg_lillicrap2019continuouscontroldeepreinforcement,
      title={Continuous control with deep reinforcement learning}, 
      author={Timothy P. Lillicrap and Jonathan J. Hunt and Alexander Pritzel and Nicolas Heess and Tom Erez and Yuval Tassa and David Silver and Daan Wierstra},
      year={2019},
      eprint={1509.02971},
      archivePrefix={arXiv},
      primaryClass={cs.LG},
      url={https://arxiv.org/abs/1509.02971}, 
}

@inproceedings{MOPO_NEURIPS2020_a322852c,
 author = {Yu, Tianhe and Thomas, Garrett and Yu, Lantao and Ermon, Stefano and Zou, James Y and Levine, Sergey and Finn, Chelsea and Ma, Tengyu},
 booktitle = {Advances in Neural Information Processing Systems},
 editor = {H. Larochelle and M. Ranzato and R. Hadsell and M.F. Balcan and H. Lin},
 pages = {14129--14142},
 publisher = {Curran Associates, Inc.},
 title = {MOPO: Model-based Offline Policy Optimization},
 url = {https://proceedings.neurips.cc/paper_files/paper/2020/file/a322852ce0df73e204b7e67cbbef0d0a-Paper.pdf},
 volume = {33},
 year = {2020}
}

@inproceedings{MBPO_NEURIPS2019_5faf461e,
 author = {Janner, Michael and Fu, Justin and Zhang, Marvin and Levine, Sergey},
 booktitle = {Advances in Neural Information Processing Systems},
 editor = {H. Wallach and H. Larochelle and A. Beygelzimer and F. d\textquotesingle Alch\'{e}-Buc and E. Fox and R. Garnett},
 pages = {},
 publisher = {Curran Associates, Inc.},
 title = {When to Trust Your Model: Model-Based Policy Optimization},
 url = {https://proceedings.neurips.cc/paper_files/paper/2019/file/5faf461eff3099671ad63c6f3f094f7f-Paper.pdf},
 volume = {32},
 year = {2019}
}

@misc{decision_diffuser_ajay2023conditionalgenerativemodelingneed,
      title={Is Conditional Generative Modeling all you need for Decision-Making?}, 
      author={Anurag Ajay and Yilun Du and Abhi Gupta and Joshua Tenenbaum and Tommi Jaakkola and Pulkit Agrawal},
      year={2023},
      eprint={2211.15657},
      archivePrefix={arXiv},
      primaryClass={cs.LG},
      url={https://arxiv.org/abs/2211.15657}, 
}

@inproceedings{trajectory_janner2021sequence,
 author = {Janner, Michael and Li, Qiyang and Levine, Sergey},
 booktitle = {Advances in Neural Information Processing Systems},
 editor = {M. Ranzato and A. Beygelzimer and Y. Dauphin and P.S. Liang and J. Wortman Vaughan},
 pages = {1273--1286},
 publisher = {Curran Associates, Inc.},
 title = {Offline Reinforcement Learning as One Big Sequence Modeling Problem},
 url = {https://proceedings.neurips.cc/paper_files/paper/2021/file/099fe6b0b444c23836c4a5d07346082b-Paper.pdf},
 volume = {34},
 year = {2021}
}

@article{deeptrader_Wang_Huang_Tu_Zhang_Xu_2021, title={DeepTrader: A Deep Reinforcement Learning Approach for Risk-Return Balanced Portfolio Management with Market Conditions Embedding}, volume={35}, url={https://ojs.aaai.org/index.php/AAAI/article/view/16144}, DOI={10.1609/aaai.v35i1.16144}, abstractNote={Most existing reinforcement learning (RL)-based portfolio management models do not take into account the market conditions, which limits their performance in risk-return balancing. In this paper, we propose DeepTrader, a deep RL method to optimize the investment policy. In particular, to tackle the risk-return balancing problem, our model embeds macro market conditions as an indicator to dynamically adjust the proportion between long and short funds, to lower the risk of market fluctuations, with the negative maximum drawdown as the reward function. Additionally, the model involves a unit to evaluate individual assets, which learns dynamic patterns from historical data with the price rising rate as the reward function. Both temporal and spatial dependencies between assets are captured hierarchically by a specific type of graph structure. Particularly, we find that the estimated causal structure best captures the interrelationships between assets, compared to industry classification and correlation. The two units are complementary and integrated to generate a suitable portfolio which fits the market trend well and strikes a balance between return and risk effectively. Experiments on three well-known stock indexes demonstrate the superiority of DeepTrader in terms of risk-gain criteria.}, number={1}, journal={Proceedings of the AAAI Conference on Artificial Intelligence}, author={Wang, Zhicheng and Huang, Biwei and Tu, Shikui and Zhang, Kun and Xu, Lei}, year={2021}, month={May}, pages={643-650} }

@inproceedings{deepscalper_10.1145/3511808.3557283,
author = {Sun, Shuo and Xue, Wanqi and Wang, Rundong and He, Xu and Zhu, Junlei and Li, Jian and An, Bo},
title = {DeepScalper: A Risk-Aware Reinforcement Learning Framework to Capture Fleeting Intraday Trading Opportunities},
year = {2022},
isbn = {9781450392365},
publisher = {Association for Computing Machinery},
address = {New York, NY, USA},
url = {https://doi.org/10.1145/3511808.3557283},
doi = {10.1145/3511808.3557283},
booktitle = {Proceedings of the 31st ACM International Conference on Information \& Knowledge Management},
pages = {1858–1867},
numpages = {10},
location = {Atlanta, GA, USA},
series = {CIKM '22}
}

@article{gu_kelly_r2_10.1093/rfs/hhaa009,
    author = {Gu, Shihao and Kelly, Bryan and Xiu, Dacheng},
    title = {Empirical Asset Pricing via Machine Learning},
    journal = {The Review of Financial Studies},
    volume = {33},
    number = {5},
    pages = {2223-2273},
    year = {2020},
    month = {02},
    issn = {0893-9454},
    doi = {10.1093/rfs/hhaa009},
    url = {https://doi.org/10.1093/rfs/hhaa009},
    eprint = {https://academic.oup.com/rfs/article-pdf/33/5/2223/33209812/hhaa009.pdf},
}

@inproceedings{
hansen2024tdmpc,
title={{TD}-{MPC}2: Scalable, Robust World Models for Continuous Control},
author={Nicklas Hansen and Hao Su and Xiaolong Wang},
booktitle={The Twelfth International Conference on Learning Representations},
year={2024},
url={https://openreview.net/forum?id=Oxh5CstDJU}
}

@inproceedings{
deb2026model,
title={Model Predictive Control with Differentiable World Models for Offline Reinforcement Learning},
author={Rohan Deb and Stephen J. Wright and Arindam Banerjee},
booktitle={ICLR 2026 the 2nd Workshop on World Models: Understanding, Modelling and Scaling},
year={2026},
url={https://openreview.net/forum?id=KqKyK9lLZI}
}

\appendix
\section{Related Work}
\label{app:related}
\textbf{Reinforcement Learning for Financial Trading.} The application of RL to portfolio management has grown substantially, with early tabular methods giving way to deep RL approaches. \citet{jiang2017deepreinforcementlearningframework} made the first deep RL attempt on portfolio management, and subsequent work introduced risk-aware objectives and augmented state representations incorporating price prediction signals \citep{deeptrader_Wang_Huang_Tu_Zhang_Xu_2021,deepscalper_10.1145/3511808.3557283,SARL_Ye_Pei_Wang_Chen_Zhu_Xiao_Li_2020}. \citet{trademaster/NEURIPS2023_b8f6f7f2} and \citet{liu2022finrldeepreinforcementlearning} consolidated these into reproducible benchmarking platforms, which we build on directly. A consistent limitation across all of these methods is that the trained policy is deployed statically; there is no mechanism for exploiting a price forecaster at inference time within a principled planning framework. Our work addresses this gap.

\textbf{Model Predictive Control and World Models.} Model-based RL methods such as MBPO and MOPO use learned world models during training to generate imagined rollouts for policy improvement \citep{MBPO_NEURIPS2019_5faf461e,MOPO_NEURIPS2020_a322852c}.  A separate line of work performs inference-time planning: Trajectory Transformer \citep{trajectory_janner2021sequence}, Decision Diffuser \citep{decision_diffuser_ajay2023conditionalgenerativemodelingneed}, and TD-MPC2 \citep{hansen2024tdmpc} sample candidate plans at test time but do not backpropagate through the rollout to adapt policy parameters. The most directly relevant work is \citet{deb2026modelpredictivecontroldifferentiable,deb2026model}, which introduced inference-time gradient-based policy adaptation through a learned differentiable world model on offline RL benchmarks. We extend this paradigm to financial trading. As we argue in Section \ref{sec:intro}, the structure of portfolio management simplifies the world modeling requirements substantially, reducing the problem to price forecasting alone.

\textbf{Model Predictive Control (MPC) in Finance.} \citet{boyd_cvx_DBLP:journals/ftopt/BoydBDKKNS17} established convex optimization-based multi-period trading as the classical MPC framework in quantitative finance, optimizing action sequences over predicted future returns subject to transaction cost penalties. Our work connects this tradition to learned RL policies, replacing the convex planner with gradient-based adaptation of a pre-trained actor.

\textbf{Price Forecasting as a Trading Signal.} Practical price movement
prediction for equities is a well-studied but fundamentally difficult problem.
Even state-of-the-art machine learning methods achieve modest out-of-sample
$R^2$ values for short-horizon equity prediction
\citep{gu_kelly_r2_10.1093/rfs/hhaa009}. Our XGBoost forecaster for DJ30 predicts daily price movements rather than
returns directly. On daily close-price moves over the $H=50$ horizon,
averaged across test years, it achieves a pooled out-of-sample $R^2$ of
0.016, and the Theta + VAR(1)-ridge ensemble for FX achieves 0.029 (Table~\ref{tab:r2_forecaster}). While
modest in absolute terms, out-of-sample $R^2$ of this magnitude can still carry
meaningful economic value \citep{campbell2008predicting}.

\section{Temporal Feature Construction \& Metrics}
\label{app:features}

For each asset $i$ at trading day $t$, we construct an 11-dimensional feature vector following the input representation used in TradeMaster~\citep{trademaster/NEURIPS2023_b8f6f7f2}, adapted from prior work on stock movement prediction~\citep{feng2019enhancingstockmovementprediction,yoo2021accurate}. The features summarize short-term price action and longer-horizon momentum entirely from the daily OHLC and adjusted-close series, requiring no auxiliary data (Table~\ref{tab:features}).

The four intra-day features encode the open, high, and low prices relative to the same day's close, together with the adjusted close, which corrects for corporate actions such as dividends and splits:
\begin{equation}
z_{\text{open}} = \frac{\text{open}_t}{\text{close}_t} - 1, \quad
z_{\text{high}} = \frac{\text{high}_t}{\text{close}_t} - 1, \quad
z_{\text{low}}  = \frac{\text{low}_t}{\text{close}_t}  - 1, \quad
z_{\text{adj}}  = \frac{\text{adj\_close}_t}{\text{close}_t} - 1.
\end{equation}

A single one-day return captures day-over-day price change:
\begin{equation}
z_{\text{close}} = \frac{\text{close}_t}{\text{close}_{t-1}} - 1.
\end{equation}

Six moving-average ratios capture momentum at horizons $k \in \{5, 10, 15, 20, 25, 30\}$ trading days, expressing the average close over the last $k$ days relative to today's close:
\begin{equation}
z_{d\_k} = \frac{\tfrac{1}{k}\sum_{j=0}^{k-1} \text{close}_{t-j}}{\text{close}_t} - 1.
\end{equation}
All 11 features are dimensionless ratios relative to the same day's or the
previous day's close, so they are comparable across assets without further
scaling. We apply no additional normalization.

\begin{table}[h]
\centering
\caption{Summary of the 11 temporal features computed per asset per day.}
\vspace{2ex}
\label{tab:features}
\begin{tabular}{ll}
\textbf{Feature} & \textbf{Definition} \\
\midrule
$z_{\text{open}}$  & $\text{open}_t / \text{close}_t - 1$ \\
$z_{\text{high}}$  & $\text{high}_t / \text{close}_t - 1$ \\
$z_{\text{low}}$   & $\text{low}_t  / \text{close}_t - 1$ \\
$z_{\text{adj}}$   & $\text{adj\_close}_t / \text{close}_t - 1$ \\
$z_{\text{close}}$ & $\text{close}_t / \text{close}_{t-1} - 1$ \\
$z_{d\_k}$, $k \in \{5,10,15,20,25,30\}$
                   & $\left(\tfrac{1}{k}\sum_{j=0}^{k-1} \text{close}_{t-j}\right) / \text{close}_t - 1$ \\
\bottomrule
\end{tabular}
\end{table}

\subsection{Evaluation Metric Definitions}
\label{app:metrics}

Let $U_t$ denote portfolio value at time $t$, $T$ the test horizon length,
and $\rho_t = U_t / U_{t-1} - 1$ the daily portfolio return. We annualize
with 252 trading days per year and assume a zero risk-free rate, consistent
with the TradeMaster benchmark.

\textbf{Total Return (TR).}
\begin{equation}
\text{TR} = \frac{U_T - U_0}{U_0}
\end{equation}

\textbf{Sharpe Ratio (SR)~\citep{sharpe1994sharpe}.} Annualized ratio of mean excess return to
return volatility:
\begin{equation}
\text{SR} = \sqrt{252} \cdot \frac{\bar{\rho}}{\sigma_\rho},
\quad
\bar{\rho} = \tfrac{1}{T}\sum_{t=1}^T \rho_t,
\quad
\sigma_\rho = \sqrt{\tfrac{1}{T-1}\sum_{t=1}^T (\rho_t - \bar{\rho})^2}
\end{equation}

\textbf{Sortino Ratio (SoR).} Annualized ratio of mean return to
downside deviation:
\begin{equation}
\text{SoR} = \sqrt{252} \cdot \frac{\bar{\rho}}{\sigma_d},
\quad
\sigma_d = \sqrt{\tfrac{1}{T}\sum_{t=1}^T \min(\rho_t, 0)^2}
\end{equation}

\textbf{Maximum Drawdown (MDD).} Worst peak-to-trough decline over the
test period:
\begin{equation}
\text{MDD} = \max_{t \in [0,T]} \left(
  \frac{\max_{s \le t} U_s - U_t}{\max_{s \le t} U_s}
\right)
\end{equation}

\textbf{Calmar Ratio (CR).} Annualized return divided by maximum
drawdown:
\begin{equation}
\text{CR} = \frac{(1 + \text{TR})^{252/T} - 1}{\text{MDD}}
\end{equation}

\begin{algorithm}[!t]
\caption{$\fpilotabs$: Inference-Time Policy Adaptation for Portfolio Management}
\label{alg:finpilot-full}
\begin{algorithmic}[1]

\Require Pre-trained actor $\pi_\theta$, frozen critic $\mathcal{V}_\psi$,
         forecaster $F$, horizon $H$, particles $K$, epochs $E$,
         step size $\alpha$, discount $\gamma$, risk weight $\lambda$,
         noise scale $\sigma$

\For{$t=1,\ldots,T$}

    \State Observe real state $s_t$, pre-rebalancing portfolio
           $w_t^{-}$, and wealth $U_t$

    \State Predict the $H$-step state trajectory
           $\{\hat{s}_{t+h}\}_{h=1}^{H}
           \leftarrow F(s_{\leq t})$
           \Comment{forecaster frozen}

    \For{$k=1,\ldots,K$}
        \For{$h=1,\ldots,H$}
            \State Sample forecast error
            $\varepsilon_{t,h}^{(k)}
            \sim
            \mathcal{N}
            (0,\sigma^2\Sigma_h^{\mathrm{val}})$

            \State Construct imagined state
            $\tilde{s}_{t+h}^{(k)}
            \leftarrow
            \operatorname{Perturb}
            (\hat{s}_{t+h},\varepsilon_{t,h}^{(k)})$
        \EndFor
    \EndFor
    \Statex \Comment{Particles are fixed for all $E$ adaptation epochs}

    \For{$e=1,\ldots,E$}

        \For{$k=1,\ldots,K$}
            \State $J_t^{(k)}
            \leftarrow
            J_t
            \big(
                \theta;
                \tilde{s}_{t:t+H}^{(k)},
                w_t^{-},
                U_t
            \big)$
            \Comment{\eqref{eq:imagined-return}}
        \EndFor

        \State $\bar{J}_t
        \leftarrow
        \frac{1}{K}
        \sum_{k=1}^{K}J_t^{(k)}$

        \State $\widehat{\mathcal{D}}_t
        \leftarrow
        \frac{1}{K}
        \sum_{k=1}^{K}
        \min
        \left(
            J_t^{(k)}-\bar{J}_t,\,
            0
        \right)^2$

        \State $\widehat{\mathcal{J}}_t
        \leftarrow
        \bar{J}_t
        -
        \lambda
        \sqrt{
            \widehat{\mathcal{D}}_t+\epsilon
        }$

        \State $\theta
        \leftarrow
        \theta
        +
        \alpha
        \nabla_\theta
        \widehat{\mathcal{J}}_t$
        \Comment{actor parameters only}

    \EndFor

    \State Execute $w_t=\pi_\theta(s_t)$ in the real environment

\EndFor

\end{algorithmic}
\end{algorithm}

\section{Experimental Setup}
\subsection{Training Details and Hyperparameter Selection}
\label{app:pre-trained-setup}
\paragraph{Baseline RL training.} The TradeMaster paper does not 
release the specific hyperparameters used for their reported baselines, 
so we re-tuned each algorithm on our setup. Following the search space 
described in the TradeMaster paper, we grid-searched batch size in 
$\{256, 512, 1024\}$, hidden size in $\{64, 128, 256\}$, and 
actor/critic learning rate in $\{3\mathrm{e}{-4}, 5\mathrm{e}{-4}, 
7\mathrm{e}{-4}, 9\mathrm{e}{-4}\}$ for each of PPO, SAC, A2C, TD3, 
and DDPG, with Adam as the optimizer in all cases. Selection was based 
on the best total return on the validation split. The selected 
configuration was then retrained across 5 random seeds to produce the 
reported baseline numbers.

\subsection{$\fpilotabs$ configuration}
\label{app:finpilot-config}

Table~\ref{tab:finpilot-hparams} lists all inference-time settings. Only the
risk weight $\lambda$ is selected per setting; every other value is shared
across algorithms, markets, and windows.

\paragraph{Shared settings.} The planning horizon was chosen from
$H \in \{1, 15, 50\}$ during preliminary development on DJ30 data through
2021, which precedes every validation and test year in our protocol, and
fixed at $H = 50$ thereafter. The step size ($\alpha = 0.01$), number of
adaptation epochs ($E = 2$), and number of forecast particles
($K = 512$) are single values shared across all settings, with no
per-setting tuning. Imagined rollouts start from the agent's current
portfolio value (initial capital $10^5$, compounding over the test period),
the same reward scale on which the critic was trained, so imagined rewards
and the terminal bootstrap share the same currency units. The actor is
updated with Adam, with gradients clipped to unit norm. The discount
$\gamma = 0.99$ matches pretraining; with $H = 50$, the terminal bootstrap
receives weight $\gamma^H \approx 0.61$. The forecast-noise covariance
$\Sigma^{\mathrm{val}}_h$ is estimated from forecast errors on the validation
split of each window and used without rescaling ($\sigma = 1$), so the
dispersion of imagined trajectories reflects the forecaster's observed error
rather than a tuned noise level.

\paragraph{Risk-weight selection.} For each market, algorithm, and window, we
run $\fpilotabs$ on the validation year with each
$\lambda \in \{0.05, 0.10, 0.15\}$ and select the value with the highest
mean validation Sortino ratio across seeds. The selected $\lambda$ is held
fixed for the corresponding test year. DJ30 and FX are tuned independently
under the same protocol. We select only $\lambda$ per setting because it
encodes a risk preference, namely the trade-off between imagined return and
downside dispersion, whose appropriate level can differ across policies with
different base risk profiles and across markets. We use the Sortino ratio
rather than total return for this choice because $\lambda$ penalizes downside
dispersion, so a downside-risk-adjusted criterion matches its role. A post-hoc sensitivity analysis over the step size and number of adaptation
epochs is given in Appendix~\ref{app:alpha}.

\begin{table}[h]
\centering
\setlength{\tabcolsep}{4pt}
\caption{$\fpilotabs$ inference-time hyperparameters. $\lambda$ is chosen on each
window's validation year; $H$ was fixed during development on pre-2022 data;
all other $\fpilotabs$ settings are shared and untuned.}
\label{tab:finpilot-hparams}
\vspace{2ex}
\begin{tabular}{llll}
\toprule
Symbol & Description & Value & How set \\
\midrule
$H$ & Planning horizon (days) & 50 & Prelim.\ dev.\ (DJ30, $\le$2021) \\
$\alpha$ & Actor step size & 0.01 & Shared \\
$E$ & Adaptation epochs per decision & 2 & Shared \\
$K$ & Forecast particles & 512 & Shared \\
$\gamma$ & Discount factor & 0.99 & Matches pretraining \\
$\sigma$ & Noise scale on $\Sigma^{\mathrm{val}}_h$ & 1 & Unscaled validation cov. \\
$\epsilon$ & Stabilizer in downside penalty & $10^{-8}$ & Numerical constant \\
$c_{\text{tx}}$ & Transaction cost & 10 bps & Environment \\
$U_0$ & Initial capital & $10^5$ & Environment \\
-- & Actor optimizer & Adam & -- \\
-- & Gradient-norm clip & 1 & -- \\
$\lambda$ & Risk weight & $\{0.05, 0.10, 0.15\}$ & Val.\ Sortino, per setting \\
\bottomrule
\end{tabular}
\end{table}

\subsection{Forecasters and forecast skill}
\label{app:r2-def}

\paragraph{Definition of $R^2_{t,i}$.} The controlled forecasts in
Section~\ref{sec:cheating} rescale errors using the original forecast's skill $R^2_{t,i}$ at
each origin $t$ and asset $i$. Let $d_{s,i}$ denote the realized one-day price
change of asset $i$ on day $s$, so that $d_{t,i,h} = d_{t+h,i}$ in the notation
of Section~\ref{sec:cheating}. The baseline is the trailing mean of realized changes available
at the origin,
\begin{equation}
\bar d_{t,i} = \frac{1}{L}\sum_{j=1}^{L} d_{t-j,i},
\end{equation}
with $L = 30$ trading days, held constant across the horizon. The skill is
\begin{equation}
R^2_{t,i} = 1 - \frac{\sum_{h=1}^{H} \bigl(\hat d_{t,i,h} - d_{t,i,h}\bigr)^2}
                     {\sum_{h=1}^{H} \bigl(\bar d_{t,i} - d_{t,i,h}\bigr)^2}.
\end{equation}
The construction requires $R^2_{t,i} < 1$, which holds at every origin in our
data. Table~\ref{tab:r2_forecaster} reports this skill for each forecaster and
test year, pooled over forecast origins, horizons, and assets.

\begin{table}[h]
\centering
\caption{Out-of-sample $R^2$ of the learned forecasters on each test year.
$R^2$ is computed on daily close-price moves over the $H = 50$ planning
horizon, relative to the trailing 30-day mean move, and pooled over forecast
origins, horizons, and assets.}
\vspace{2ex}
\label{tab:r2_forecaster}
\begin{tabular}{llcccc}
\toprule
 & & \multicolumn{3}{c}{Test year} & \\
\cmidrule(lr){3-5}
Market & Forecaster & 2023 & 2024 & 2025 & Mean \\
\midrule
DJ30 & XGBoost              & 0.027 & 0.005 & 0.014 & 0.016 \\
FX   & Theta + VAR(1)-ridge & 0.039 & 0.029 & 0.019 & 0.029 \\
\bottomrule
\end{tabular}
\end{table}

\subsection{Non-RL baselines}
\label{app:baselines}

All baselines run in the same environment as $\fpilotabs$, with the same
transaction cost ($c_{\text{tx}} = 10$ bps on $\ell_1$ turnover).

\paragraph{Passive allocators.} \emph{Buy-and-hold} invests $1/N$ in each of
the $N$ risky assets on the first day of the test year and never trades again.
\emph{1/N} rebalances to equal weights across the $N$ risky assets every day.
Neither holds cash or uses a forecast.

\paragraph{Forecast-driven optimizers.} Following
\citet{boyd_cvx_DBLP:journals/ftopt/BoydBDKKNS17}, adapted to our long-only,
unlevered setting with the environment's linear transaction cost (no holding
cost or market-impact term), the single-period (SPO, $H = 1$) and multi-period
(MPO, $H = 50$, matching $\fpilotabs$'s planning horizon) optimizers choose
weights over cash and the $N$ assets by solving, at each decision step $t$,
\begin{equation}
\max_{w_1,\ldots,w_H}\; \sum_{h=1}^{H} \Big(
  \hat\mu_{t,h}^{\top} w_h
  - \gamma_{\text{risk}}\, w_h^{\top} \hat\Sigma_t\, w_h
  - \gamma_{\text{trade}}\, c_{\text{tx}} \|w_h - w_{h-1}\|_1 \Big)
\quad \text{s.t.}\; w_h \ge 0,\; \mathbf 1^{\top} w_h = 1,
\end{equation}
with $w_0$ the current pre-rebalancing weights, where cash has zero expected
return and zero risk, and execute only $w_1$ before replanning at the next
step. The expected returns $\hat\mu_{t,h}$ are the predicted one-day
close-price returns at horizon $h$, taken from exactly the same forecasts that
$\fpilotabs$ receives. The planner ignores weight drift within the horizon;
the environment applies the realized dynamics. The risk model $\hat\Sigma_t$
is the factor model of
\citet[Sec.~7.3]{boyd_cvx_DBLP:journals/ftopt/BoydBDKKNS17} with 15 factors,
estimated from the trailing 500 trading days of realized daily returns and
re-estimated on the first trading day of each month, using only returns
realized up to that date. Problems are solved with CVXPY using OSQP.

\paragraph{Hyperparameter selection.} We search 87 distinct pairs
$(\gamma_{\text{risk}}, \gamma_{\text{trade}})$, the union of the coarse grid
\[
  \{0.1, 1, 10, 100, 1000\} \times \{1, 2, 5, 10, 20\}
\]
and the finer grid
\[
  \{0.03, 0.056, 0.1, 0.178, 0.316, 0.562, 1, 1.78\}
  \times \{7, 8, 9, 12, 15, 20, 30, 50\}.
\] For each test year $Y$, we select the pair
with the highest mean final portfolio value over the years preceding it
(2022 through $Y{-}1$), each evaluated with a forecaster trained only on data
preceding that year; no result from year $Y$ is used. Ties go to the smaller
$\gamma_{\text{risk}}$, then the smaller $\gamma_{\text{trade}}$.

\paragraph{Cash-only solutions on FX.} On FX, the forecast-implied expected
returns are small relative to the trading penalty: every grid pair with
$\gamma_{\text{trade}} \ge 7$ remains essentially in cash (final value within
0.01\% of the initial value) in all three test years, and the pairs selected
on the preceding years all have $\gamma_{\text{trade}} \ge 20$, reflecting losses from trading in 2022, the first selection year.


\section{Additional Experimental Results}
This appendix reports the per-window results behind the three-window averages
in the main text (Appendices~\ref{app:empirical} and~\ref{app:non-rl}), the
full results of the controlled forecast-skill experiments
(Appendix~\ref{app:controlled-results}), and a post-hoc sensitivity analysis
over the step size and number of adaptation epochs (Appendix~\ref{app:alpha}).

\subsection{Per-Window Comparison with Frozen Policies}
\label{app:empirical}

Tables~\ref{tab:dj30_final} and~\ref{tab:fx_final} report per-window results
underlying the three-window averages in Table~\ref{tab:main} and the
per-window changes in Figure~\ref{fig:window-improvement}. On DJ30, $\fpilotabs$
improves total return over Base in all 15 algorithm--windows and Sharpe in 13;
in the two exceptions (A2C in 2024 and DDPG in 2025), return still improves
while Sharpe falls slightly. Maximum drawdown increases in 12 of the 15 DJ30
algorithm--windows, consistent with Section~\ref{sec:frozen-policy-results}. On FX, total return improves
in 13 of 15 algorithm--windows and Sharpe in 14. PPO and A2C in 2025 lose
return but improve Sharpe and reduce drawdown, and SAC in 2024 gains return
but loses Sharpe. In 2024, when all five base policies lost between 5.7\% and
7.4\%, $\fpilotabs$ reduced both the loss and the maximum drawdown for every
algorithm (Table~\ref{tab:fx_final}). Because each $\fpilotabs$ run starts from the Base policy with the
same seed, comparisons are paired; overlap of the $\pm$ ranges does not by
itself imply that a difference is not significant.

\begin{table}[!h]
\centering
\caption{Per-window performance on DJ30 (three rolling test windows, XGBoost
forecaster). Values are mean $\pm$ one standard deviation across random seeds
($n=5$ per window cell). Base denotes the frozen pre-trained policy; FinPILOT denotes the same agent under our method. For each window,
algorithm, and metric, the better of the two means is shown in \textbf{bold}
(higher is better for all metrics except MDD, where lower is better); values
tied at two decimals are both bolded.}
\label{tab:dj30_final}
\vspace{2ex}
\setlength{\tabcolsep}{5pt}
\resizebox{\textwidth}{!}{%
\begin{tabular}{lll rrrrr}
\toprule
Window & Algo & Method & Return (\%) & Sharpe & Sortino & Calmar & MDD (\%) \\
\midrule
\multirow{10}{*}{2023} & \multirow{2}{*}{PPO} & Base & $13.84 \pm 0.39$ & $1.25 \pm 0.02$ & $2.07 \pm 0.04$ & $1.64 \pm 0.04$ & {\boldmath$8.27 \pm 0.05$} \\
 & & FinPILOT & {\boldmath$26.16 \pm 0.43$} & {\boldmath$1.50 \pm 0.02$} & {\boldmath$2.34 \pm 0.04$} & {\boldmath$2.03 \pm 0.03$} & $12.11 \pm 0.00$ \\
\cmidrule(l){2-8}
 & \multirow{2}{*}{SAC} & Base & $14.25 \pm 0.01$ & $1.28 \pm 0.00$ & $2.12 \pm 0.00$ & $1.68 \pm 0.00$ & {\boldmath$8.30 \pm 0.00$} \\
 & & FinPILOT & {\boldmath$18.82 \pm 0.01$} & {\boldmath$1.51 \pm 0.00$} & {\boldmath$2.61 \pm 0.00$} & {\boldmath$2.01 \pm 0.00$} & $8.94 \pm 0.00$ \\
\cmidrule(l){2-8}
 & \multirow{2}{*}{A2C} & Base & $14.15 \pm 0.12$ & $1.27 \pm 0.01$ & $2.10 \pm 0.02$ & $1.67 \pm 0.01$ & {\boldmath$8.31 \pm 0.03$} \\
 & & FinPILOT & {\boldmath$44.11 \pm 0.84$} & {\boldmath$2.20 \pm 0.03$} & {\boldmath$3.51 \pm 0.04$} & {\boldmath$4.21 \pm 0.16$} & $9.04 \pm 0.22$ \\
\cmidrule(l){2-8}
 & \multirow{2}{*}{TD3} & Base & $11.85 \pm 1.84$ & $1.03 \pm 0.18$ & $1.64 \pm 0.30$ & $1.18 \pm 0.25$ & {\boldmath$10.32 \pm 1.77$} \\
 & & FinPILOT & {\boldmath$53.69 \pm 8.19$} & {\boldmath$2.05 \pm 0.18$} & {\boldmath$3.56 \pm 0.39$} & {\boldmath$3.69 \pm 0.60$} & $12.45 \pm 1.61$ \\
\cmidrule(l){2-8}
 & \multirow{2}{*}{DDPG} & Base & $9.09 \pm 2.25$ & $0.85 \pm 0.19$ & $1.36 \pm 0.32$ & $1.05 \pm 0.30$ & {\boldmath$9.07 \pm 1.49$} \\
 & & FinPILOT & {\boldmath$64.31 \pm 6.98$} & {\boldmath$2.42 \pm 0.30$} & {\boldmath$4.55 \pm 0.78$} & {\boldmath$4.53 \pm 0.96$} & $11.73 \pm 1.53$ \\
\midrule
\multirow{10}{*}{2024} & \multirow{2}{*}{PPO} & Base & $10.78 \pm 0.13$ & $1.05 \pm 0.01$ & $1.52 \pm 0.02$ & $1.77 \pm 0.03$ & {\boldmath$6.09 \pm 0.07$} \\
 & & FinPILOT & {\boldmath$21.01 \pm 1.12$} & {\boldmath$1.29 \pm 0.05$} & {\boldmath$1.83 \pm 0.09$} & {\boldmath$1.88 \pm 0.09$} & $10.80 \pm 0.17$ \\
\cmidrule(l){2-8}
 & \multirow{2}{*}{SAC} & Base & $10.95 \pm 0.01$ & $1.06 \pm 0.00$ & $1.54 \pm 0.00$ & $1.77 \pm 0.00$ & $6.15 \pm 0.00$ \\
 & & FinPILOT & {\boldmath$11.31 \pm 0.00$} & {\boldmath$1.09 \pm 0.00$} & {\boldmath$1.62 \pm 0.00$} & {\boldmath$2.01 \pm 0.00$} & {\boldmath$5.60 \pm 0.00$} \\
\cmidrule(l){2-8}
 & \multirow{2}{*}{A2C} & Base & $10.93 \pm 0.05$ & {\boldmath$1.06 \pm 0.01$} & {\boldmath$1.53 \pm 0.01$} & {\boldmath$1.77 \pm 0.02$} & {\boldmath$6.18 \pm 0.04$} \\
 & & FinPILOT & {\boldmath$12.29 \pm 0.32$} & $0.98 \pm 0.02$ & $1.42 \pm 0.03$ & $1.37 \pm 0.04$ & $9.07 \pm 0.06$ \\
\cmidrule(l){2-8}
 & \multirow{2}{*}{TD3} & Base & $11.12 \pm 7.02$ & $1.01 \pm 0.62$ & $1.37 \pm 0.71$ & $1.64 \pm 1.04$ & {\boldmath$6.92 \pm 0.97$} \\
 & & FinPILOT & {\boldmath$26.65 \pm 6.93$} & {\boldmath$1.41 \pm 0.39$} & {\boldmath$2.10 \pm 0.66$} & {\boldmath$2.37 \pm 0.87$} & $11.80 \pm 3.99$ \\
\cmidrule(l){2-8}
 & \multirow{2}{*}{DDPG} & Base & $10.37 \pm 5.26$ & $0.86 \pm 0.40$ & $1.22 \pm 0.62$ & $1.38 \pm 0.85$ & {\boldmath$8.39 \pm 1.63$} \\
 & & FinPILOT & {\boldmath$23.86 \pm 5.94$} & {\boldmath$1.41 \pm 0.36$} & {\boldmath$2.06 \pm 0.70$} & {\boldmath$2.36 \pm 0.68$} & $9.97 \pm 2.25$ \\
\midrule
\multirow{10}{*}{2025} & \multirow{2}{*}{PPO} & Base & $15.65 \pm 0.28$ & $1.01 \pm 0.02$ & $1.34 \pm 0.03$ & $1.04 \pm 0.02$ & $15.23 \pm 0.15$ \\
 & & FinPILOT & {\boldmath$23.68 \pm 2.23$} & {\boldmath$1.17 \pm 0.09$} & {\boldmath$1.61 \pm 0.13$} & {\boldmath$1.61 \pm 0.11$} & {\boldmath$14.42 \pm 0.38$} \\
\cmidrule(l){2-8}
 & \multirow{2}{*}{SAC} & Base & $16.09 \pm 0.08$ & $1.04 \pm 0.00$ & $1.37 \pm 0.00$ & $1.08 \pm 0.00$ & {\boldmath$14.92 \pm 0.01$} \\
 & & FinPILOT & {\boldmath$17.73 \pm 0.06$} & {\boldmath$1.05 \pm 0.00$} & {\boldmath$1.44 \pm 0.00$} & {\boldmath$1.17 \pm 0.00$} & $15.15 \pm 0.01$ \\
\cmidrule(l){2-8}
 & \multirow{2}{*}{A2C} & Base & $16.11 \pm 0.11$ & $1.04 \pm 0.00$ & $1.37 \pm 0.01$ & $1.08 \pm 0.01$ & $14.96 \pm 0.04$ \\
 & & FinPILOT & {\boldmath$25.11 \pm 0.80$} & {\boldmath$1.18 \pm 0.03$} & {\boldmath$1.76 \pm 0.05$} & {\boldmath$1.80 \pm 0.06$} & {\boldmath$13.67 \pm 0.09$} \\
\cmidrule(l){2-8}
 & \multirow{2}{*}{TD3} & Base & $15.82 \pm 4.09$ & $0.95 \pm 0.15$ & $1.28 \pm 0.19$ & $0.93 \pm 0.13$ & {\boldmath$17.09 \pm 2.22$} \\
 & & FinPILOT & {\boldmath$28.56 \pm 12.28$} & {\boldmath$1.18 \pm 0.40$} & {\boldmath$1.80 \pm 0.70$} & {\boldmath$1.59 \pm 0.61$} & $18.38 \pm 5.07$ \\
\cmidrule(l){2-8}
 & \multirow{2}{*}{DDPG} & Base & $12.94 \pm 4.77$ & {\boldmath$0.85 \pm 0.29$} & $1.10 \pm 0.34$ & $0.81 \pm 0.26$ & {\boldmath$16.73 \pm 1.96$} \\
 & & FinPILOT & {\boldmath$20.38 \pm 9.89$} & $0.81 \pm 0.32$ & {\boldmath$1.29 \pm 0.55$} & {\boldmath$1.30 \pm 0.57$} & $17.42 \pm 2.61$ \\
\bottomrule
\end{tabular}}
\end{table}

\begin{table}[h]
\centering
\caption{Per-window performance on FX (three rolling
test windows, Theta\,+\,VAR(1)-ridge forecaster). Values are mean $\pm$ one
standard deviation across random seeds ($n=5$ per window cell). Base denotes the frozen pre-trained policy; FinPILOT denotes the same agent under
our method. For each window, algorithm, and metric, the better of the two means
is shown in \textbf{bold} (higher is better for all metrics except MDD, where
lower is better); values tied at two decimals are both bolded.}
\label{tab:fx_final}
\vspace{2ex}
\setlength{\tabcolsep}{5pt}
\resizebox{\textwidth}{!}{%
\begin{tabular}{lll rrrrr}
\toprule
Window & Algo & Method & Return (\%) & Sharpe & Sortino & Calmar & MDD (\%) \\
\midrule
\multirow{10}{*}{2023} & \multirow{2}{*}{PPO} & Base & $2.05 \pm 0.13$ & $0.39 \pm 0.02$ & $0.65 \pm 0.04$ & $0.40 \pm 0.03$ & $5.51 \pm 0.22$ \\
 & & FinPILOT & {\boldmath$2.33 \pm 0.97$} & {\boldmath$0.54 \pm 0.21$} & {\boldmath$0.92 \pm 0.33$} & {\boldmath$0.45 \pm 0.20$} & {\boldmath$5.42 \pm 0.52$} \\
\cmidrule(l){2-8}
 & \multirow{2}{*}{SAC} & Base & $2.32 \pm 0.02$ & $0.45 \pm 0.00$ & $0.74 \pm 0.01$ & {\boldmath$0.46 \pm 0.00$} & $5.35 \pm 0.01$ \\
 & & FinPILOT & {\boldmath$2.33 \pm 0.02$} & {\boldmath$0.49 \pm 0.00$} & {\boldmath$0.82 \pm 0.01$} & {\boldmath$0.46 \pm 0.00$} & {\boldmath$5.28 \pm 0.01$} \\
\cmidrule(l){2-8}
 & \multirow{2}{*}{A2C} & Base & $2.31 \pm 0.03$ & $0.45 \pm 0.01$ & $0.74 \pm 0.01$ & $0.46 \pm 0.01$ & $5.35 \pm 0.01$ \\
 & & FinPILOT & {\boldmath$2.78 \pm 0.03$} & {\boldmath$0.61 \pm 0.01$} & {\boldmath$1.06 \pm 0.02$} & {\boldmath$0.54 \pm 0.01$} & {\boldmath$5.27 \pm 0.05$} \\
\cmidrule(l){2-8}
 & \multirow{2}{*}{TD3} & Base & $2.63 \pm 0.84$ & $0.51 \pm 0.15$ & $0.82 \pm 0.23$ & $0.50 \pm 0.19$ & {\boldmath$5.67 \pm 0.63$} \\
 & & FinPILOT & {\boldmath$4.76 \pm 1.78$} & {\boldmath$0.90 \pm 0.27$} & {\boldmath$1.49 \pm 0.47$} & {\boldmath$0.75 \pm 0.26$} & $6.45 \pm 1.20$ \\
\cmidrule(l){2-8}
 & \multirow{2}{*}{DDPG} & Base & $2.62 \pm 1.77$ & $0.51 \pm 0.31$ & $0.81 \pm 0.51$ & $0.50 \pm 0.30$ & {\boldmath$5.70 \pm 0.83$} \\
 & & FinPILOT & {\boldmath$4.96 \pm 2.60$} & {\boldmath$1.00 \pm 0.44$} & {\boldmath$1.63 \pm 0.80$} & {\boldmath$0.79 \pm 0.38$} & $6.31 \pm 0.83$ \\
\midrule
\multirow{10}{*}{2024} & \multirow{2}{*}{PPO} & Base & $-5.82 \pm 0.10$ & $-1.36 \pm 0.01$ & $-1.87 \pm 0.02$ & {\boldmath$-0.90 \pm 0.01$} & $6.53 \pm 0.10$ \\
 & & FinPILOT & {\boldmath$-4.60 \pm 0.24$} & {\boldmath$-1.24 \pm 0.05$} & {\boldmath$-1.67 \pm 0.08$} & $-0.95 \pm 0.04$ & {\boldmath$4.85 \pm 0.34$} \\
\cmidrule(l){2-8}
 & \multirow{2}{*}{SAC} & Base & $-5.74 \pm 0.07$ & {\boldmath$-1.35 \pm 0.01$} & {\boldmath$-1.87 \pm 0.02$} & {\boldmath$-0.90 \pm 0.01$} & $6.49 \pm 0.04$ \\
 & & FinPILOT & {\boldmath$-5.53 \pm 0.05$} & $-1.42 \pm 0.01$ & $-1.89 \pm 0.01$ & $-1.00 \pm 0.01$ & {\boldmath$5.60 \pm 0.01$} \\
\cmidrule(l){2-8}
 & \multirow{2}{*}{A2C} & Base & $-5.78 \pm 0.00$ & $-1.36 \pm 0.00$ & $-1.88 \pm 0.00$ & $-0.90 \pm 0.00$ & $6.51 \pm 0.01$ \\
 & & FinPILOT & {\boldmath$-3.79 \pm 0.24$} & {\boldmath$-1.05 \pm 0.05$} & {\boldmath$-1.43 \pm 0.07$} & {\boldmath$-0.84 \pm 0.02$} & {\boldmath$4.51 \pm 0.19$} \\
\cmidrule(l){2-8}
 & \multirow{2}{*}{TD3} & Base & $-7.36 \pm 1.37$ & $-1.56 \pm 0.24$ & $-2.24 \pm 0.27$ & $-0.99 \pm 0.05$ & $7.43 \pm 1.21$ \\
 & & FinPILOT & {\boldmath$-3.23 \pm 1.51$} & {\boldmath$-0.83 \pm 0.33$} & {\boldmath$-1.18 \pm 0.39$} & {\boldmath$-0.67 \pm 0.14$} & {\boldmath$4.55 \pm 1.44$} \\
\cmidrule(l){2-8}
 & \multirow{2}{*}{DDPG} & Base & $-6.78 \pm 1.31$ & $-1.62 \pm 0.28$ & $-2.33 \pm 0.49$ & $-1.01 \pm 0.03$ & $6.68 \pm 1.20$ \\
 & & FinPILOT & {\boldmath$-4.36 \pm 1.58$} & {\boldmath$-1.25 \pm 0.48$} & {\boldmath$-1.61 \pm 0.48$} & {\boldmath$-0.87 \pm 0.15$} & {\boldmath$4.93 \pm 1.48$} \\
\midrule
\multirow{10}{*}{2025} & \multirow{2}{*}{PPO} & Base & {\boldmath$7.00 \pm 0.26$} & $1.40 \pm 0.04$ & $2.41 \pm 0.08$ & $2.70 \pm 0.11$ & $2.55 \pm 0.04$ \\
 & & FinPILOT & $6.14 \pm 0.78$ & {\boldmath$1.54 \pm 0.15$} & {\boldmath$3.05 \pm 0.29$} & {\boldmath$3.31 \pm 0.85$} & {\boldmath$1.91 \pm 0.45$} \\
\cmidrule(l){2-8}
 & \multirow{2}{*}{SAC} & Base & $7.02 \pm 0.07$ & $1.39 \pm 0.00$ & $2.39 \pm 0.01$ & $2.70 \pm 0.01$ & $2.56 \pm 0.03$ \\
 & & FinPILOT & {\boldmath$7.06 \pm 0.04$} & {\boldmath$1.54 \pm 0.01$} & {\boldmath$2.80 \pm 0.02$} & {\boldmath$3.07 \pm 0.01$} & {\boldmath$2.25 \pm 0.00$} \\
\cmidrule(l){2-8}
 & \multirow{2}{*}{A2C} & Base & {\boldmath$7.06 \pm 0.02$} & $1.39 \pm 0.00$ & $2.39 \pm 0.01$ & {\boldmath$2.71 \pm 0.01$} & $2.57 \pm 0.01$ \\
 & & FinPILOT & $5.34 \pm 0.08$ & {\boldmath$1.43 \pm 0.01$} & {\boldmath$2.93 \pm 0.02$} & $2.66 \pm 0.04$ & {\boldmath$1.98 \pm 0.01$} \\
\cmidrule(l){2-8}
 & \multirow{2}{*}{TD3} & Base & $6.35 \pm 1.05$ & $1.33 \pm 0.19$ & $2.32 \pm 0.29$ & $2.64 \pm 0.74$ & $2.47 \pm 0.47$ \\
 & & FinPILOT & {\boldmath$8.74 \pm 0.86$} & {\boldmath$1.53 \pm 0.09$} & {\boldmath$2.86 \pm 0.18$} & {\boldmath$3.55 \pm 0.18$} & {\boldmath$2.40 \pm 0.22$} \\
\cmidrule(l){2-8}
 & \multirow{2}{*}{DDPG} & Base & $3.52 \pm 1.85$ & $0.79 \pm 0.39$ & $1.35 \pm 0.67$ & {\boldmath$1.33 \pm 0.97$} & {\boldmath$3.18 \pm 1.10$} \\
 & & FinPILOT & {\boldmath$3.84 \pm 0.91$} & {\boldmath$0.82 \pm 0.17$} & {\boldmath$1.51 \pm 0.28$} & $1.07 \pm 0.39$ & $3.76 \pm 0.62$ \\
\bottomrule
\end{tabular}}
\end{table}

\clearpage

\subsection{Comparison with non-RL baselines}
\label{app:non-rl}
Tables~\ref{tab:nonrl_dj30_per_year} and~\ref{tab:nonrl_fx_per_year} report
per-window results for the non-RL baselines, underlying the three-window
averages in Table~\ref{tab:nonrl}; their construction and hyperparameter
selection are described in Appendix~\ref{app:baselines}. On DJ30, the best $\fpilotabs$-adapted policy exceeds every baseline in total
return and Sortino ratio in all three windows, while Boyd SPO attains the
highest Sharpe ratio in 2024 and Boyd MPO the lowest drawdown in 2023.
Individual adapted policies do not always beat the passive allocators; for
example, $\fpilotabs$-SAC and $\fpilotabs$-A2C trail buy-and-hold in 2024 return.
On FX, the Boyd optimizers remain essentially in cash in every window, so
their near-zero returns and drawdowns reflect inactivity. Among portfolios
exposed to the market, a $\fpilotabs$-adapted policy attains the highest return
in 2023 and 2025, the smallest loss in 2024 ($-3.23\%$ for TD3 vs.\
$-5.91\%$ for buy-and-hold), and the lowest drawdown in all three windows.

\begin{table}[!ht]
\centering
\caption{Per-window comparison of FinPILOT with non-RL baselines on DJ30
(XGBoost forecaster). FinPILOT values are mean $\pm$ one standard deviation
across random seeds ($n=5$ per window cell); the baselines
are deterministic. SPO and MPO use the same forecast as FinPILOT, with
hyperparameters selected on the years preceding each test year
(Appendix~\ref{app:baselines}); the other baselines use no forecast. Best mean
per window and column in \textbf{bold} (lower is better for MDD); values tied
at two decimals are all bolded.}
\vspace{2ex}
\label{tab:nonrl_dj30_per_year}
\footnotesize
\setlength{\tabcolsep}{3.25pt}
\resizebox{\textwidth}{!}{%
\begin{tabular}{ll *{5}{r@{}l}}
\toprule
Window & Method & \multicolumn{2}{c}{Return (\%)} & \multicolumn{2}{c}{Sharpe} & \multicolumn{2}{c}{Sortino} & \multicolumn{2}{c}{Calmar} & \multicolumn{2}{c}{MDD (\%)} \\
\midrule
\multirow{9}{*}{2023} & Buy-and-hold & $15.61$ &  & $1.33$ &  & $2.22$ &  & $1.72$ &  & $8.84$ &  \\
 & 1/N (daily rebalanced) & $14.79$ &  & $1.28$ &  & $2.12$ &  & $1.68$ &  & $8.59$ &  \\
 & SPO ($H{=}1$) & $-4.94$ &  & $-0.37$ &  & $-0.42$ &  & $-0.28$ &  & $15.40$ &  \\
 & MPO ($H{=}50$) & $2.24$ &  & $0.33$ &  & $0.31$ &  & $0.35$ &  & {\boldmath$7.30$} &  \\
\cmidrule(l){2-12}
 & FinPILOT-PPO & $26.16$ & $\,{\pm}\,0.43$ & $1.50$ & $\,{\pm}\,0.02$ & $2.34$ & $\,{\pm}\,0.04$ & $2.03$ & $\,{\pm}\,0.03$ & $12.11$ & $\,{\pm}\,0.00$ \\
 & FinPILOT-SAC & $18.82$ & $\,{\pm}\,0.01$ & $1.51$ & $\,{\pm}\,0.00$ & $2.61$ & $\,{\pm}\,0.00$ & $2.01$ & $\,{\pm}\,0.00$ & $8.94$ & $\,{\pm}\,0.00$ \\
 & FinPILOT-A2C & $44.11$ & $\,{\pm}\,0.84$ & $2.20$ & $\,{\pm}\,0.03$ & $3.51$ & $\,{\pm}\,0.04$ & $4.21$ & $\,{\pm}\,0.16$ & $9.04$ & $\,{\pm}\,0.22$ \\
 & FinPILOT-TD3 & $53.69$ & $\,{\pm}\,8.19$ & $2.05$ & $\,{\pm}\,0.18$ & $3.56$ & $\,{\pm}\,0.39$ & $3.69$ & $\,{\pm}\,0.60$ & $12.45$ & $\,{\pm}\,1.61$ \\
 & FinPILOT-DDPG & {\boldmath$64.31$} & {\boldmath$\,{\pm}\,6.98$} & {\boldmath$2.42$} & {\boldmath$\,{\pm}\,0.30$} & {\boldmath$4.55$} & {\boldmath$\,{\pm}\,0.78$} & {\boldmath$4.53$} & {\boldmath$\,{\pm}\,0.96$} & $11.73$ & $\,{\pm}\,1.53$ \\
\midrule
\multirow{9}{*}{2024} & Buy-and-hold & $12.62$ &  & $1.18$ &  & $1.75$ &  & $2.14$ &  & $5.81$ &  \\
 & 1/N (daily rebalanced) & $11.34$ &  & $1.06$ &  & $1.54$ &  & $1.78$ &  & $6.37$ &  \\
 & SPO ($H{=}1$) & $20.54$ &  & {\boldmath$1.42$} &  & $1.64$ &  & $2.27$ &  & $8.66$ &  \\
 & MPO ($H{=}50$) & $4.42$ &  & $0.30$ &  & $0.37$ &  & $0.37$ &  & $20.47$ &  \\
\cmidrule(l){2-12}
 & FinPILOT-PPO & $21.01$ & $\,{\pm}\,1.12$ & $1.29$ & $\,{\pm}\,0.05$ & $1.83$ & $\,{\pm}\,0.09$ & $1.88$ & $\,{\pm}\,0.09$ & $10.80$ & $\,{\pm}\,0.17$ \\
 & FinPILOT-SAC & $11.31$ & $\,{\pm}\,0.00$ & $1.09$ & $\,{\pm}\,0.00$ & $1.62$ & $\,{\pm}\,0.00$ & $2.01$ & $\,{\pm}\,0.00$ & {\boldmath$5.60$} & {\boldmath$\,{\pm}\,0.00$} \\
 & FinPILOT-A2C & $12.29$ & $\,{\pm}\,0.32$ & $0.98$ & $\,{\pm}\,0.02$ & $1.42$ & $\,{\pm}\,0.03$ & $1.37$ & $\,{\pm}\,0.04$ & $9.07$ & $\,{\pm}\,0.06$ \\
 & FinPILOT-TD3 & {\boldmath$26.65$} & {\boldmath$\,{\pm}\,6.93$} & $1.41$ & $\,{\pm}\,0.39$ & {\boldmath$2.10$} & {\boldmath$\,{\pm}\,0.66$} & {\boldmath$2.37$} & {\boldmath$\,{\pm}\,0.87$} & $11.80$ & $\,{\pm}\,3.99$ \\
 & FinPILOT-DDPG & $23.86$ & $\,{\pm}\,5.94$ & $1.41$ & $\,{\pm}\,0.36$ & $2.06$ & $\,{\pm}\,0.70$ & $2.36$ & $\,{\pm}\,0.68$ & $9.97$ & $\,{\pm}\,2.25$ \\
\midrule
\multirow{9}{*}{2025} & Buy-and-hold & $15.86$ &  & $1.01$ &  & $1.34$ &  & $1.04$ &  & $15.43$ &  \\
 & 1/N (daily rebalanced) & $16.64$ &  & $1.04$ &  & $1.37$ &  & $1.08$ &  & $15.42$ &  \\
 & SPO ($H{=}1$) & $11.26$ &  & $0.49$ &  & $0.68$ &  & $0.55$ &  & $28.81$ &  \\
 & MPO ($H{=}50$) & $12.08$ &  & $0.49$ &  & $0.78$ &  & $0.65$ &  & $29.66$ &  \\
\cmidrule(l){2-12}
 & FinPILOT-PPO & $23.68$ & $\,{\pm}\,2.23$ & $1.17$ & $\,{\pm}\,0.09$ & $1.61$ & $\,{\pm}\,0.13$ & $1.61$ & $\,{\pm}\,0.11$ & $14.42$ & $\,{\pm}\,0.38$ \\
 & FinPILOT-SAC & $17.73$ & $\,{\pm}\,0.06$ & $1.05$ & $\,{\pm}\,0.00$ & $1.44$ & $\,{\pm}\,0.00$ & $1.17$ & $\,{\pm}\,0.00$ & $15.15$ & $\,{\pm}\,0.01$ \\
 & FinPILOT-A2C & $25.11$ & $\,{\pm}\,0.80$ & {\boldmath$1.18$} & {\boldmath$\,{\pm}\,0.03$} & $1.76$ & $\,{\pm}\,0.05$ & {\boldmath$1.80$} & {\boldmath$\,{\pm}\,0.06$} & {\boldmath$13.67$} & {\boldmath$\,{\pm}\,0.09$} \\
 & FinPILOT-TD3 & {\boldmath$28.56$} & {\boldmath$\,{\pm}\,12.28$} & {\boldmath$1.18$} & {\boldmath$\,{\pm}\,0.40$} & {\boldmath$1.80$} & {\boldmath$\,{\pm}\,0.70$} & $1.59$ & $\,{\pm}\,0.61$ & $18.38$ & $\,{\pm}\,5.07$ \\
 & FinPILOT-DDPG & $20.38$ & $\,{\pm}\,9.89$ & $0.81$ & $\,{\pm}\,0.32$ & $1.29$ & $\,{\pm}\,0.55$ & $1.30$ & $\,{\pm}\,0.57$ & $17.42$ & $\,{\pm}\,2.61$ \\
\bottomrule
\end{tabular}}
\end{table}

\begin{table}[!ht]
\centering
\caption{Per-window comparison of FinPILOT with non-RL baselines on FX
(Theta + VAR(1)-ridge forecaster). FinPILOT values are mean $\pm$ one standard
deviation across random seeds ($n=5$ per window cell);
the baselines are deterministic. SPO and MPO use the same forecast as
FinPILOT, with hyperparameters selected on the years preceding each test year
(Appendix~\ref{app:baselines}); the other baselines use no forecast.
$^\dagger$SPO and MPO remain essentially in cash, losing less than
\$1 on \$100{,}000 in every window; their returns and drawdowns are below
0.01\% in magnitude. Their Calmar and MDD are excluded when marking the best
value, and their Sharpe, Sortino, and Calmar ratios, computed from these
near-zero returns, are numerically unstable. Best mean per window and column in
\textbf{bold} (lower is better for MDD); values tied at two decimals are all
bolded.}
\vspace{2ex}
\label{tab:nonrl_fx_per_year}
\footnotesize
\setlength{\tabcolsep}{2.91pt}
\resizebox{\textwidth}{!}{%
\begin{tabular}{ll *{5}{r@{}l}}
\toprule
Window & Method & \multicolumn{2}{c}{Return (\%)} & \multicolumn{2}{c}{Sharpe} & \multicolumn{2}{c}{Sortino} & \multicolumn{2}{c}{Calmar} & \multicolumn{2}{c}{MDD (\%)} \\
\midrule
\multirow{9}{*}{2023} & Buy-and-hold & $2.39$ &  & $0.45$ &  & $0.74$ &  & $0.45$ &  & $5.59$ &  \\
 & 1/N (daily rebalanced) & $2.42$ &  & $0.45$ &  & $0.74$ &  & $0.46$ &  & $5.58$ &  \\
 & SPO ($H{=}1$) & $0.00$ & $^\dagger$ & $-0.01$ &  & $-0.02$ &  & $-0.01$ & $^\dagger$ & $<\!0.01$ & $^\dagger$ \\
 & MPO ($H{=}50$) & $0.00$ & $^\dagger$ & $-1.75$ &  & $-2.07$ &  & $-0.88$ & $^\dagger$ & $<\!0.01$ & $^\dagger$ \\
\cmidrule(l){2-12}
 & FinPILOT-PPO & $2.33$ & $\,{\pm}\,0.97$ & $0.54$ & $\,{\pm}\,0.21$ & $0.92$ & $\,{\pm}\,0.33$ & $0.45$ & $\,{\pm}\,0.20$ & $5.42$ & $\,{\pm}\,0.52$ \\
 & FinPILOT-SAC & $2.33$ & $\,{\pm}\,0.02$ & $0.49$ & $\,{\pm}\,0.00$ & $0.82$ & $\,{\pm}\,0.01$ & $0.46$ & $\,{\pm}\,0.00$ & $5.28$ & $\,{\pm}\,0.01$ \\
 & FinPILOT-A2C & $2.78$ & $\,{\pm}\,0.03$ & $0.61$ & $\,{\pm}\,0.01$ & $1.06$ & $\,{\pm}\,0.02$ & $0.54$ & $\,{\pm}\,0.01$ & {\boldmath$5.27$} & {\boldmath$\,{\pm}\,0.05$} \\
 & FinPILOT-TD3 & $4.76$ & $\,{\pm}\,1.78$ & $0.90$ & $\,{\pm}\,0.27$ & $1.49$ & $\,{\pm}\,0.47$ & $0.75$ & $\,{\pm}\,0.26$ & $6.45$ & $\,{\pm}\,1.20$ \\
 & FinPILOT-DDPG & {\boldmath$4.96$} & {\boldmath$\,{\pm}\,2.60$} & {\boldmath$1.00$} & {\boldmath$\,{\pm}\,0.44$} & {\boldmath$1.63$} & {\boldmath$\,{\pm}\,0.80$} & {\boldmath$0.79$} & {\boldmath$\,{\pm}\,0.38$} & $6.31$ & $\,{\pm}\,0.83$ \\
\midrule
\multirow{9}{*}{2024} & Buy-and-hold & $-5.91$ &  & $-1.34$ &  & $-1.83$ &  & $-0.89$ &  & $6.73$ &  \\
 & 1/N (daily rebalanced) & $-6.03$ &  & $-1.36$ &  & $-1.87$ &  & $-0.90$ &  & $6.80$ &  \\
 & SPO ($H{=}1$) & {\boldmath$0.00$} & {\boldmath$^\dagger$} & $-1.56$ &  & $-1.69$ &  & $-0.92$ & $^\dagger$ & $<\!0.01$ & $^\dagger$ \\
 & MPO ($H{=}50$) & {\boldmath$0.00$} & {\boldmath$^\dagger$} & $-4.29$ &  & $-4.70$ &  & $-1.00$ & $^\dagger$ & $<\!0.01$ & $^\dagger$ \\
\cmidrule(l){2-12}
 & FinPILOT-PPO & $-4.60$ & $\,{\pm}\,0.24$ & $-1.24$ & $\,{\pm}\,0.05$ & $-1.67$ & $\,{\pm}\,0.08$ & $-0.95$ & $\,{\pm}\,0.04$ & $4.85$ & $\,{\pm}\,0.34$ \\
 & FinPILOT-SAC & $-5.53$ & $\,{\pm}\,0.05$ & $-1.42$ & $\,{\pm}\,0.01$ & $-1.89$ & $\,{\pm}\,0.01$ & $-1.00$ & $\,{\pm}\,0.01$ & $5.60$ & $\,{\pm}\,0.01$ \\
 & FinPILOT-A2C & $-3.79$ & $\,{\pm}\,0.24$ & $-1.05$ & $\,{\pm}\,0.05$ & $-1.43$ & $\,{\pm}\,0.07$ & $-0.84$ & $\,{\pm}\,0.02$ & {\boldmath$4.51$} & {\boldmath$\,{\pm}\,0.19$} \\
 & FinPILOT-TD3 & $-3.23$ & $\,{\pm}\,1.51$ & {\boldmath$-0.83$} & {\boldmath$\,{\pm}\,0.33$} & {\boldmath$-1.18$} & {\boldmath$\,{\pm}\,0.39$} & {\boldmath$-0.67$} & {\boldmath$\,{\pm}\,0.14$} & $4.55$ & $\,{\pm}\,1.44$ \\
 & FinPILOT-DDPG & $-4.36$ & $\,{\pm}\,1.58$ & $-1.25$ & $\,{\pm}\,0.48$ & $-1.61$ & $\,{\pm}\,0.48$ & $-0.87$ & $\,{\pm}\,0.15$ & $4.93$ & $\,{\pm}\,1.48$ \\
\midrule
\multirow{9}{*}{2025} & Buy-and-hold & $7.46$ &  & $1.39$ &  & $2.39$ &  & $2.72$ &  & $2.70$ &  \\
 & 1/N (daily rebalanced) & $7.38$ &  & $1.39$ &  & $2.40$ &  & $2.70$ &  & $2.69$ &  \\
 & SPO ($H{=}1$) & $0.00$ & $^\dagger$ & $-0.57$ &  & $-0.71$ &  & $-0.64$ & $^\dagger$ & $<\!0.01$ & $^\dagger$ \\
 & MPO ($H{=}50$) & $0.00$ & $^\dagger$ & $-2.89$ &  & $-3.77$ &  & $-0.98$ & $^\dagger$ & $<\!0.01$ & $^\dagger$ \\
\cmidrule(l){2-12}
 & FinPILOT-PPO & $6.14$ & $\,{\pm}\,0.78$ & {\boldmath$1.54$} & {\boldmath$\,{\pm}\,0.15$} & {\boldmath$3.05$} & {\boldmath$\,{\pm}\,0.29$} & $3.31$ & $\,{\pm}\,0.85$ & {\boldmath$1.91$} & {\boldmath$\,{\pm}\,0.45$} \\
 & FinPILOT-SAC & $7.06$ & $\,{\pm}\,0.04$ & {\boldmath$1.54$} & {\boldmath$\,{\pm}\,0.01$} & $2.80$ & $\,{\pm}\,0.02$ & $3.07$ & $\,{\pm}\,0.01$ & $2.25$ & $\,{\pm}\,0.00$ \\
 & FinPILOT-A2C & $5.34$ & $\,{\pm}\,0.08$ & $1.43$ & $\,{\pm}\,0.01$ & $2.93$ & $\,{\pm}\,0.02$ & $2.66$ & $\,{\pm}\,0.04$ & $1.98$ & $\,{\pm}\,0.01$ \\
 & FinPILOT-TD3 & {\boldmath$8.74$} & {\boldmath$\,{\pm}\,0.86$} & $1.53$ & $\,{\pm}\,0.09$ & $2.86$ & $\,{\pm}\,0.18$ & {\boldmath$3.55$} & {\boldmath$\,{\pm}\,0.18$} & $2.40$ & $\,{\pm}\,0.22$ \\
 & FinPILOT-DDPG & $3.84$ & $\,{\pm}\,0.91$ & $0.82$ & $\,{\pm}\,0.17$ & $1.51$ & $\,{\pm}\,0.28$ & $1.07$ & $\,{\pm}\,0.39$ & $3.76$ & $\,{\pm}\,0.62$ \\
\bottomrule
\end{tabular}}
\end{table}




\FloatBarrier
\subsection{Controlled forecast-skill results}
\label{app:controlled-results}
Tables~\ref{tab:r2-subset-dj30-2023}--\ref{tab:r2-subset-fx-2025} report
per-window results for every target $R^2_\star$, algorithm, and market,
underlying Figure~\ref{fig:forecast-skill}. Controlled forecasts are
constructed as in Section~4.4, with $R^2_{t,i}$ defined in
Appendix~\ref{app:r2-def}. The Base row is the frozen pre-trained policy,
and the Original forecast row uses the learned forecaster.

\begin{table}[h]
\caption{$\fpilotabs$ performance with controlled forecasts on DJ30, test year 2023,
at target skill levels from $R^2_\star=0.005$ to $R^2_\star=0.8$
(Section~\ref{sec:cheating}). Values are mean $\pm$ one standard deviation across
5 seeds. Base is the frozen pre-trained policy and does not depend on $R^2_\star$;
Original forecast is the unmodified forecast ($c_{t,i}=1$).}
  \vspace{2ex}
  
  \label{tab:r2-subset-dj30-2023}
  \footnotesize
  \centering
  
  \resizebox{0.9\textwidth}{!}{%
  \begin{tabular}{ll ccccc}
    \toprule
    Algorithm & 	$R^2_\star$ & Total return (\%) $\uparrow$ & Sharpe $\uparrow$ & Sortino $\uparrow$ & Calmar $\uparrow$ & MDD (\%) $\downarrow$ \\
    \midrule
    \textbf{PPO}   	& Base & 13.84 $\pm$ 0.39 & 1.25 $\pm$ 0.02 & 2.07 $\pm$ 0.04 & 1.64 $\pm$ 0.04 & 8.27 $\pm$ 0.05 \\
    \cmidrule(lr){2-7}
                & Original forecast & 26.16 $\pm$ 0.43 & 1.50 $\pm$ 0.02 & 2.34 $\pm$ 0.04 & 2.03 $\pm$ 0.03 & 12.11 $\pm$ 0.00 \\
                   & 	$R^2_\star=0.005$ & 25.44 $\pm$ 0.38 & 1.52 $\pm$ 0.02 & 2.44 $\pm$ 0.03 & 1.95 $\pm$ 0.03 & 12.27 $\pm$ 0.07 \\
                   & 	$R^2_\star=0.01$ & 25.74 $\pm$ 0.36 & 1.54 $\pm$ 0.02 & 2.46 $\pm$ 0.03 & 1.97 $\pm$ 0.03 & 12.29 $\pm$ 0.06 \\
                   & 	$R^2_\star=0.025$ & 26.73 $\pm$ 0.34 & 1.58 $\pm$ 0.02 & 2.55 $\pm$ 0.03 & 2.02 $\pm$ 0.03 & 12.35 $\pm$ 0.04 \\
                   & 	$R^2_\star=0.05$ & 28.67 $\pm$ 0.28 & 1.67 $\pm$ 0.01 & 2.74 $\pm$ 0.03 & 2.13 $\pm$ 0.02 & 12.41 $\pm$ 0.02 \\
                   & 	$R^2_\star=0.1$ & 32.11 $\pm$ 0.36 & 1.83 $\pm$ 0.02 & 3.00 $\pm$ 0.03 & 2.34 $\pm$ 0.02 & 12.46 $\pm$ 0.03 \\
                   & 	$R^2_\star=0.2$ & 51.38 $\pm$ 0.63 & 2.66 $\pm$ 0.02 & 4.73 $\pm$ 0.05 & 3.46 $\pm$ 0.03 & 12.36 $\pm$ 0.10 \\
                   & 	$R^2_\star=0.4$ & 93.19 $\pm$ 0.91 & 3.66 $\pm$ 0.02 & 7.32 $\pm$ 0.08 & 6.70 $\pm$ 0.21 & 10.11 $\pm$ 0.25 \\
                   & 	$R^2_\star=0.8$ & 128.31 $\pm$ 1.00 & 4.49 $\pm$ 0.02 & 8.94 $\pm$ 0.06 & 9.68 $\pm$ 0.10 & 8.73 $\pm$ 0.11 \\
    \midrule
    \textbf{SAC}  	& Base & 14.25 $\pm$ 0.01 & 1.28 $\pm$ 0.00 & 2.12 $\pm$ 0.00 & 1.68 $\pm$ 0.00 & 8.30 $\pm$ 0.00 \\
    \cmidrule(lr){2-7}
    & Original forecast & 18.82 $\pm$ 0.01 & 1.51 $\pm$ 0.00 & 2.61 $\pm$ 0.00 & 2.01 $\pm$ 0.00 & 8.94 $\pm$ 0.00 \\
                   & 	$R^2_\star=0.005$ & 17.85 $\pm$ 0.01 & 1.44 $\pm$ 0.00 & 2.47 $\pm$ 0.00 & 1.86 $\pm$ 0.00 & 9.24 $\pm$ 0.00 \\
                   & 	$R^2_\star=0.01$ & 18.05 $\pm$ 0.01 & 1.45 $\pm$ 0.00 & 2.49 $\pm$ 0.00 & 1.88 $\pm$ 0.00 & 9.22 $\pm$ 0.00 \\
                   & 	$R^2_\star=0.025$ & 18.64 $\pm$ 0.01 & 1.49 $\pm$ 0.00 & 2.57 $\pm$ 0.00 & 1.94 $\pm$ 0.00 & 9.17 $\pm$ 0.00 \\
                   & 	$R^2_\star=0.05$ & 19.56 $\pm$ 0.01 & 1.56 $\pm$ 0.00 & 2.67 $\pm$ 0.00 & 2.05 $\pm$ 0.00 & 9.08 $\pm$ 0.00 \\
                   & 	$R^2_\star=0.1$ & 21.65 $\pm$ 0.01 & 1.70 $\pm$ 0.00 & 2.93 $\pm$ 0.00 & 2.29 $\pm$ 0.00 & 8.88 $\pm$ 0.00 \\
                   & 	$R^2_\star=0.2$ & 26.78 $\pm$ 0.02 & 2.04 $\pm$ 0.00 & 3.54 $\pm$ 0.00 & 2.96 $\pm$ 0.00 & 8.27 $\pm$ 0.00 \\
                   & 	$R^2_\star=0.4$ & 35.61 $\pm$ 0.01 & 2.58 $\pm$ 0.00 & 4.55 $\pm$ 0.00 & 4.16 $\pm$ 0.00 & 7.50 $\pm$ 0.00 \\
                   & 	$R^2_\star=0.8$ & 42.19 $\pm$ 0.01 & 3.06 $\pm$ 0.00 & 5.49 $\pm$ 0.00 & 6.10 $\pm$ 0.00 & 5.89 $\pm$ 0.00 \\
    \midrule
    \textbf{A2C}   & Base & 14.15 $\pm$ 0.12 & 1.27 $\pm$ 0.01 & 2.10 $\pm$ 0.02 & 1.67 $\pm$ 0.01 & 8.31 $\pm$ 0.03 \\
    \cmidrule(lr){2-7}
     & Original forecast & 44.11 $\pm$ 0.84 & 2.20 $\pm$ 0.03 & 3.51 $\pm$ 0.04 & 4.21 $\pm$ 0.16 & 9.04 $\pm$ 0.22 \\
                   & 	$R^2_\star=0.005$ & 39.71 $\pm$ 0.95 & 1.99 $\pm$ 0.03 & 3.07 $\pm$ 0.05 & 3.63 $\pm$ 0.17 & 9.67 $\pm$ 0.26 \\
                   & 	$R^2_\star=0.01$ & 43.23 $\pm$ 0.98 & 2.09 $\pm$ 0.03 & 3.27 $\pm$ 0.06 & 4.23 $\pm$ 0.22 & 8.90 $\pm$ 0.31 \\
                   & 	$R^2_\star=0.025$ & 52.50 $\pm$ 0.24 & 2.31 $\pm$ 0.01 & 3.81 $\pm$ 0.02 & 5.60 $\pm$ 0.03 & 7.86 $\pm$ 0.02 \\
                   & 	$R^2_\star=0.05$ & 56.62 $\pm$ 0.28 & 2.34 $\pm$ 0.01 & 4.10 $\pm$ 0.01 & 5.99 $\pm$ 0.02 & 7.84 $\pm$ 0.00 \\
                   & 	$R^2_\star=0.1$ & 70.14 $\pm$ 0.91 & 2.57 $\pm$ 0.04 & 4.77 $\pm$ 0.09 & 6.10 $\pm$ 0.43 & 9.14 $\pm$ 0.58 \\
                   & 	$R^2_\star=0.2$ & 79.49 $\pm$ 0.29 & 2.89 $\pm$ 0.01 & 5.25 $\pm$ 0.02 & 6.89 $\pm$ 0.12 & 8.82 $\pm$ 0.13 \\
                   & 	$R^2_\star=0.4$ & 119.15 $\pm$ 0.71 & 4.07 $\pm$ 0.02 & 7.40 $\pm$ 0.02 & 10.20 $\pm$ 0.04 & 7.89 $\pm$ 0.00 \\
                   & 	$R^2_\star=0.8$ & 145.02 $\pm$ 0.30 & 4.25 $\pm$ 0.01 & 7.91 $\pm$ 0.02 & 9.64 $\pm$ 0.04 & 9.56 $\pm$ 0.03 \\
    \midrule
    \textbf{TD3}   & Base & 11.85 $\pm$ 1.84 & 1.03 $\pm$ 0.18 & 1.64 $\pm$ 0.30 & 1.18 $\pm$ 0.25 & 10.32 $\pm$ 1.77 \\
    \cmidrule(lr){2-7}
    & Original forecast & 53.69 $\pm$ 8.19 & 2.05 $\pm$ 0.18 & 3.56 $\pm$ 0.39 & 3.69 $\pm$ 0.60 & 12.45 $\pm$ 1.61 \\
                   & 	$R^2_\star=0.005$ & 41.57 $\pm$ 17.56 & 1.66 $\pm$ 0.49 & 2.81 $\pm$ 1.12 & 2.76 $\pm$ 0.96 & 13.45 $\pm$ 0.92 \\
                   & 	$R^2_\star=0.01$ & 49.98 $\pm$ 12.59 & 1.90 $\pm$ 0.29 & 3.15 $\pm$ 0.88 & 3.55 $\pm$ 0.58 & 12.10 $\pm$ 1.13 \\
                   & 	$R^2_\star=0.025$ & 50.96 $\pm$ 14.85 & 1.97 $\pm$ 0.40 & 3.42 $\pm$ 0.93 & 3.90 $\pm$ 1.16 & 11.51 $\pm$ 2.29 \\
                   & 	$R^2_\star=0.05$ & 58.75 $\pm$ 13.94 & 2.21 $\pm$ 0.36 & 3.95 $\pm$ 0.76 & 3.89 $\pm$ 0.93 & 12.66 $\pm$ 1.68 \\
                   & 	$R^2_\star=0.1$ & 70.21 $\pm$ 11.19 & 2.54 $\pm$ 0.33 & 4.59 $\pm$ 0.74 & 5.16 $\pm$ 1.37 & 11.14 $\pm$ 1.99 \\
                   & 	$R^2_\star=0.2$ & 69.98 $\pm$ 12.03 & 2.42 $\pm$ 0.25 & 4.45 $\pm$ 0.55 & 4.75 $\pm$ 1.61 & 12.24 $\pm$ 2.08 \\
                   & 	$R^2_\star=0.4$ & 103.54 $\pm$ 19.85 & 3.34 $\pm$ 0.31 & 6.49 $\pm$ 0.82 & 9.20 $\pm$ 2.48 & 8.25 $\pm$ 1.48 \\
                   & 	$R^2_\star=0.8$ & 548.88 $\pm$ 396.61 & 7.07 $\pm$ 1.20 & 18.05 $\pm$ 5.49 & 33.08 $\pm$ 17.86 & 5.89 $\pm$ 1.72 \\
    \midrule
    \textbf{DDPG}  & Base & 9.09 $\pm$ 2.25 & 0.85 $\pm$ 0.19 & 1.36 $\pm$ 0.32 & 1.05 $\pm$ 0.30 & 9.07 $\pm$ 1.49 \\
    \cmidrule(lr){2-7}
     & Original forecast & 64.31 $\pm$ 6.98 & 2.42 $\pm$ 0.30 & 4.55 $\pm$ 0.78 & 4.53 $\pm$ 0.96 & 11.73 $\pm$ 1.53 \\
                   & 	$R^2_\star=0.005$ & 55.04 $\pm$ 15.26 & 2.17 $\pm$ 0.32 & 3.88 $\pm$ 0.71 & 3.84 $\pm$ 0.93 & 12.12 $\pm$ 2.03 \\
                   & 	$R^2_\star=0.01$ & 59.53 $\pm$ 8.64 & 2.25 $\pm$ 0.25 & 4.20 $\pm$ 0.58 & 3.72 $\pm$ 0.19 & 13.17 $\pm$ 1.41 \\
                   & 	$R^2_\star=0.025$ & 64.80 $\pm$ 10.41 & 2.42 $\pm$ 0.24 & 4.41 $\pm$ 0.63 & 4.48 $\pm$ 0.94 & 11.94 $\pm$ 2.12 \\
                   & 	$R^2_\star=0.05$ & 63.00 $\pm$ 13.43 & 2.27 $\pm$ 0.31 & 4.06 $\pm$ 0.78 & 3.94 $\pm$ 0.99 & 13.28 $\pm$ 1.46 \\
                   & 	$R^2_\star=0.1$ & 74.40 $\pm$ 2.61 & 2.60 $\pm$ 0.15 & 4.77 $\pm$ 0.28 & 4.51 $\pm$ 0.61 & 13.06 $\pm$ 1.54 \\
                   & 	$R^2_\star=0.2$ & 82.85 $\pm$ 13.71 & 2.77 $\pm$ 0.35 & 5.16 $\pm$ 0.81 & 6.22 $\pm$ 1.82 & 10.59 $\pm$ 2.46 \\
                   & 	$R^2_\star=0.4$ & 111.89 $\pm$ 31.58 & 3.33 $\pm$ 0.77 & 6.59 $\pm$ 1.63 & 8.59 $\pm$ 3.72 & 9.90 $\pm$ 2.83 \\
                   & 	$R^2_\star=0.8$ & 993.53 $\pm$ 601.88 & 8.23 $\pm$ 1.78 & 20.18 $\pm$ 5.22 & 33.67 $\pm$ 22.23 & 7.97 $\pm$ 3.17 \\
    \bottomrule
  \end{tabular}
  }%
\end{table}

                   
\begin{table}[ht]
\caption{$\fpilotabs$ performance with controlled forecasts on DJ30, test year 2024,
at target skill levels from $R^2_\star=0.005$ to $R^2_\star=0.8$
(Section~\ref{sec:cheating}). Values are mean $\pm$ one standard deviation across
5 seeds.
Base is the frozen pre-trained policy and does not depend on $R^2_\star$;
Original forecast is the unmodified forecast ($c_{t,i}=1$).}
  \vspace{2ex}
  
  \label{tab:r2-subset-dj30-2024}
  \centering
  \resizebox{\textwidth}{!}{%
  \begin{tabular}{ll ccccc}
    \toprule
    Algorithm & 	$R^2_\star$ & Total return (\%) $\uparrow$ & Sharpe $\uparrow$ & Sortino $\uparrow$ & Calmar $\uparrow$ & MDD (\%) $\downarrow$ \\
    \midrule
    \textbf{PPO}   & Base & 10.78 $\pm$ 0.13 & 1.05 $\pm$ 0.01 & 1.52 $\pm$ 0.02 & 1.77 $\pm$ 0.03 & 6.09 $\pm$ 0.07 \\
    \cmidrule(lr){2-7}
    & Original forecast & 21.01 $\pm$ 1.12 & 1.29 $\pm$ 0.05 & 1.83 $\pm$ 0.09 & 1.88 $\pm$ 0.09 & 10.80 $\pm$ 0.17 \\
                   & 	$R^2_\star=0.005$ & 16.07 $\pm$ 0.25 & 1.04 $\pm$ 0.01 & 1.62 $\pm$ 0.02 & 1.74 $\pm$ 0.02 & 9.26 $\pm$ 0.00 \\
                   & 	$R^2_\star=0.01$ & 16.95 $\pm$ 0.48 & 1.09 $\pm$ 0.03 & 1.69 $\pm$ 0.04 & 1.82 $\pm$ 0.04 & 9.27 $\pm$ 0.00 \\
                   & 	$R^2_\star=0.025$ & 20.43 $\pm$ 0.11 & 1.28 $\pm$ 0.01 & 1.99 $\pm$ 0.01 & 2.14 $\pm$ 0.01 & 9.27 $\pm$ 0.00 \\
                   & 	$R^2_\star=0.05$ & 24.75 $\pm$ 0.19 & 1.51 $\pm$ 0.01 & 2.36 $\pm$ 0.02 & 2.53 $\pm$ 0.02 & 9.20 $\pm$ 0.05 \\
                   & 	$R^2_\star=0.1$ & 45.10 $\pm$ 0.26 & 2.37 $\pm$ 0.02 & 4.13 $\pm$ 0.02 & 4.21 $\pm$ 0.02 & 9.16 $\pm$ 0.04 \\
                   & 	$R^2_\star=0.2$ & 62.02 $\pm$ 0.19 & 2.78 $\pm$ 0.01 & 5.03 $\pm$ 0.02 & 5.57 $\pm$ 0.08 & 8.97 $\pm$ 0.12 \\
                   & 	$R^2_\star=0.4$ & 101.24 $\pm$ 0.34 & 4.28 $\pm$ 0.00 & 7.24 $\pm$ 0.03 & 11.85 $\pm$ 0.03 & 6.03 $\pm$ 0.00 \\
                   & 	$R^2_\star=0.8$ & 128.77 $\pm$ 0.31 & 5.10 $\pm$ 0.01 & 9.34 $\pm$ 0.01 & 17.28 $\pm$ 0.03 & 4.87 $\pm$ 0.00 \\
    \midrule
    \textbf{SAC}   & Base & 10.95 $\pm$ 0.01 & 1.06 $\pm$ 0.00 & 1.54 $\pm$ 0.00 & 1.77 $\pm$ 0.00 & 6.15 $\pm$ 0.00 \\
    \cmidrule(lr){2-7}
    & Original forecast & 11.31 $\pm$ 0.00 & 1.09 $\pm$ 0.00 & 1.62 $\pm$ 0.00 & 2.01 $\pm$ 0.00 & 5.60 $\pm$ 0.00 \\
                   & 	$R^2_\star=0.005$ & 11.67 $\pm$ 0.00 & 1.11 $\pm$ 0.00 & 1.67 $\pm$ 0.00 & 2.03 $\pm$ 0.00 & 5.71 $\pm$ 0.00 \\
                   & 	$R^2_\star=0.01$ & 12.02 $\pm$ 0.00 & 1.14 $\pm$ 0.00 & 1.71 $\pm$ 0.00 & 2.09 $\pm$ 0.00 & 5.71 $\pm$ 0.00 \\
                   & 	$R^2_\star=0.025$ & 13.08 $\pm$ 0.00 & 1.22 $\pm$ 0.00 & 1.85 $\pm$ 0.00 & 2.26 $\pm$ 0.00 & 5.69 $\pm$ 0.00 \\
                   & 	$R^2_\star=0.05$ & 14.83 $\pm$ 0.01 & 1.36 $\pm$ 0.00 & 2.07 $\pm$ 0.00 & 2.53 $\pm$ 0.00 & 5.69 $\pm$ 0.00 \\
                   & 	$R^2_\star=0.1$ & 18.22 $\pm$ 0.00 & 1.62 $\pm$ 0.00 & 2.48 $\pm$ 0.00 & 3.04 $\pm$ 0.00 & 5.70 $\pm$ 0.00 \\
                   & 	$R^2_\star=0.2$ & 24.22 $\pm$ 0.01 & 2.07 $\pm$ 0.00 & 3.22 $\pm$ 0.00 & 4.07 $\pm$ 0.00 & 5.48 $\pm$ 0.00 \\
                   & 	$R^2_\star=0.4$ & 31.77 $\pm$ 0.01 & 2.55 $\pm$ 0.00 & 4.02 $\pm$ 0.00 & 5.80 $\pm$ 0.00 & 4.87 $\pm$ 0.00 \\
                   & 	$R^2_\star=0.8$ & 36.50 $\pm$ 0.01 & 2.80 $\pm$ 0.00 & 4.33 $\pm$ 0.00 & 6.92 $\pm$ 0.00 & 4.59 $\pm$ 0.00 \\
    \midrule
    \textbf{A2C}   & Base & 10.93 $\pm$ 0.05 & 1.06 $\pm$ 0.01 & 1.53 $\pm$ 0.01 & 1.77 $\pm$ 0.02 & 6.18 $\pm$ 0.04 \\
    \cmidrule(lr){2-7}
    & Original forecast & 12.29 $\pm$ 0.32 & 0.98 $\pm$ 0.02 & 1.42 $\pm$ 0.03 & 1.37 $\pm$ 0.04 & 9.07 $\pm$ 0.06 \\
                   & 	$R^2_\star=0.005$ & 15.54 $\pm$ 1.07 & 1.19 $\pm$ 0.06 & 1.76 $\pm$ 0.10 & 2.01 $\pm$ 0.14 & 7.59 $\pm$ 0.05 \\
                   & 	$R^2_\star=0.01$ & 17.92 $\pm$ 0.39 & 1.32 $\pm$ 0.02 & 1.99 $\pm$ 0.04 & 2.34 $\pm$ 0.06 & 7.43 $\pm$ 0.05 \\
                   & 	$R^2_\star=0.025$ & 20.51 $\pm$ 0.44 & 1.44 $\pm$ 0.03 & 2.24 $\pm$ 0.04 & 2.83 $\pm$ 0.08 & 6.91 $\pm$ 0.10 \\
                   & 	$R^2_\star=0.05$ & 27.09 $\pm$ 0.47 & 1.78 $\pm$ 0.04 & 2.86 $\pm$ 0.05 & 3.41 $\pm$ 0.12 & 7.33 $\pm$ 0.24 \\
                   & 	$R^2_\star=0.1$ & 32.74 $\pm$ 0.43 & 1.88 $\pm$ 0.03 & 3.13 $\pm$ 0.05 & 3.33 $\pm$ 0.04 & 8.87 $\pm$ 0.01 \\
                   & 	$R^2_\star=0.2$ & 51.26 $\pm$ 0.45 & 2.70 $\pm$ 0.02 & 4.63 $\pm$ 0.03 & 5.23 $\pm$ 0.06 & 8.15 $\pm$ 0.03 \\
                   & 	$R^2_\star=0.4$ & 98.09 $\pm$ 0.48 & 4.37 $\pm$ 0.02 & 8.33 $\pm$ 0.04 & 14.21 $\pm$ 0.32 & 4.91 $\pm$ 0.09 \\
                   & 	$R^2_\star=0.8$ & 107.48 $\pm$ 0.28 & 4.34 $\pm$ 0.01 & 7.66 $\pm$ 0.02 & 13.04 $\pm$ 0.03 & 5.72 $\pm$ 0.01 \\
    \midrule
    \textbf{TD3}   & Base & 11.12 $\pm$ 7.02 & 1.01 $\pm$ 0.62 & 1.37 $\pm$ 0.71 & 1.64 $\pm$ 1.04 & 6.92 $\pm$ 0.97 \\
    \cmidrule(lr){2-7}
     & Original forecast & 26.65 $\pm$ 6.93 & 1.41 $\pm$ 0.39 & 2.10 $\pm$ 0.66 & 2.37 $\pm$ 0.87 & 11.80 $\pm$ 3.99 \\
                   & 	$R^2_\star=0.005$ & 10.88 $\pm$ 11.59 & 0.69 $\pm$ 0.69 & 0.97 $\pm$ 0.97 & 1.19 $\pm$ 1.39 & 12.57 $\pm$ 3.11 \\
                   & 	$R^2_\star=0.01$ & 14.25 $\pm$ 8.97 & 0.87 $\pm$ 0.52 & 1.23 $\pm$ 0.76 & 1.37 $\pm$ 0.99 & 12.75 $\pm$ 3.83 \\
                   & 	$R^2_\star=0.025$ & 25.09 $\pm$ 6.45 & 1.50 $\pm$ 0.35 & 2.29 $\pm$ 0.69 & 3.06 $\pm$ 1.51 & 9.06 $\pm$ 3.69 \\
                   & 	$R^2_\star=0.05$ & 34.36 $\pm$ 7.55 & 1.75 $\pm$ 0.42 & 2.69 $\pm$ 0.88 & 4.11 $\pm$ 1.91 & 8.72 $\pm$ 3.83 \\
                   & 	$R^2_\star=0.1$ & 49.20 $\pm$ 13.67 & 2.26 $\pm$ 0.50 & 3.57 $\pm$ 1.10 & 4.82 $\pm$ 1.36 & 8.80 $\pm$ 1.53 \\
                   & 	$R^2_\star=0.2$ & 103.32 $\pm$ 39.67 & 3.77 $\pm$ 0.90 & 6.79 $\pm$ 1.97 & 10.54 $\pm$ 3.42 & 7.08 $\pm$ 1.59 \\
                   & 	$R^2_\star=0.4$ & 123.49 $\pm$ 59.28 & 4.22 $\pm$ 0.85 & 7.53 $\pm$ 2.12 & 11.98 $\pm$ 4.53 & 6.72 $\pm$ 1.42 \\
                   & 	$R^2_\star=0.8$ & 577.55 $\pm$ 306.04 & 7.47 $\pm$ 1.87 & 17.74 $\pm$ 7.87 & 43.59 $\pm$ 21.61 & 4.52 $\pm$ 1.23 \\
    \midrule
    \textbf{DDPG}  & Base & 10.37 $\pm$ 5.26 & 0.86 $\pm$ 0.40 & 1.22 $\pm$ 0.62 & 1.38 $\pm$ 0.85 & 8.39 $\pm$ 1.63 \\
    \cmidrule(lr){2-7}
    & Original forecast & 23.86 $\pm$ 5.94 & 1.41 $\pm$ 0.36 & 2.06 $\pm$ 0.70 & 2.36 $\pm$ 0.68 & 9.97 $\pm$ 2.25 \\
                   & 	$R^2_\star=0.005$ & 8.68 $\pm$ 8.52 & 0.54 $\pm$ 0.35 & 0.74 $\pm$ 0.50 & 0.82 $\pm$ 0.73 & 14.20 $\pm$ 4.85 \\
                   & 	$R^2_\star=0.01$ & 6.87 $\pm$ 14.42 & 0.43 $\pm$ 0.74 & 0.70 $\pm$ 1.19 & 0.65 $\pm$ 1.06 & 15.44 $\pm$ 4.61 \\
                   & 	$R^2_\star=0.025$ & 22.69 $\pm$ 10.79 & 1.32 $\pm$ 0.63 & 1.97 $\pm$ 1.10 & 2.49 $\pm$ 1.32 & 9.20 $\pm$ 1.29 \\
                   & 	$R^2_\star=0.05$ & 30.27 $\pm$ 3.72 & 1.62 $\pm$ 0.23 & 2.44 $\pm$ 0.29 & 2.68 $\pm$ 0.80 & 11.16 $\pm$ 3.15 \\
                   & 	$R^2_\star=0.1$ & 36.10 $\pm$ 7.94 & 1.97 $\pm$ 0.29 & 2.97 $\pm$ 0.44 & 3.72 $\pm$ 0.70 & 8.63 $\pm$ 0.83 \\
                   & 	$R^2_\star=0.2$ & 66.62 $\pm$ 36.42 & 2.74 $\pm$ 0.65 & 4.69 $\pm$ 1.63 & 6.39 $\pm$ 3.18 & 8.25 $\pm$ 0.83 \\
                   & 	$R^2_\star=0.4$ & 140.11 $\pm$ 42.70 & 3.95 $\pm$ 0.88 & 7.16 $\pm$ 1.22 & 10.90 $\pm$ 3.27 & 8.38 $\pm$ 1.09 \\
                   & 	$R^2_\star=0.8$ & 1102.24 $\pm$ 400.08 & 8.82 $\pm$ 1.11 & 30.78 $\pm$ 5.14 & 74.20 $\pm$ 19.77 & 3.58 $\pm$ 1.28 \\
    \bottomrule
  \end{tabular}
  }%
\end{table}

\begin{table}[ht]
\caption{$\fpilotabs$ performance with controlled forecasts on DJ30, test year 2025,
at target skill levels from $R^2_\star=0.005$ to $R^2_\star=0.8$
(Section~\ref{sec:cheating}). Values are mean $\pm$ one standard deviation across
5 seeds. Base is the frozen pre-trained policy and does not depend on $R^2_\star$;
Original forecast is the unmodified forecast ($c_{t,i}=1$).}
  \vspace{2ex}
  
  \label{tab:r2-subset-dj30-2025}
  \centering
  \resizebox{\textwidth}{!}{%
  \begin{tabular}{ll ccccc}
    \toprule
    Algorithm & 	$R^2_\star$ & Total return (\%) $\uparrow$ & Sharpe $\uparrow$ & Sortino $\uparrow$ & Calmar $\uparrow$ & MDD (\%) $\downarrow$ \\
    \midrule
    \textbf{PPO}   & Base & 15.65 $\pm$ 0.28 & 1.01 $\pm$ 0.02 & 1.34 $\pm$ 0.03 & 1.04 $\pm$ 0.02 & 15.23 $\pm$ 0.15 \\
    \cmidrule(lr){2-7}
     & Original forecast & 23.68 $\pm$ 2.23 & 1.17 $\pm$ 0.09 & 1.61 $\pm$ 0.13 & 1.61 $\pm$ 0.11 & 14.42 $\pm$ 0.38 \\
                   & 	$R^2_\star=0.005$ & 15.38 $\pm$ 1.35 & 0.85 $\pm$ 0.06 & 1.15 $\pm$ 0.09 & 1.11 $\pm$ 0.08 & 14.46 $\pm$ 0.39 \\
                   & 	$R^2_\star=0.01$ & 17.11 $\pm$ 1.57 & 0.94 $\pm$ 0.07 & 1.26 $\pm$ 0.10 & 1.24 $\pm$ 0.09 & 14.24 $\pm$ 0.41 \\
                   & 	$R^2_\star=0.025$ & 22.69 $\pm$ 1.56 & 1.20 $\pm$ 0.07 & 1.63 $\pm$ 0.09 & 1.64 $\pm$ 0.08 & 13.56 $\pm$ 0.46 \\
                   & 	$R^2_\star=0.05$ & 33.12 $\pm$ 1.91 & 1.69 $\pm$ 0.07 & 2.31 $\pm$ 0.11 & 2.45 $\pm$ 0.12 & 12.36 $\pm$ 0.48 \\
                   & 	$R^2_\star=0.1$ & 49.73 $\pm$ 1.65 & 2.44 $\pm$ 0.06 & 3.35 $\pm$ 0.18 & 3.85 $\pm$ 0.22 & 10.90 $\pm$ 0.44 \\
                   & 	$R^2_\star=0.2$ & 96.02 $\pm$ 4.72 & 3.60 $\pm$ 0.15 & 5.15 $\pm$ 0.37 & 6.72 $\pm$ 0.61 & 10.34 $\pm$ 0.63 \\
                   & 	$R^2_\star=0.4$ & 151.72 $\pm$ 1.68 & 3.97 $\pm$ 0.05 & 5.96 $\pm$ 0.12 & 8.05 $\pm$ 0.24 & 11.84 $\pm$ 0.30 \\
                   & 	$R^2_\star=0.8$ & 282.87 $\pm$ 3.72 & 5.20 $\pm$ 0.01 & 8.90 $\pm$ 0.10 & 13.71 $\pm$ 0.34 & 10.07 $\pm$ 0.21 \\
    \midrule
    \textbf{SAC}   & Base & 16.09 $\pm$ 0.08 & 1.04 $\pm$ 0.00 & 1.37 $\pm$ 0.00 & 1.08 $\pm$ 0.00 & 14.92 $\pm$ 0.01 \\
    \cmidrule(lr){2-7}
     & Original forecast & 17.73 $\pm$ 0.06 & 1.05 $\pm$ 0.00 & 1.44 $\pm$ 0.00 & 1.17 $\pm$ 0.00 & 15.15 $\pm$ 0.01 \\
                   & 	$R^2_\star=0.005$ & 15.68 $\pm$ 0.05 & 0.95 $\pm$ 0.00 & 1.30 $\pm$ 0.00 & 1.05 $\pm$ 0.00 & 15.25 $\pm$ 0.01 \\
                   & 	$R^2_\star=0.01$ & 16.08 $\pm$ 0.05 & 0.97 $\pm$ 0.00 & 1.33 $\pm$ 0.00 & 1.07 $\pm$ 0.00 & 15.22 $\pm$ 0.01 \\
                   & 	$R^2_\star=0.025$ & 17.32 $\pm$ 0.05 & 1.03 $\pm$ 0.00 & 1.41 $\pm$ 0.00 & 1.15 $\pm$ 0.00 & 15.11 $\pm$ 0.01 \\
                   & 	$R^2_\star=0.05$ & 19.69 $\pm$ 0.05 & 1.15 $\pm$ 0.00 & 1.57 $\pm$ 0.00 & 1.30 $\pm$ 0.00 & 14.98 $\pm$ 0.01 \\
                   & 	$R^2_\star=0.1$ & 23.71 $\pm$ 0.07 & 1.35 $\pm$ 0.00 & 1.84 $\pm$ 0.00 & 1.56 $\pm$ 0.00 & 14.60 $\pm$ 0.01 \\
                   & 	$R^2_\star=0.2$ & 30.41 $\pm$ 0.08 & 1.67 $\pm$ 0.00 & 2.28 $\pm$ 0.00 & 2.02 $\pm$ 0.01 & 13.82 $\pm$ 0.01 \\
                   & 	$R^2_\star=0.4$ & 39.33 $\pm$ 0.08 & 2.08 $\pm$ 0.00 & 2.87 $\pm$ 0.00 & 2.75 $\pm$ 0.01 & 12.59 $\pm$ 0.02 \\
                   & 	$R^2_\star=0.8$ & 46.28 $\pm$ 0.06 & 2.36 $\pm$ 0.00 & 3.32 $\pm$ 0.00 & 3.35 $\pm$ 0.01 & 11.78 $\pm$ 0.02 \\
    \midrule
    \textbf{A2C}   & Base & 16.11 $\pm$ 0.11 & 1.04 $\pm$ 0.00 & 1.37 $\pm$ 0.01 & 1.08 $\pm$ 0.01 & 14.96 $\pm$ 0.04 \\
    \cmidrule(lr){2-7}
     & Original forecast & 25.11 $\pm$ 0.80 & 1.18 $\pm$ 0.03 & 1.76 $\pm$ 0.05 & 1.80 $\pm$ 0.06 & 13.67 $\pm$ 0.09 \\
                   & 	$R^2_\star=0.005$ & 15.99 $\pm$ 0.64 & 0.84 $\pm$ 0.03 & 1.16 $\pm$ 0.04 & 1.15 $\pm$ 0.04 & 14.62 $\pm$ 0.13 \\
                   & 	$R^2_\star=0.01$ & 17.62 $\pm$ 0.59 & 0.91 $\pm$ 0.03 & 1.26 $\pm$ 0.04 & 1.28 $\pm$ 0.05 & 14.27 $\pm$ 0.15 \\
                   & 	$R^2_\star=0.025$ & 24.40 $\pm$ 0.48 & 1.21 $\pm$ 0.02 & 1.72 $\pm$ 0.04 & 1.82 $\pm$ 0.05 & 13.06 $\pm$ 0.17 \\
                   & 	$R^2_\star=0.05$ & 36.56 $\pm$ 0.78 & 1.72 $\pm$ 0.03 & 2.46 $\pm$ 0.06 & 2.79 $\pm$ 0.10 & 11.87 $\pm$ 0.24 \\
                   & 	$R^2_\star=0.1$ & 60.70 $\pm$ 0.59 & 2.64 $\pm$ 0.02 & 3.66 $\pm$ 0.03 & 4.53 $\pm$ 0.05 & 10.87 $\pm$ 0.05 \\
                   & 	$R^2_\star=0.2$ & 100.71 $\pm$ 0.60 & 3.63 $\pm$ 0.01 & 5.41 $\pm$ 0.02 & 6.67 $\pm$ 0.04 & 10.75 $\pm$ 0.05 \\
                   & 	$R^2_\star=0.4$ & 181.77 $\pm$ 0.73 & 4.57 $\pm$ 0.01 & 7.39 $\pm$ 0.04 & 10.69 $\pm$ 0.05 & 9.96 $\pm$ 0.03 \\
                   & 	$R^2_\star=0.8$ & 340.94 $\pm$ 0.75 & 6.58 $\pm$ 0.01 & 12.70 $\pm$ 0.01 & 27.88 $\pm$ 0.05 & 5.43 $\pm$ 0.01 \\
    \midrule
    \textbf{TD3}   & Base & 15.82 $\pm$ 4.09 & 0.95 $\pm$ 0.15 & 1.28 $\pm$ 0.19 & 0.93 $\pm$ 0.13 & 17.09 $\pm$ 2.22 \\
    \cmidrule(lr){2-7}
    & Original forecast & 28.56 $\pm$ 12.28 & 1.18 $\pm$ 0.40 & 1.80 $\pm$ 0.70 & 1.59 $\pm$ 0.61 & 18.38 $\pm$ 5.07 \\
                   & 	$R^2_\star=0.005$ & 27.34 $\pm$ 9.44 & 1.03 $\pm$ 0.19 & 1.60 $\pm$ 0.40 & 1.48 $\pm$ 0.38 & 18.79 $\pm$ 3.54 \\
                   & 	$R^2_\star=0.01$ & 23.05 $\pm$ 12.33 & 0.94 $\pm$ 0.37 & 1.43 $\pm$ 0.67 & 1.32 $\pm$ 0.74 & 19.93 $\pm$ 5.66 \\
                   & 	$R^2_\star=0.025$ & 21.31 $\pm$ 4.12 & 0.93 $\pm$ 0.23 & 1.39 $\pm$ 0.31 & 1.22 $\pm$ 0.38 & 19.35 $\pm$ 4.10 \\
                   & 	$R^2_\star=0.05$ & 40.38 $\pm$ 23.19 & 1.41 $\pm$ 0.45 & 2.15 $\pm$ 0.75 & 2.09 $\pm$ 1.13 & 17.56 $\pm$ 2.77 \\
                   & 	$R^2_\star=0.1$ & 46.43 $\pm$ 34.17 & 1.53 $\pm$ 0.68 & 2.40 $\pm$ 1.32 & 2.39 $\pm$ 1.43 & 16.99 $\pm$ 4.04 \\
                   & 	$R^2_\star=0.2$ & 78.92 $\pm$ 38.57 & 2.40 $\pm$ 0.65 & 4.01 $\pm$ 1.51 & 4.40 $\pm$ 2.19 & 14.29 $\pm$ 3.08 \\
                   & 	$R^2_\star=0.4$ & 73.19 $\pm$ 52.79 & 2.50 $\pm$ 0.63 & 4.35 $\pm$ 1.95 & 5.89 $\pm$ 3.44 & 9.30 $\pm$ 1.58 \\
                   & 	$R^2_\star=0.8$ & 1393.49 $\pm$ 576.38 & 6.49 $\pm$ 1.44 & 19.24 $\pm$ 5.86 & 34.29 $\pm$ 14.39 & 9.09 $\pm$ 3.25 \\
    \midrule
    \textbf{DDPG}  & Base & 12.94 $\pm$ 4.77 & 0.85 $\pm$ 0.29 & 1.10 $\pm$ 0.34 & 0.81 $\pm$ 0.26 & 16.73 $\pm$ 1.96 \\
    \cmidrule(lr){2-7}
    & Original forecast & 20.38 $\pm$ 9.89 & 0.81 $\pm$ 0.32 & 1.29 $\pm$ 0.55 & 1.30 $\pm$ 0.57 & 17.42 $\pm$ 2.61 \\
                   & 	$R^2_\star=0.005$ & 14.52 $\pm$ 10.45 & 0.68 $\pm$ 0.35 & 1.03 $\pm$ 0.56 & 0.80 $\pm$ 0.41 & 22.21 $\pm$ 4.92 \\
                   & 	$R^2_\star=0.01$ & 17.40 $\pm$ 17.50 & 0.75 $\pm$ 0.63 & 1.13 $\pm$ 1.03 & 0.93 $\pm$ 0.88 & 24.13 $\pm$ 5.49 \\
                   & 	$R^2_\star=0.025$ & 13.17 $\pm$ 7.18 & 0.63 $\pm$ 0.27 & 0.94 $\pm$ 0.42 & 0.83 $\pm$ 0.37 & 20.09 $\pm$ 4.94 \\
                   & 	$R^2_\star=0.05$ & 35.93 $\pm$ 8.78 & 1.40 $\pm$ 0.25 & 2.08 $\pm$ 0.43 & 2.32 $\pm$ 0.43 & 14.54 $\pm$ 1.96 \\
                   & 	$R^2_\star=0.1$ & 65.89 $\pm$ 6.70 & 1.90 $\pm$ 0.23 & 3.07 $\pm$ 0.49 & 3.71 $\pm$ 0.70 & 15.15 $\pm$ 2.77 \\
                   & 	$R^2_\star=0.2$ & 196.86 $\pm$ 149.24 & 3.66 $\pm$ 0.87 & 7.51 $\pm$ 3.68 & 9.86 $\pm$ 6.53 & 11.64 $\pm$ 2.17 \\
                   & 	$R^2_\star=0.4$ & 462.97 $\pm$ 355.25 & 4.18 $\pm$ 0.67 & 8.80 $\pm$ 2.42 & 13.54 $\pm$ 6.71 & 13.28 $\pm$ 4.93 \\
                   & 	$R^2_\star=0.8$ & 2360.66 $\pm$ 1928.59 & 7.75 $\pm$ 1.55 & 25.06 $\pm$ 7.14 & 48.98 $\pm$ 19.01 & 6.64 $\pm$ 1.70 \\
    \bottomrule
  \end{tabular}
  }%
\end{table}

\begin{table}[ht]
  \caption{$\fpilotabs$ performance with controlled forecasts on FX, test year 2023,
at target skill levels from $R^2_\star=0.005$ to $R^2_\star=0.8$
(Section~\ref{sec:cheating}). Values are mean $\pm$ one standard deviation across
5 seeds. Base is the frozen pre-trained policy and does not depend on $R^2_\star$;
Original forecast is the unmodified forecast ($c_{t,i}=1$).}
  \vspace{2ex}
  \label{tab:r2-subset-fx-2023}
  \centering
  \resizebox{\textwidth}{!}{%
  \begin{tabular}{ll ccccc}
    \toprule
    Algorithm & 	$R^2_\star$ & Total return (\%) $\uparrow$ & Sharpe $\uparrow$ & Sortino $\uparrow$ & Calmar $\uparrow$ & MDD (\%) $\downarrow$ \\
    \midrule
    \textbf{PPO}   & Base & 2.05 $\pm$ 0.13 & 0.39 $\pm$ 0.02 & 0.65 $\pm$ 0.04 & 0.40 $\pm$ 0.03 & 5.51 $\pm$ 0.22 \\
    \cmidrule(lr){2-7}
                   & Original forecast & 2.33 $\pm$ 0.97 & 0.54 $\pm$ 0.21 & 0.92 $\pm$ 0.33 & 0.45 $\pm$ 0.20 & 5.42 $\pm$ 0.52 \\
                   & 	$R^2_\star=0.005$ & 1.44 $\pm$ 1.19 & 0.34 $\pm$ 0.26 & 0.58 $\pm$ 0.44 & 0.28 $\pm$ 0.23 & 5.98 $\pm$ 0.96 \\
                   & 	$R^2_\star=0.01$ & 1.54 $\pm$ 1.20 & 0.36 $\pm$ 0.27 & 0.62 $\pm$ 0.44 & 0.30 $\pm$ 0.24 & 5.91 $\pm$ 0.97 \\
                   & 	$R^2_\star=0.025$ & 1.86 $\pm$ 1.22 & 0.44 $\pm$ 0.27 & 0.74 $\pm$ 0.45 & 0.36 $\pm$ 0.25 & 5.76 $\pm$ 0.90 \\
                   & 	$R^2_\star=0.05$ & 2.42 $\pm$ 1.14 & 0.56 $\pm$ 0.25 & 0.96 $\pm$ 0.40 & 0.47 $\pm$ 0.24 & 5.46 $\pm$ 0.63 \\
                   & 	$R^2_\star=0.1$ & 3.75 $\pm$ 0.83 & 0.83 $\pm$ 0.21 & 1.39 $\pm$ 0.29 & 0.77 $\pm$ 0.19 & 4.93 $\pm$ 0.37 \\
                   & 	$R^2_\star=0.2$ & 6.03 $\pm$ 0.69 & 1.23 $\pm$ 0.18 & 2.02 $\pm$ 0.24 & 1.17 $\pm$ 0.18 & 5.13 $\pm$ 0.40 \\
                   & 	$R^2_\star=0.4$ & 13.32 $\pm$ 1.48 & 2.46 $\pm$ 0.28 & 4.05 $\pm$ 0.52 & 2.71 $\pm$ 0.33 & 4.67 $\pm$ 0.28 \\
                   & 	$R^2_\star=0.8$ & 22.70 $\pm$ 1.44 & 3.28 $\pm$ 0.12 & 5.78 $\pm$ 0.37 & 3.89 $\pm$ 0.17 & 5.32 $\pm$ 0.22 \\
    \midrule
    \textbf{SAC}   & Base & 2.32 $\pm$ 0.02 & 0.45 $\pm$ 0.00 & 0.74 $\pm$ 0.01 & 0.46 $\pm$ 0.00 & 5.35 $\pm$ 0.01 \\
    \cmidrule(lr){2-7}
                   & Original forecast & 2.33 $\pm$ 0.02 & 0.49 $\pm$ 0.00 & 0.82 $\pm$ 0.01 & 0.46 $\pm$ 0.00 & 5.28 $\pm$ 0.01 \\
                   & 	$R^2_\star=0.005$ & 2.13 $\pm$ 0.03 & 0.45 $\pm$ 0.00 & 0.75 $\pm$ 0.01 & 0.42 $\pm$ 0.00 & 5.33 $\pm$ 0.01 \\
                   & 	$R^2_\star=0.01$ & 2.16 $\pm$ 0.03 & 0.46 $\pm$ 0.00 & 0.76 $\pm$ 0.01 & 0.42 $\pm$ 0.00 & 5.33 $\pm$ 0.01 \\
                   & 	$R^2_\star=0.025$ & 2.25 $\pm$ 0.02 & 0.47 $\pm$ 0.00 & 0.79 $\pm$ 0.01 & 0.44 $\pm$ 0.00 & 5.33 $\pm$ 0.01 \\
                   & 	$R^2_\star=0.05$ & 2.40 $\pm$ 0.02 & 0.50 $\pm$ 0.00 & 0.84 $\pm$ 0.01 & 0.47 $\pm$ 0.00 & 5.32 $\pm$ 0.01 \\
                   & 	$R^2_\star=0.1$ & 2.70 $\pm$ 0.02 & 0.56 $\pm$ 0.00 & 0.94 $\pm$ 0.01 & 0.53 $\pm$ 0.00 & 5.30 $\pm$ 0.00 \\
                   & 	$R^2_\star=0.2$ & 3.38 $\pm$ 0.02 & 0.69 $\pm$ 0.00 & 1.16 $\pm$ 0.01 & 0.66 $\pm$ 0.00 & 5.21 $\pm$ 0.01 \\
                   & 	$R^2_\star=0.4$ & 4.88 $\pm$ 0.04 & 0.98 $\pm$ 0.01 & 1.65 $\pm$ 0.02 & 0.99 $\pm$ 0.01 & 4.96 $\pm$ 0.02 \\
                   & 	$R^2_\star=0.8$ & 6.43 $\pm$ 0.06 & 1.21 $\pm$ 0.01 & 2.05 $\pm$ 0.02 & 1.29 $\pm$ 0.01 & 4.94 $\pm$ 0.02 \\
    \midrule
    \textbf{A2C}   & Base & 2.31 $\pm$ 0.03 & 0.45 $\pm$ 0.01 & 0.74 $\pm$ 0.01 & 0.46 $\pm$ 0.01 & 5.35 $\pm$ 0.01 \\
    \cmidrule(lr){2-7}
                   & Original forecast & 2.78 $\pm$ 0.03 & 0.61 $\pm$ 0.01 & 1.06 $\pm$ 0.02 & 0.54 $\pm$ 0.01 & 5.27 $\pm$ 0.05 \\
                   & 	$R^2_\star=0.005$ & 1.87 $\pm$ 0.05 & 0.43 $\pm$ 0.01 & 0.74 $\pm$ 0.02 & 0.33 $\pm$ 0.01 & 6.03 $\pm$ 0.03 \\
                   & 	$R^2_\star=0.01$ & 1.98 $\pm$ 0.03 & 0.45 $\pm$ 0.01 & 0.77 $\pm$ 0.01 & 0.35 $\pm$ 0.01 & 5.94 $\pm$ 0.03 \\
                   & 	$R^2_\star=0.025$ & 2.27 $\pm$ 0.05 & 0.50 $\pm$ 0.02 & 0.85 $\pm$ 0.03 & 0.41 $\pm$ 0.01 & 5.77 $\pm$ 0.10 \\
                   & 	$R^2_\star=0.05$ & 3.41 $\pm$ 0.06 & 0.71 $\pm$ 0.01 & 1.17 $\pm$ 0.02 & 0.61 $\pm$ 0.01 & 5.74 $\pm$ 0.07 \\
                   & 	$R^2_\star=0.1$ & 4.70 $\pm$ 0.04 & 0.95 $\pm$ 0.01 & 1.56 $\pm$ 0.01 & 0.84 $\pm$ 0.01 & 5.64 $\pm$ 0.06 \\
                   & 	$R^2_\star=0.2$ & 6.97 $\pm$ 0.08 & 1.36 $\pm$ 0.01 & 2.26 $\pm$ 0.01 & 1.27 $\pm$ 0.02 & 5.42 $\pm$ 0.05 \\
                   & 	$R^2_\star=0.4$ & 15.06 $\pm$ 0.15 & 2.63 $\pm$ 0.04 & 4.34 $\pm$ 0.06 & 2.89 $\pm$ 0.06 & 4.91 $\pm$ 0.10 \\
                   & 	$R^2_\star=0.8$ & 24.61 $\pm$ 0.13 & 3.18 $\pm$ 0.01 & 5.62 $\pm$ 0.03 & 3.42 $\pm$ 0.03 & 6.51 $\pm$ 0.05 \\
    \midrule
    \textbf{TD3}   & Base & 2.63 $\pm$ 0.84 & 0.51 $\pm$ 0.15 & 0.82 $\pm$ 0.23 & 0.50 $\pm$ 0.19 & 5.67 $\pm$ 0.63 \\
    \cmidrule(lr){2-7}
                   & Original forecast & 4.76 $\pm$ 1.78 & 0.90 $\pm$ 0.27 & 1.49 $\pm$ 0.47 & 0.75 $\pm$ 0.26 & 6.45 $\pm$ 1.20 \\
                   & 	$R^2_\star=0.005$ & 2.28 $\pm$ 2.10 & 0.41 $\pm$ 0.36 & 0.65 $\pm$ 0.60 & 0.33 $\pm$ 0.31 & 7.71 $\pm$ 1.28 \\
                   & 	$R^2_\star=0.01$ & 5.43 $\pm$ 2.35 & 1.00 $\pm$ 0.46 & 1.56 $\pm$ 0.73 & 0.85 $\pm$ 0.49 & 6.90 $\pm$ 1.09 \\
                   & 	$R^2_\star=0.025$ & 5.28 $\pm$ 4.02 & 0.92 $\pm$ 0.75 & 1.51 $\pm$ 1.17 & 0.78 $\pm$ 0.55 & 7.35 $\pm$ 1.21 \\
                   & 	$R^2_\star=0.05$ & 6.48 $\pm$ 1.95 & 1.14 $\pm$ 0.32 & 1.79 $\pm$ 0.49 & 0.90 $\pm$ 0.25 & 7.17 $\pm$ 0.26 \\
                   & 	$R^2_\star=0.1$ & 9.33 $\pm$ 5.16 & 1.53 $\pm$ 0.75 & 2.45 $\pm$ 1.36 & 1.35 $\pm$ 0.70 & 6.42 $\pm$ 0.99 \\
                   & 	$R^2_\star=0.2$ & 12.19 $\pm$ 1.62 & 1.86 $\pm$ 0.28 & 2.91 $\pm$ 0.63 & 1.63 $\pm$ 0.22 & 7.17 $\pm$ 0.32 \\
                   & 	$R^2_\star=0.4$ & 28.96 $\pm$ 9.42 & 3.05 $\pm$ 0.45 & 4.78 $\pm$ 1.08 & 4.31 $\pm$ 1.47 & 6.10 $\pm$ 1.02 \\
                   & 	$R^2_\star=0.8$ & 66.77 $\pm$ 7.57 & 4.81 $\pm$ 0.65 & 9.90 $\pm$ 2.07 & 11.87 $\pm$ 4.09 & 4.73 $\pm$ 1.34 \\
    \midrule
    \textbf{DDPG}  & Base & 2.62 $\pm$ 1.77 & 0.51 $\pm$ 0.31 & 0.81 $\pm$ 0.51 & 0.50 $\pm$ 0.30 & 5.70 $\pm$ 0.83 \\
    \cmidrule(lr){2-7}
                   & Original forecast & 4.96 $\pm$ 2.60 & 1.00 $\pm$ 0.44 & 1.63 $\pm$ 0.80 & 0.79 $\pm$ 0.38 & 6.31 $\pm$ 0.83 \\
                   & 	$R^2_\star=0.005$ & 3.75 $\pm$ 2.66 & 0.73 $\pm$ 0.42 & 1.18 $\pm$ 0.71 & 0.55 $\pm$ 0.35 & 6.69 $\pm$ 0.45 \\
                   & 	$R^2_\star=0.01$ & 4.45 $\pm$ 3.90 & 0.83 $\pm$ 0.54 & 1.38 $\pm$ 1.00 & 0.66 $\pm$ 0.52 & 6.54 $\pm$ 0.48 \\
                   & 	$R^2_\star=0.025$ & 6.12 $\pm$ 3.74 & 1.10 $\pm$ 0.49 & 1.74 $\pm$ 0.81 & 0.90 $\pm$ 0.45 & 6.41 $\pm$ 0.84 \\
                   & 	$R^2_\star=0.05$ & 7.29 $\pm$ 4.62 & 1.27 $\pm$ 0.63 & 2.01 $\pm$ 1.11 & 1.01 $\pm$ 0.55 & 6.80 $\pm$ 0.69 \\
                   & 	$R^2_\star=0.1$ & 8.65 $\pm$ 1.71 & 1.50 $\pm$ 0.29 & 2.32 $\pm$ 0.56 & 1.28 $\pm$ 0.26 & 6.71 $\pm$ 1.17 \\
                   & 	$R^2_\star=0.2$ & 12.75 $\pm$ 2.69 & 1.96 $\pm$ 0.22 & 3.12 $\pm$ 0.42 & 1.92 $\pm$ 0.44 & 6.45 $\pm$ 1.04 \\
                   & 	$R^2_\star=0.4$ & 27.15 $\pm$ 3.87 & 3.00 $\pm$ 0.37 & 4.67 $\pm$ 0.80 & 3.97 $\pm$ 1.17 & 6.45 $\pm$ 1.44 \\
                   & 	$R^2_\star=0.8$ & 54.27 $\pm$ 12.83 & 4.14 $\pm$ 0.57 & 8.26 $\pm$ 1.79 & 8.20 $\pm$ 2.72 & 5.60 $\pm$ 1.19 \\
    \bottomrule
  \end{tabular}
  }%
\end{table}

\begin{table}[ht]
  \caption{$\fpilotabs$ performance with controlled forecasts on FX, test year 2024,
at target skill levels from $R^2_\star=0.005$ to $R^2_\star=0.8$
(Section~\ref{sec:cheating}). Values are mean $\pm$ one standard deviation across
5 seeds. Base is the frozen pre-trained policy and does not depend on $R^2_\star$;
Original forecast is the unmodified forecast ($c_{t,i}=1$).}
  \vspace{2ex}
  
  \label{tab:r2-subset-fx-2024}
  \centering
  \resizebox{\textwidth}{!}{%
  \begin{tabular}{ll ccccc}
    \toprule
    Algorithm & 	$R^2_\star$ & Total return (\%) $\uparrow$ & Sharpe $\uparrow$ & Sortino $\uparrow$ & Calmar $\uparrow$ & MDD (\%) $\downarrow$ \\
    \midrule
    \textbf{PPO}   & Base & -5.82 $\pm$ 0.10 & -1.36 $\pm$ 0.01 & -1.87 $\pm$ 0.02 & -0.90 $\pm$ 0.01 & 6.53 $\pm$ 0.10 \\
    \cmidrule(lr){2-7}
                   & Original forecast & -4.60 $\pm$ 0.24 & -1.24 $\pm$ 0.05 & -1.67 $\pm$ 0.08 & -0.95 $\pm$ 0.04 & 4.85 $\pm$ 0.34 \\
                   & 	$R^2_\star=0.005$ & -4.67 $\pm$ 0.20 & -1.25 $\pm$ 0.07 & -1.69 $\pm$ 0.10 & -0.93 $\pm$ 0.05 & 4.99 $\pm$ 0.44 \\
                   & 	$R^2_\star=0.01$ & -4.61 $\pm$ 0.19 & -1.23 $\pm$ 0.07 & -1.67 $\pm$ 0.09 & -0.92 $\pm$ 0.06 & 4.99 $\pm$ 0.47 \\
                   & 	$R^2_\star=0.025$ & -4.46 $\pm$ 0.10 & -1.18 $\pm$ 0.05 & -1.61 $\pm$ 0.07 & -0.89 $\pm$ 0.07 & 5.05 $\pm$ 0.47 \\
                   & 	$R^2_\star=0.05$ & -4.16 $\pm$ 0.14 & -1.10 $\pm$ 0.05 & -1.51 $\pm$ 0.06 & -0.83 $\pm$ 0.08 & 5.05 $\pm$ 0.47 \\
                   & 	$R^2_\star=0.1$ & -3.75 $\pm$ 0.23 & -0.98 $\pm$ 0.09 & -1.36 $\pm$ 0.12 & -0.73 $\pm$ 0.07 & 5.16 $\pm$ 0.29 \\
                   & 	$R^2_\star=0.2$ & -2.77 $\pm$ 0.21 & -0.72 $\pm$ 0.04 & -1.01 $\pm$ 0.05 & -0.53 $\pm$ 0.04 & 5.20 $\pm$ 0.30 \\
                   & 	$R^2_\star=0.4$ & -0.93 $\pm$ 0.42 & -0.24 $\pm$ 0.11 & -0.34 $\pm$ 0.15 & -0.18 $\pm$ 0.08 & 4.91 $\pm$ 0.56 \\
                   & 	$R^2_\star=0.8$ & 7.83 $\pm$ 1.15 & 1.74 $\pm$ 0.25 & 2.62 $\pm$ 0.38 & 1.76 $\pm$ 0.35 & 4.40 $\pm$ 0.32 \\
    \midrule
    \textbf{SAC}   & Base & -5.74 $\pm$ 0.07 & -1.35 $\pm$ 0.01 & -1.87 $\pm$ 0.02 & -0.90 $\pm$ 0.01 & 6.49 $\pm$ 0.04 \\
    \cmidrule(lr){2-7}
                   & Original forecast & -5.53 $\pm$ 0.05 & -1.42 $\pm$ 0.01 & -1.89 $\pm$ 0.01 & -1.00 $\pm$ 0.01 & 5.60 $\pm$ 0.01 \\
                   & 	$R^2_\star=0.005$ & -5.46 $\pm$ 0.05 & -1.41 $\pm$ 0.01 & -1.86 $\pm$ 0.01 & -0.98 $\pm$ 0.01 & 5.64 $\pm$ 0.01 \\
                   & 	$R^2_\star=0.01$ & -5.40 $\pm$ 0.05 & -1.39 $\pm$ 0.01 & -1.84 $\pm$ 0.01 & -0.97 $\pm$ 0.01 & 5.64 $\pm$ 0.02 \\
                   & 	$R^2_\star=0.025$ & -5.24 $\pm$ 0.05 & -1.35 $\pm$ 0.01 & -1.78 $\pm$ 0.01 & -0.94 $\pm$ 0.01 & 5.63 $\pm$ 0.01 \\
                   & 	$R^2_\star=0.05$ & -4.97 $\pm$ 0.05 & -1.28 $\pm$ 0.01 & -1.69 $\pm$ 0.01 & -0.90 $\pm$ 0.01 & 5.60 $\pm$ 0.01 \\
                   & 	$R^2_\star=0.1$ & -4.46 $\pm$ 0.05 & -1.14 $\pm$ 0.01 & -1.51 $\pm$ 0.01 & -0.81 $\pm$ 0.01 & 5.57 $\pm$ 0.01 \\
                   & 	$R^2_\star=0.2$ & -3.56 $\pm$ 0.04 & -0.90 $\pm$ 0.01 & -1.20 $\pm$ 0.01 & -0.65 $\pm$ 0.01 & 5.48 $\pm$ 0.01 \\
                   & 	$R^2_\star=0.4$ & -2.25 $\pm$ 0.03 & -0.55 $\pm$ 0.01 & -0.75 $\pm$ 0.01 & -0.42 $\pm$ 0.01 & 5.28 $\pm$ 0.01 \\
                   & 	$R^2_\star=0.8$ & -0.88 $\pm$ 0.03 & -0.20 $\pm$ 0.01 & -0.27 $\pm$ 0.01 & -0.16 $\pm$ 0.00 & 5.12 $\pm$ 0.01 \\
    \midrule
    \textbf{A2C}   & Base & -5.78 $\pm$ 0.00 & -1.36 $\pm$ 0.00 & -1.88 $\pm$ 0.00 & -0.90 $\pm$ 0.00 & 6.51 $\pm$ 0.01 \\
    \cmidrule(lr){2-7}
                   & Original forecast & -3.79 $\pm$ 0.24 & -1.05 $\pm$ 0.05 & -1.43 $\pm$ 0.07 & -0.84 $\pm$ 0.02 & 4.51 $\pm$ 0.19 \\
                   & 	$R^2_\star=0.005$ & -4.37 $\pm$ 0.10 & -1.19 $\pm$ 0.02 & -1.60 $\pm$ 0.03 & -0.89 $\pm$ 0.01 & 4.93 $\pm$ 0.08 \\
                   & 	$R^2_\star=0.01$ & -4.36 $\pm$ 0.13 & -1.18 $\pm$ 0.03 & -1.59 $\pm$ 0.04 & -0.88 $\pm$ 0.01 & 4.99 $\pm$ 0.15 \\
                   & 	$R^2_\star=0.025$ & -4.35 $\pm$ 0.16 & -1.16 $\pm$ 0.04 & -1.59 $\pm$ 0.06 & -0.85 $\pm$ 0.01 & 5.17 $\pm$ 0.15 \\
                   & 	$R^2_\star=0.05$ & -3.91 $\pm$ 0.11 & -1.03 $\pm$ 0.03 & -1.44 $\pm$ 0.04 & -0.75 $\pm$ 0.02 & 5.20 $\pm$ 0.04 \\
                   & 	$R^2_\star=0.1$ & -3.39 $\pm$ 0.10 & -0.89 $\pm$ 0.03 & -1.24 $\pm$ 0.04 & -0.65 $\pm$ 0.02 & 5.20 $\pm$ 0.04 \\
                   & 	$R^2_\star=0.2$ & -2.53 $\pm$ 0.08 & -0.67 $\pm$ 0.01 & -0.94 $\pm$ 0.01 & -0.48 $\pm$ 0.01 & 5.17 $\pm$ 0.09 \\
                   & 	$R^2_\star=0.4$ & -0.44 $\pm$ 0.20 & -0.12 $\pm$ 0.06 & -0.18 $\pm$ 0.09 & -0.09 $\pm$ 0.05 & 4.12 $\pm$ 0.15 \\
                   & 	$R^2_\star=0.8$ & 7.38 $\pm$ 0.24 & 1.69 $\pm$ 0.05 & 2.56 $\pm$ 0.08 & 1.68 $\pm$ 0.06 & 4.28 $\pm$ 0.03 \\
    \midrule
    \textbf{TD3}   & Base & -7.36 $\pm$ 1.37 & -1.56 $\pm$ 0.24 & -2.24 $\pm$ 0.27 & -0.99 $\pm$ 0.05 & 7.43 $\pm$ 1.21 \\
    \cmidrule(lr){2-7}
                   & Original forecast & -3.23 $\pm$ 1.51 & -0.83 $\pm$ 0.33 & -1.18 $\pm$ 0.39 & -0.67 $\pm$ 0.14 & 4.55 $\pm$ 1.44 \\
                   & 	$R^2_\star=0.005$ & -4.84 $\pm$ 3.54 & -1.08 $\pm$ 0.52 & -1.47 $\pm$ 0.59 & -0.79 $\pm$ 0.15 & 5.64 $\pm$ 3.15 \\
                   & 	$R^2_\star=0.01$ & -3.92 $\pm$ 2.46 & -0.93 $\pm$ 0.39 & -1.27 $\pm$ 0.44 & -0.73 $\pm$ 0.15 & 4.93 $\pm$ 2.10 \\
                   & 	$R^2_\star=0.025$ & -4.48 $\pm$ 3.66 & -1.00 $\pm$ 0.55 & -1.36 $\pm$ 0.66 & -0.74 $\pm$ 0.19 & 5.43 $\pm$ 3.08 \\
                   & 	$R^2_\star=0.05$ & -2.81 $\pm$ 0.97 & -0.76 $\pm$ 0.22 & -1.09 $\pm$ 0.25 & -0.66 $\pm$ 0.14 & 4.08 $\pm$ 0.71 \\
                   & 	$R^2_\star=0.1$ & -0.73 $\pm$ 0.98 & -0.22 $\pm$ 0.29 & -0.32 $\pm$ 0.43 & -0.18 $\pm$ 0.30 & 3.34 $\pm$ 0.71 \\
                   & 	$R^2_\star=0.2$ & 3.03 $\pm$ 2.46 & 0.66 $\pm$ 0.30 & 0.90 $\pm$ 0.36 & 0.66 $\pm$ 0.18 & 4.41 $\pm$ 3.02 \\
                   & 	$R^2_\star=0.4$ & 12.90 $\pm$ 4.91 & 2.17 $\pm$ 1.06 & 3.33 $\pm$ 1.91 & 4.03 $\pm$ 3.79 & 5.09 $\pm$ 2.97 \\
                   & 	$R^2_\star=0.8$ & 17.26 $\pm$ 7.15 & 2.47 $\pm$ 0.95 & 3.78 $\pm$ 1.76 & 4.86 $\pm$ 3.38 & 4.61 $\pm$ 2.73 \\
    \midrule
    \textbf{DDPG}  & Base & -6.78 $\pm$ 1.31 & -1.62 $\pm$ 0.28 & -2.33 $\pm$ 0.49 & -1.01 $\pm$ 0.03 & 6.68 $\pm$ 1.20 \\
    \cmidrule(lr){2-7}
                   & Original forecast & -4.36 $\pm$ 1.58 & -1.25 $\pm$ 0.48 & -1.61 $\pm$ 0.48 & -0.87 $\pm$ 0.15 & 4.93 $\pm$ 1.48 \\
                   & 	$R^2_\star=0.005$ & -4.61 $\pm$ 3.12 & -1.22 $\pm$ 0.73 & -1.51 $\pm$ 0.65 & -0.78 $\pm$ 0.22 & 5.58 $\pm$ 2.66 \\
                   & 	$R^2_\star=0.01$ & -3.83 $\pm$ 2.18 & -1.09 $\pm$ 0.67 & -1.44 $\pm$ 0.73 & -0.74 $\pm$ 0.26 & 4.78 $\pm$ 1.40 \\
                   & 	$R^2_\star=0.025$ & -3.32 $\pm$ 1.38 & -0.90 $\pm$ 0.30 & -1.25 $\pm$ 0.32 & -0.72 $\pm$ 0.17 & 4.51 $\pm$ 1.27 \\
                   & 	$R^2_\star=0.05$ & -3.04 $\pm$ 1.34 & -0.76 $\pm$ 0.27 & -1.10 $\pm$ 0.31 & -0.64 $\pm$ 0.07 & 4.60 $\pm$ 1.54 \\
                   & 	$R^2_\star=0.1$ & -0.01 $\pm$ 1.49 & 0.03 $\pm$ 0.44 & 0.06 $\pm$ 0.74 & 0.14 $\pm$ 0.67 & 3.18 $\pm$ 0.78 \\
                   & 	$R^2_\star=0.2$ & 4.83 $\pm$ 2.24 & 1.21 $\pm$ 0.89 & 1.80 $\pm$ 1.49 & 1.49 $\pm$ 1.90 & 5.39 $\pm$ 2.61 \\
                   & 	$R^2_\star=0.4$ & 15.53 $\pm$ 7.56 & 2.45 $\pm$ 1.22 & 3.54 $\pm$ 1.87 & 4.24 $\pm$ 2.98 & 5.03 $\pm$ 2.88 \\
                   & 	$R^2_\star=0.8$ & 22.37 $\pm$ 7.44 & 3.69 $\pm$ 0.60 & 5.70 $\pm$ 0.97 & 9.17 $\pm$ 3.72 & 2.59 $\pm$ 1.30 \\
    \bottomrule
  \end{tabular}
  }%
\end{table}

\begin{table}[ht]
  \caption{$\fpilotabs$ performance with controlled forecasts on FX, test year 2025,
at target skill levels from $R^2_\star=0.005$ to $R^2_\star=0.8$
(Section~\ref{sec:cheating}). Values are mean $\pm$ one standard deviation across
5 seeds. Base is the frozen pre-trained policy and does not depend on $R^2_\star$;
Original forecast is the unmodified forecast ($c_{t,i}=1$).}
  \vspace{2ex}
  
  \label{tab:r2-subset-fx-2025}
  \centering
  \resizebox{\textwidth}{!}{%
  \begin{tabular}{ll ccccc}
    \toprule
    Algorithm & 	$R^2_\star$ & Total return (\%) $\uparrow$ & Sharpe $\uparrow$ & Sortino $\uparrow$ & Calmar $\uparrow$ & MDD (\%) $\downarrow$ \\
    \midrule
    \textbf{PPO}   & Base & 7.00 $\pm$ 0.26 & 1.40 $\pm$ 0.04 & 2.41 $\pm$ 0.08 & 2.70 $\pm$ 0.11 & 2.55 $\pm$ 0.04 \\
    \cmidrule(lr){2-7}
                   & Original forecast & 6.14 $\pm$ 0.78 & 1.54 $\pm$ 0.15 & 3.05 $\pm$ 0.29 & 3.31 $\pm$ 0.85 & 1.91 $\pm$ 0.45 \\
                   & 	$R^2_\star=0.005$ & 6.35 $\pm$ 0.80 & 1.54 $\pm$ 0.16 & 3.02 $\pm$ 0.32 & 3.27 $\pm$ 0.86 & 1.99 $\pm$ 0.43 \\
                   & 	$R^2_\star=0.01$ & 6.53 $\pm$ 0.79 & 1.56 $\pm$ 0.15 & 3.08 $\pm$ 0.30 & 3.28 $\pm$ 0.83 & 2.03 $\pm$ 0.41 \\
                   & 	$R^2_\star=0.025$ & 7.21 $\pm$ 0.75 & 1.65 $\pm$ 0.13 & 3.30 $\pm$ 0.27 & 3.55 $\pm$ 0.86 & 2.06 $\pm$ 0.41 \\
                   & 	$R^2_\star=0.05$ & 8.28 $\pm$ 0.70 & 1.78 $\pm$ 0.14 & 3.63 $\pm$ 0.30 & 4.03 $\pm$ 0.96 & 2.07 $\pm$ 0.40 \\
                   & 	$R^2_\star=0.1$ & 10.20 $\pm$ 0.66 & 1.94 $\pm$ 0.12 & 4.04 $\pm$ 0.30 & 4.44 $\pm$ 0.72 & 2.25 $\pm$ 0.27 \\
                   & 	$R^2_\star=0.2$ & 13.83 $\pm$ 0.52 & 2.27 $\pm$ 0.07 & 4.90 $\pm$ 0.27 & 5.63 $\pm$ 0.58 & 2.35 $\pm$ 0.20 \\
                   & 	$R^2_\star=0.4$ & 18.82 $\pm$ 1.01 & 2.64 $\pm$ 0.14 & 5.59 $\pm$ 0.46 & 6.91 $\pm$ 0.86 & 2.56 $\pm$ 0.29 \\
                   & 	$R^2_\star=0.8$ & 21.25 $\pm$ 0.47 & 2.71 $\pm$ 0.05 & 5.43 $\pm$ 0.15 & 6.10 $\pm$ 0.20 & 3.20 $\pm$ 0.06 \\
    \midrule
    \textbf{SAC}   & Base & 7.02 $\pm$ 0.07 & 1.39 $\pm$ 0.00 & 2.39 $\pm$ 0.01 & 2.70 $\pm$ 0.01 & 2.56 $\pm$ 0.03 \\
    \cmidrule(lr){2-7}
                   & Original forecast & 7.06 $\pm$ 0.04 & 1.54 $\pm$ 0.01 & 2.80 $\pm$ 0.02 & 3.07 $\pm$ 0.01 & 2.25 $\pm$ 0.00 \\
                   & 	$R^2_\star=0.005$ & 7.00 $\pm$ 0.05 & 1.52 $\pm$ 0.01 & 2.75 $\pm$ 0.03 & 3.01 $\pm$ 0.01 & 2.28 $\pm$ 0.00 \\
                   & 	$R^2_\star=0.01$ & 7.05 $\pm$ 0.05 & 1.53 $\pm$ 0.01 & 2.77 $\pm$ 0.03 & 3.03 $\pm$ 0.02 & 2.28 $\pm$ 0.00 \\
                   & 	$R^2_\star=0.025$ & 7.20 $\pm$ 0.04 & 1.55 $\pm$ 0.01 & 2.82 $\pm$ 0.02 & 3.09 $\pm$ 0.01 & 2.28 $\pm$ 0.00 \\
                   & 	$R^2_\star=0.05$ & 7.47 $\pm$ 0.05 & 1.58 $\pm$ 0.01 & 2.90 $\pm$ 0.03 & 3.20 $\pm$ 0.02 & 2.28 $\pm$ 0.00 \\
                   & 	$R^2_\star=0.1$ & 8.00 $\pm$ 0.06 & 1.65 $\pm$ 0.01 & 3.03 $\pm$ 0.03 & 3.40 $\pm$ 0.02 & 2.29 $\pm$ 0.00 \\
                   & 	$R^2_\star=0.2$ & 9.11 $\pm$ 0.07 & 1.76 $\pm$ 0.01 & 3.24 $\pm$ 0.02 & 3.79 $\pm$ 0.02 & 2.34 $\pm$ 0.01 \\
                   & 	$R^2_\star=0.4$ & 10.79 $\pm$ 0.07 & 1.93 $\pm$ 0.01 & 3.54 $\pm$ 0.02 & 4.51 $\pm$ 0.02 & 2.31 $\pm$ 0.00 \\
                   & 	$R^2_\star=0.8$ & 12.56 $\pm$ 0.07 & 2.14 $\pm$ 0.01 & 3.94 $\pm$ 0.02 & 5.39 $\pm$ 0.02 & 2.22 $\pm$ 0.00 \\
    \midrule
    \textbf{A2C}   & Base & 7.06 $\pm$ 0.02 & 1.39 $\pm$ 0.00 & 2.39 $\pm$ 0.01 & 2.71 $\pm$ 0.01 & 2.57 $\pm$ 0.01 \\
    \cmidrule(lr){2-7}
                   & Original forecast & 5.34 $\pm$ 0.08 & 1.43 $\pm$ 0.01 & 2.93 $\pm$ 0.02 & 2.66 $\pm$ 0.04 & 1.98 $\pm$ 0.01 \\
                   & 	$R^2_\star=0.005$ & 5.67 $\pm$ 0.09 & 1.43 $\pm$ 0.01 & 2.95 $\pm$ 0.03 & 2.67 $\pm$ 0.04 & 2.10 $\pm$ 0.00 \\
                   & 	$R^2_\star=0.01$ & 5.92 $\pm$ 0.11 & 1.47 $\pm$ 0.02 & 3.05 $\pm$ 0.04 & 2.77 $\pm$ 0.05 & 2.11 $\pm$ 0.00 \\
                   & 	$R^2_\star=0.025$ & 6.63 $\pm$ 0.08 & 1.58 $\pm$ 0.01 & 3.29 $\pm$ 0.01 & 3.06 $\pm$ 0.04 & 2.13 $\pm$ 0.01 \\
                   & 	$R^2_\star=0.05$ & 7.58 $\pm$ 0.07 & 1.69 $\pm$ 0.01 & 3.50 $\pm$ 0.04 & 3.33 $\pm$ 0.04 & 2.22 $\pm$ 0.03 \\
                   & 	$R^2_\star=0.1$ & 9.54 $\pm$ 0.05 & 1.91 $\pm$ 0.01 & 3.92 $\pm$ 0.01 & 4.46 $\pm$ 0.07 & 2.07 $\pm$ 0.03 \\
                   & 	$R^2_\star=0.2$ & 14.77 $\pm$ 0.32 & 2.36 $\pm$ 0.03 & 5.19 $\pm$ 0.06 & 6.41 $\pm$ 0.16 & 2.18 $\pm$ 0.07 \\
                   & 	$R^2_\star=0.4$ & 23.23 $\pm$ 0.09 & 3.09 $\pm$ 0.01 & 7.52 $\pm$ 0.03 & 9.05 $\pm$ 0.08 & 2.33 $\pm$ 0.02 \\
                   & 	$R^2_\star=0.8$ & 26.68 $\pm$ 0.04 & 3.19 $\pm$ 0.01 & 7.16 $\pm$ 0.02 & 7.69 $\pm$ 0.07 & 3.11 $\pm$ 0.03 \\
    \midrule
    \textbf{TD3}   & Base & 6.35 $\pm$ 1.05 & 1.33 $\pm$ 0.19 & 2.32 $\pm$ 0.29 & 2.64 $\pm$ 0.74 & 2.47 $\pm$ 0.47 \\
    \cmidrule(lr){2-7}
                   & Original forecast & 8.74 $\pm$ 0.86 & 1.53 $\pm$ 0.09 & 2.86 $\pm$ 0.18 & 3.55 $\pm$ 0.18 & 2.40 $\pm$ 0.22 \\
                   & 	$R^2_\star=0.005$ & 6.63 $\pm$ 2.13 & 1.26 $\pm$ 0.30 & 2.35 $\pm$ 0.56 & 2.46 $\pm$ 1.06 & 2.87 $\pm$ 0.70 \\
                   & 	$R^2_\star=0.01$ & 6.81 $\pm$ 2.35 & 1.27 $\pm$ 0.35 & 2.39 $\pm$ 0.65 & 2.45 $\pm$ 1.14 & 2.94 $\pm$ 0.60 \\
                   & 	$R^2_\star=0.025$ & 6.37 $\pm$ 1.72 & 1.19 $\pm$ 0.25 & 2.34 $\pm$ 0.55 & 2.25 $\pm$ 0.80 & 2.92 $\pm$ 0.45 \\
                   & 	$R^2_\star=0.05$ & 9.35 $\pm$ 2.24 & 1.56 $\pm$ 0.26 & 3.13 $\pm$ 0.54 & 3.66 $\pm$ 1.15 & 2.56 $\pm$ 0.37 \\
                   & 	$R^2_\star=0.1$ & 11.26 $\pm$ 0.96 & 1.79 $\pm$ 0.21 & 3.54 $\pm$ 0.42 & 4.52 $\pm$ 0.99 & 2.46 $\pm$ 0.34 \\
                   & 	$R^2_\star=0.2$ & 15.73 $\pm$ 2.72 & 2.14 $\pm$ 0.19 & 4.56 $\pm$ 0.57 & 5.77 $\pm$ 1.36 & 2.62 $\pm$ 0.35 \\
                   & 	$R^2_\star=0.4$ & 24.97 $\pm$ 6.59 & 2.70 $\pm$ 0.44 & 5.61 $\pm$ 1.70 & 8.09 $\pm$ 2.43 & 2.87 $\pm$ 0.47 \\
                   & 	$R^2_\star=0.8$ & 35.56 $\pm$ 11.51 & 3.41 $\pm$ 0.89 & 7.49 $\pm$ 3.59 & 10.58 $\pm$ 4.76 & 3.11 $\pm$ 0.71 \\
    \midrule
    \textbf{DDPG}  & Base & 3.52 $\pm$ 1.85 & 0.79 $\pm$ 0.39 & 1.35 $\pm$ 0.67 & 1.33 $\pm$ 0.97 & 3.18 $\pm$ 1.10 \\
    \cmidrule(lr){2-7}
                   & Original forecast & 3.84 $\pm$ 0.91 & 0.82 $\pm$ 0.17 & 1.51 $\pm$ 0.28 & 1.07 $\pm$ 0.39 & 3.76 $\pm$ 0.62 \\
                   & 	$R^2_\star=0.005$ & 2.82 $\pm$ 1.69 & 0.65 $\pm$ 0.39 & 1.17 $\pm$ 0.70 & 0.91 $\pm$ 0.67 & 3.99 $\pm$ 1.87 \\
                   & 	$R^2_\star=0.01$ & 2.53 $\pm$ 1.10 & 0.57 $\pm$ 0.23 & 1.05 $\pm$ 0.43 & 0.70 $\pm$ 0.30 & 3.78 $\pm$ 0.74 \\
                   & 	$R^2_\star=0.025$ & 3.19 $\pm$ 1.43 & 0.69 $\pm$ 0.28 & 1.27 $\pm$ 0.55 & 0.96 $\pm$ 0.55 & 3.60 $\pm$ 0.85 \\
                   & $R^2_\star=0.05$ & 4.61 $\pm$ 1.41 & 0.97 $\pm$ 0.24 & 1.84 $\pm$ 0.49 & 1.65 $\pm$ 0.56 & 2.87 $\pm$ 0.53 \\
                   & $R^2_\star=0.1$ & 8.71 $\pm$ 3.17 & 1.54 $\pm$ 0.51 & 3.11 $\pm$ 1.09 & 3.19 $\pm$ 1.79 & 3.05 $\pm$ 1.06 \\
                   & $R^2_\star=0.2$ & 14.90 $\pm$ 5.38 & 2.07 $\pm$ 0.45 & 4.49 $\pm$ 1.04 & 4.70 $\pm$ 2.41 & 3.27 $\pm$ 1.03 \\
                   & $R^2_\star=0.4$ & 21.18 $\pm$ 4.11 & 2.53 $\pm$ 0.41 & 5.45 $\pm$ 1.18 & 6.60 $\pm$ 2.33 & 3.13 $\pm$ 0.77 \\
                   & $R^2_\star=0.8$ & 30.65 $\pm$ 7.20 & 3.26 $\pm$ 0.44 & 6.51 $\pm$ 1.01 & 10.41 $\pm$ 3.97 & 2.79 $\pm$ 0.74 \\
    \bottomrule
  \end{tabular}
  }%
\end{table}

\FloatBarrier
\subsection{Sensitivity to step size and adaptation epochs}
\label{app:alpha}
\begin{figure}[h]
\centering
\includegraphics[width=\linewidth]{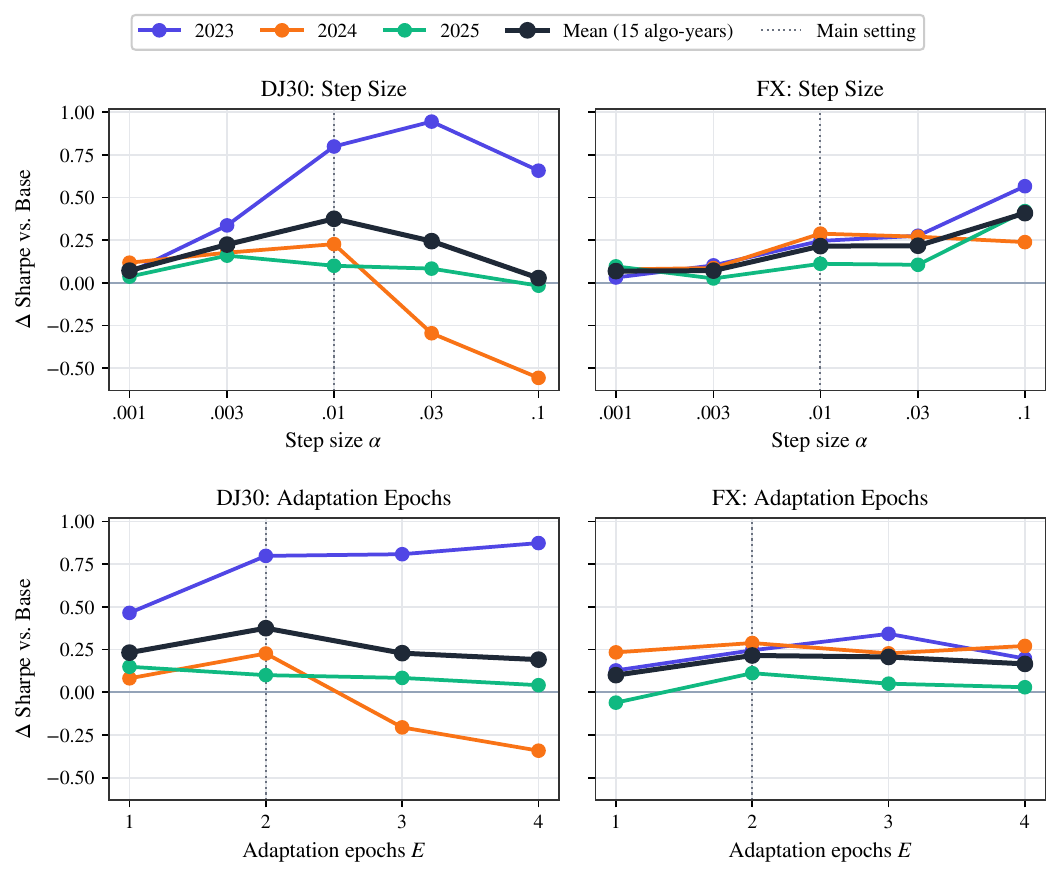}
\caption{Post-hoc sensitivity of $\fpilotabs$ to the step size $\alpha$ (top;
$E = 2$) and the number of adaptation epochs $E$ (bottom; $\alpha = 0.01$).
Colored lines show the mean $\Delta$Sharpe over the five algorithms in each
test year; the black line is the mean over all 15 algorithm--years in each
market. The dotted line marks the setting used in all main results. Pooled
forecaster $R^2$ for 2023/2024/2025 is 0.027/0.005/0.014 on DJ30 and
0.039/0.029/0.019 on FX (Table~\ref{tab:r2_forecaster}). Lines show means
only; seed-level variability at the main setting is reported in
Tables~\ref{tab:dj30_final} and~\ref{tab:fx_final}.}
\label{fig:sensitivity}
\end{figure}

We vary the step size $\alpha \in \{0.001, 0.003, 0.01, 0.03, 0.1\}$ (with
$E = 2$) and the number of adaptation epochs $E \in \{1, 2, 3, 4\}$ (with
$\alpha = 0.01$), holding all other settings, including the selected $\lambda$
values and seeds, at those of the main experiments. Both analyses are post-hoc
and were not used to select any setting. Table~\ref{tab:sensitivity} reports
averages over the 15 algorithm--years per market, and
Figure~\ref{fig:sensitivity} breaks $\Delta$Sharpe down by test year.

\begin{table}[h]
\centering
\small
\caption{Post-hoc sensitivity of $\fpilotabs$ to the step size $\alpha$ (with $E = 2$) and the number of adaptation epochs $E$ (with $\alpha = 0.01$). All other settings match the main experiments. Values are changes relative to Base, averaged over the 15 algorithm--years per market; the subscript is the seed-level standard deviation of the change, computed from paired per-seed differences ($\fpilotabs$ minus Base on the same seed) within each algorithm--year and pooled over the 15 algorithm--years. Sharpe$\uparrow$ counts algorithm--years where Sharpe improves, and Best counts, within each sweep, the algorithm--years in which that setting gives the highest $\Delta$Sharpe among the values tested. The main experiments use $\alpha = 0.01$ and $E = 2$; per-year results are shown in Figure~\ref{fig:sensitivity}.}
\vspace{2ex}
\label{tab:sensitivity}
\setlength{\tabcolsep}{3.5pt}
\resizebox{\linewidth}{!}{%
\begin{tabular}{l *{5}{r@{}l} rr}
\toprule
Setting & \multicolumn{2}{c}{$\Delta$Ret.\ (pp)} & \multicolumn{2}{c}{$\Delta$Sharpe} & \multicolumn{2}{c}{$\Delta$Sortino}
        & \multicolumn{2}{c}{$\Delta$Calmar} & \multicolumn{2}{c}{$\Delta$MDD (pp)$\downarrow$} & Sharpe$\uparrow$ & Best \\
\midrule
\multicolumn{13}{l}{\textit{DJ30}} \\
$\alpha = 0.001$ & $+2.08$ & ${}_{\pm 2.80}$ & $+0.07$ & ${}_{\pm 0.16}$ & $+0.13$ & ${}_{\pm 0.25}$ & $+0.19$ & ${}_{\pm 0.24}$ & $+0.31$ & ${}_{\pm 0.85}$ & 12/15 & 0/15 \\
$\alpha = 0.003$ & $+8.89$ & ${}_{\pm 6.93}$ & $+0.22$ & ${}_{\pm 0.26}$ & $+0.39$ & ${}_{\pm 0.39}$ & $+0.50$ & ${}_{\pm 0.55}$ & $+1.43$ & ${}_{\pm 2.10}$ & 15/15 & 3/15 \\
$\alpha = 0.01$ & $+14.92$ & ${}_{\pm 6.15}$ & $+0.38$ & ${}_{\pm 0.30}$ & $+0.70$ & ${}_{\pm 0.51}$ & $+0.90$ & ${}_{\pm 0.66}$ & $+1.57$ & ${}_{\pm 2.42}$ & 13/15 & 7/15 \\
$\alpha = 0.03$ & $+14.20$ & ${}_{\pm 9.22}$ & $+0.24$ & ${}_{\pm 0.37}$ & $+0.54$ & ${}_{\pm 0.62}$ & $+0.71$ & ${}_{\pm 0.71}$ & $+3.26$ & ${}_{\pm 2.78}$ & 10/15 & 1/15 \\
$\alpha = 0.1$ & $+8.60$ & ${}_{\pm 5.41}$ & $+0.03$ & ${}_{\pm 0.30}$ & $+0.18$ & ${}_{\pm 0.47}$ & $+0.25$ & ${}_{\pm 0.54}$ & $+3.37$ & ${}_{\pm 2.28}$ & 7/15 & 4/15 \\
\cmidrule(l){1-13}
$E = 1$ & $+9.75$ & ${}_{\pm 7.01}$ & $+0.23$ & ${}_{\pm 0.29}$ & $+0.43$ & ${}_{\pm 0.41}$ & $+0.62$ & ${}_{\pm 0.65}$ & $+1.44$ & ${}_{\pm 2.63}$ & 14/15 & 3/15 \\
$E = 2$ & $+14.92$ & ${}_{\pm 6.15}$ & $+0.38$ & ${}_{\pm 0.30}$ & $+0.70$ & ${}_{\pm 0.51}$ & $+0.90$ & ${}_{\pm 0.66}$ & $+1.57$ & ${}_{\pm 2.42}$ & 13/15 & 6/15 \\
$E = 3$ & $+13.70$ & ${}_{\pm 11.13}$ & $+0.23$ & ${}_{\pm 0.51}$ & $+0.51$ & ${}_{\pm 0.73}$ & $+0.66$ & ${}_{\pm 1.06}$ & $+3.25$ & ${}_{\pm 3.68}$ & 10/15 & 2/15 \\
$E = 4$ & $+13.93$ & ${}_{\pm 12.76}$ & $+0.19$ & ${}_{\pm 0.49}$ & $+0.43$ & ${}_{\pm 0.74}$ & $+0.64$ & ${}_{\pm 1.04}$ & $+3.63$ & ${}_{\pm 2.85}$ & 9/15 & 4/15 \\
\midrule
\multicolumn{13}{l}{\textit{FX}} \\
$\alpha = 0.001$ & $+0.40$ & ${}_{\pm 0.69}$ & $+0.07$ & ${}_{\pm 0.14}$ & $+0.16$ & ${}_{\pm 0.23}$ & $+0.07$ & ${}_{\pm 0.20}$ & $-0.38$ & ${}_{\pm 0.51}$ & 12/15 & 1/15 \\
$\alpha = 0.003$ & $+0.42$ & ${}_{\pm 1.23}$ & $+0.07$ & ${}_{\pm 0.25}$ & $+0.20$ & ${}_{\pm 0.40}$ & $-0.06$ & ${}_{\pm 0.56}$ & $-0.26$ & ${}_{\pm 0.64}$ & 9/15 & 0/15 \\
$\alpha = 0.01$ & $+1.02$ & ${}_{\pm 1.15}$ & $+0.22$ & ${}_{\pm 0.25}$ & $+0.46$ & ${}_{\pm 0.38}$ & $+0.18$ & ${}_{\pm 0.40}$ & $-0.61$ & ${}_{\pm 0.93}$ & 14/15 & 2/15 \\
$\alpha = 0.03$ & $+0.83$ & ${}_{\pm 1.35}$ & $+0.22$ & ${}_{\pm 0.29}$ & $+0.45$ & ${}_{\pm 0.48}$ & $+0.27$ & ${}_{\pm 0.45}$ & $-0.68$ & ${}_{\pm 0.95}$ & 13/15 & 2/15 \\
$\alpha = 0.1$ & $+1.55$ & ${}_{\pm 1.06}$ & $+0.41$ & ${}_{\pm 0.17}$ & $+0.75$ & ${}_{\pm 0.30}$ & $+0.70$ & ${}_{\pm 0.30}$ & $-0.97$ & ${}_{\pm 1.00}$ & 14/15 & 10/15 \\
\cmidrule(l){1-13}
$E = 1$ & $+0.51$ & ${}_{\pm 1.08}$ & $+0.10$ & ${}_{\pm 0.20}$ & $+0.24$ & ${}_{\pm 0.32}$ & $-0.16$ & ${}_{\pm 0.41}$ & $-0.32$ & ${}_{\pm 0.57}$ & 9/15 & 3/15 \\
$E = 2$ & $+1.02$ & ${}_{\pm 1.15}$ & $+0.22$ & ${}_{\pm 0.25}$ & $+0.46$ & ${}_{\pm 0.38}$ & $+0.18$ & ${}_{\pm 0.40}$ & $-0.61$ & ${}_{\pm 0.93}$ & 14/15 & 4/15 \\
$E = 3$ & $+1.00$ & ${}_{\pm 1.57}$ & $+0.21$ & ${}_{\pm 0.32}$ & $+0.44$ & ${}_{\pm 0.49}$ & $+0.17$ & ${}_{\pm 0.51}$ & $-0.74$ & ${}_{\pm 0.90}$ & 13/15 & 3/15 \\
$E = 4$ & $+0.64$ & ${}_{\pm 1.53}$ & $+0.17$ & ${}_{\pm 0.28}$ & $+0.38$ & ${}_{\pm 0.42}$ & $+0.11$ & ${}_{\pm 0.44}$ & $-0.45$ & ${}_{\pm 0.84}$ & 12/15 & 5/15 \\
\bottomrule
\end{tabular}}
\end{table}

\paragraph{Step size.} On DJ30, gains rise with the step size up to
$\alpha = 0.01$ and flatten or decline beyond it: mean return is similar at
$\alpha = 0.01$ and $0.03$ ($+14.92$ vs.\ $+14.20$~pp), but Sharpe falls from
$+0.38$ to $+0.24$ and the increase in maximum drawdown over Base roughly
doubles ($+1.57$ to $+3.26$~pp). At $\alpha = 0.1$, mean $\Delta$Sharpe drops
to $+0.03$. The decline is concentrated in 2024, the
year with the weakest forecaster, where Sharpe falls below the base policy
for all five algorithms at $\alpha \ge 0.03$. On FX, gains generally increase
with $\alpha$: $\alpha = 0.1$ attains the largest mean improvement on every
metric, raising mean return from $+1.02$ to $+1.55$~pp, so our shared step
size is conservative for this market.

\paragraph{Adaptation epochs.} With $E = 1$, $\fpilotabs$ retains most of its
DJ30 gain ($+9.75$ vs.\ $+14.92$~pp) but roughly halves the FX gain, improving
FX Sharpe in only 9 of 15 algorithm--years. In DJ30 2023, $E \in \{3, 4\}$
gives the largest returns for every algorithm, whereas in DJ30 2024 Sharpe
falls below the base policy for four of five algorithms at $E = 3$ and all
five at $E = 4$, and the mean DJ30 drawdown increase over Base roughly doubles
relative to $E = 2$ ($+1.57$ to $+3.25$ and $+3.63$~pp). On FX, $E = 2$ and
$E = 3$ perform similarly. Averaged over algorithm--years, $E = 2$ attains the
highest mean return and Sharpe in both markets among the values tested,
although it is best in only a minority of individual algorithm--years.

Mean return and Sharpe improvements over Base are positive at every tested
value of $\alpha$ and $E$ in both markets: these settings affect the
magnitude of the gains but not their direction. Both sweeps are consistent
with the appropriate magnitude of adaptation depending on forecast skill: the
clearest degradation from larger updates occurs in DJ30 2024, the year with
the weakest forecaster ($R^2 = 0.005$).

\end{document}